\documentclass[]{bytedance_seed}

\makeatletter
\AtBeginDocument{%
  \setlength{\headheight}{34pt}%
  \setlength{\headsep}{18pt}%
}

\let\old@maketitle\maketitle
\renewcommand{\maketitle}{%
  \old@maketitle
  \vspace{8pt}%
}
\makeatother

\usepackage[utf8]{inputenc}
\usepackage[T1]{fontenc}

\usepackage{amsmath}
\usepackage{amsfonts}
\usepackage{amssymb}
\usepackage{bbm}
\usepackage{nicefrac}

\usepackage{graphicx}
\usepackage{subcaption}
\usepackage{wrapfig}
\graphicspath{{figures/}{draft_figures/}}

\usepackage{booktabs}
\usepackage{multirow}
\usepackage{threeparttable}
\usepackage{tabularx}
\usepackage{tabularray}
\usepackage{array}
\usepackage{colortbl}
\usepackage{makecell}
\usepackage{ragged2e}

\usepackage{microtype}
\usepackage{float}
\usepackage{enumitem}
\usepackage{multicol}
\usepackage{xspace}
\usepackage{pifont}
\usepackage{fontawesome5}
\usepackage{simpleicons}
\usepackage{dirtree}
\usepackage[most]{tcolorbox}
\tcbuselibrary{listings,breakable,skins}
\microtypesetup{expansion=false}

\usepackage{hyperref}
\usepackage{url}

\newcommand{\newxcommand}[2]{\newcommand{#1}{#2\xspace}}
\newxcommand{\numberOfParticipants}{$10$}
\newxcommand{\datasetHours}{$52.9$}
\newxcommand{\datasetName}{\texttt{SuperMemory-VQA}}
\newxcommand{\numQA}{$4,853$}
\newxcommand{\dataURL}{\url{https://huggingface.co/datasets/OSU-AIoT-MLSys-Lab/SuperMemory-VQA}}
\newxcommand{\codeURL}{\url{https://github.com/AIoT-MLSys-Lab/supermemory-vqa}}

\newcommand{\modelname}[1]{%
  \mbox{\textcolor{black!80}{#1}}%
}

\definecolor{NeutralText}{HTML}{1F2933}
\definecolor{HeaderGray}{HTML}{E9E9E9}
\definecolor{LightGray}{HTML}{F5F5F5}
\definecolor{RuleGray}{HTML}{7D838C}
\definecolor{WarmNote}{HTML}{FBF8F1}

\definecolor{HeaderBg}{HTML}{2B3A4A}
\definecolor{HeaderFg}{HTML}{FFFFFF}

\definecolor{tblHeader}{HTML}{DCE6F1}
\definecolor{tblSubHeader}{HTML}{EEF3F8}
\definecolor{tblBest}{HTML}{FFF1B8}
\definecolor{tblOurs}{HTML}{E6F0E8}
\definecolor{tblObject}{HTML}{FFF8E8}
\definecolor{tblInput}{HTML}{F1F6FB}
\definecolor{tblOutput}{HTML}{EAF4EC}

\colorlet{tblText}{NeutralText}
\colorlet{tblRule}{RuleGray}
\colorlet{tblGroup}{HeaderGray}
\colorlet{tblStripe}{LightGray}
\colorlet{tblNote}{WarmNote}

\definecolor{resHeader}{HTML}{DCE6F1}
\definecolor{resHeaderFg}{HTML}{1A1A1A}
\definecolor{resSubHeader}{HTML}{EEF3F8}

\definecolor{colorGemini}{HTML}{1A73E8}
\definecolor{colorGPT}{HTML}{10A37F}
\definecolor{colorInternVL}{HTML}{F44336}
\definecolor{colorQwen}{HTML}{FF9800}
\definecolor{colorGemma}{HTML}{9C27B0}

\definecolor{resOpenGroup}{HTML}{D8EBDD}
\definecolor{resOpenRow}{HTML}{F4FBF6}

\definecolor{resClosedGroup}{HTML}{E8E1F2}
\definecolor{resClosedRow}{HTML}{FAF7FD}

\definecolor{resBest}{HTML}{FFF1B8}
\definecolor{resRule}{HTML}{7D838C}
\definecolor{resText}{HTML}{1F2933}

\colorlet{tblOpen}{tblInput}
\colorlet{tblClosed}{tblOutput}

\definecolor{bgBlue}{HTML}{EFF6FF}
\definecolor{bgGreen}{HTML}{ECFDF5}
\definecolor{bgOrange}{HTML}{FFF7ED}
\definecolor{bgPurple}{HTML}{FAF5FF}
\definecolor{bgRed}{HTML}{FEF2F2}
\definecolor{bgYellow}{HTML}{FEFCE8}

\definecolor{txtBlue}{HTML}{1D4ED8}
\definecolor{txtGreen}{HTML}{047857}
\definecolor{txtOrange}{HTML}{C2410C}
\definecolor{txtPurple}{HTML}{6D28D9}
\definecolor{txtRed}{HTML}{B91C1C}
\definecolor{txtYellow}{HTML}{A16207}

\definecolor{TaskBlue}{HTML}{1F4E79}
\definecolor{TaskTint}{HTML}{EAF2F8}
\definecolor{RowTint}{HTML}{F7F9FC}
\definecolor{QuestionBlue}{HTML}{1F4E79}
\definecolor{AnswerGreen}{HTML}{2E6B3F}

\definecolor{qaCorrectBg}{HTML}{DDEBD7}
\definecolor{qaCorrectBorder}{HTML}{7EA36E}

\definecolor{qaVagueBg}{HTML}{F2E6BD}
\definecolor{qaVagueBorder}{HTML}{C2A84A}

\definecolor{qaWrongBg}{HTML}{EBCFCB}
\definecolor{qaWrongBorder}{HTML}{B56B63}

\definecolor{qaNABg}{HTML}{DCE6F6}
\definecolor{qaNABorder}{HTML}{6F8FC8}

\definecolor{schemaBg}{HTML}{F7F9FC}
\definecolor{schemaFrame}{HTML}{8F9AA8}
\definecolor{schemaTitle}{HTML}{DCE6F1}
\definecolor{schemaText}{HTML}{1F2933}

\newcolumntype{Y}{>{\RaggedRight\arraybackslash}X}

\newtcbox{\correcttag}{
  on line,
  colback=qaCorrectBg,
  colframe=qaCorrectBorder,
  boxrule=0.5pt,
  arc=2mm,
  left=5pt,
  right=5pt,
  top=2pt,
  bottom=2pt,
  boxsep=0pt
}

\newtcbox{\vaguetag}{
  on line,
  colback=qaVagueBg,
  colframe=qaVagueBorder,
  boxrule=0.5pt,
  arc=2mm,
  left=5pt,
  right=5pt,
  top=2pt,
  bottom=2pt,
  boxsep=0pt
}

\newtcbox{\wrongtag}{
  on line,
  colback=qaWrongBg,
  colframe=qaWrongBorder,
  boxrule=0.5pt,
  arc=2mm,
  left=5pt,
  right=5pt,
  top=2pt,
  bottom=2pt,
  boxsep=0pt
}

\newtcbox{\natag}{
  on line,
  colback=qaNABg,
  colframe=qaNABorder,
  boxrule=0.5pt,
  arc=2mm,
  left=7pt,
  right=7pt,
  top=2pt,
  bottom=2pt,
  boxsep=0pt
}

\DeclareRobustCommand{\CorrectTag}{\correcttag{\strut Correct}}
\DeclareRobustCommand{\VagueTag}{\vaguetag{\strut Vague}}
\DeclareRobustCommand{\WrongTag}{\wrongtag{\strut Wrong}}
\DeclareRobustCommand{\NATag}{\natag{\strut N/A}}

\newtcolorbox{schemabox}[2][]{
  enhanced,
  breakable,
  title={#2},
  colback=schemaBg,
  colframe=schemaFrame,
  coltitle=schemaText,
  colbacktitle=schemaTitle,
  fonttitle=\bfseries\footnotesize,
  boxrule=0.5pt,
  arc=2mm,
  left=2mm,
  right=2mm,
  top=1mm,
  bottom=1mm,
  #1
}

\definecolor{ResourceLinkBlue}{HTML}{3366BB}
\definecolor{HuggingFaceGold}{HTML}{FFCC4D}

\newcommand{\resourceIconLink}[4]{%
  \href{#1}{\textcolor{#3}{\raisebox{-0.05em}{#2}}\hspace{0.35em}\textcolor{ResourceLinkBlue}{#4}}%
}

\title{SuperMemory-VQA: \\ An Egocentric Visual Question-Answering Benchmark for Long-Horizon Memory}

\author[1]{Samiul Alam}
\author[1]{Shakhrul Iman Siam}
\author[2]{Michael J. Proulx}
\author[2]{James Fort}
\author[2]{Richard Newcombe}
\author[2]{Hyo Jin Kim}
\author[1]{Mi Zhang}

\affiliation{
  {\fontsize{11}{11}\selectfont
  \textsuperscript{1}The Ohio State University,
  \textsuperscript{2}Meta\\[2pt]

  \resourceIconLink{https://supermemory-vqa.github.io}{\faHome}{black}{Project Page} \quad
  \resourceIconLink{https://github.com/AIoT-MLSys-Lab/supermemory-vqa}{\simpleicon{github}}{black}{GitHub Code} \quad
  \resourceIconLink{https://huggingface.co/datasets/OSU-AIoT-MLSys-Lab/SuperMemory-VQA}{\simpleicon{huggingface}}{HuggingFaceGold}{Dataset}
  }
}

\abstract{
AI glasses present a compelling platform for AI agents to serve as personalized memory assistants.
To be genuinely useful, such systems must move beyond short-term video comprehension and address memory gaps that humans experience for practical, personal, or social purposes over longitudinal egocentric video streams.
However, existing egocentric datasets predominantly focus on action recognition or generic QAs from short clips, measuring perceptual capabilities rather than realistic human memory needs.
We introduce \textbf{SuperMemory-VQA}, an egocentric visual question answering (VQA) dataset for evaluating AI assistants on practical, long-horizon memory tasks.
It contains \textbf{52.9 hours} of everyday activities recorded with AI glasses, including synchronized RGB video, audio transcription, eye gaze, IMU, and SLAM trajectories.
Through a human-verified annotation pipeline, we construct grounded \textbf{4,853} question-answer pairs that span object and location memory, intent recall, visual scene recall, timeline reconstruction, conversational memory, and in-context retrieval.
Each question is posed as multiple-choice with an explicit ``unanswerable'' option to test hallucination robustness.
Benchmarking leading agentic frameworks and LLM backbones reveals that existing systems remain far from reliable on real-world memory tasks, highlighting the need for new architectures for grounded AI memory that can answer only when evidence is sufficient.
A participant survey further supports that our questions are realistic, useful, and aligned with everyday memory needs.
}

\begin{document}
\maketitle
\begingroup
\renewcommand{\thefootnote}{}
\footnotetext{Dataset collection and experiments were conducted by OSU researchers following an IRB-reviewed protocol.}
\endgroup

\setlist[itemize]{leftmargin=5.5mm}

\section{Introduction}
\label{sec:intro}

The integration of AI agents into AI glasses represents a transformative leap in personal computing. By continuously observing a user's daily life, these agents have the potential to act as personalized memory systems -- helping users locate misplaced items, revisit conversations, and piece together past events. However, delivering genuine utility requires moving beyond short-term visual understanding to systems that process continuous, multi-modal sensor streams over extended periods to answer natural questions rooted in real human memory gaps.

\newpage
Progress toward reliable AI memory assistants is currently hindered by limitations in both model capability and evaluation dataset.
On the modeling front,
even models with million-token context windows~\citep{team2023gemini} degrade over long inputs and suffer from the ``lost in the middle'' effect~\citep{liu2024lost}. Retrieval-Augmented Generation (RAG)~\citep{lewis2020retrieval} can mitigate the context limit, but often fails at compositional, multi-hop reasoning over vast temporal gaps~\citep{xue2025adavideorag}.
Concurrently, existing egocentric datasets do not fully capture realistic memory demands. Most of them focus on evaluating action recognition or VQAs focusing on short-term visual perception capabilities~\citep{epickitchens100,qaego4d2022,egoschema2023,perrett2025hdepic}. Recently, EgoLife~\citep{yang2025egolife} introduced a week-long egocentric data collection.
However, it does not systematically evaluate whether systems can answer natural, user-centered memory questions across multi-session recordings.

\begin{figure}[t]
    \centering
\includegraphics[width=\linewidth]{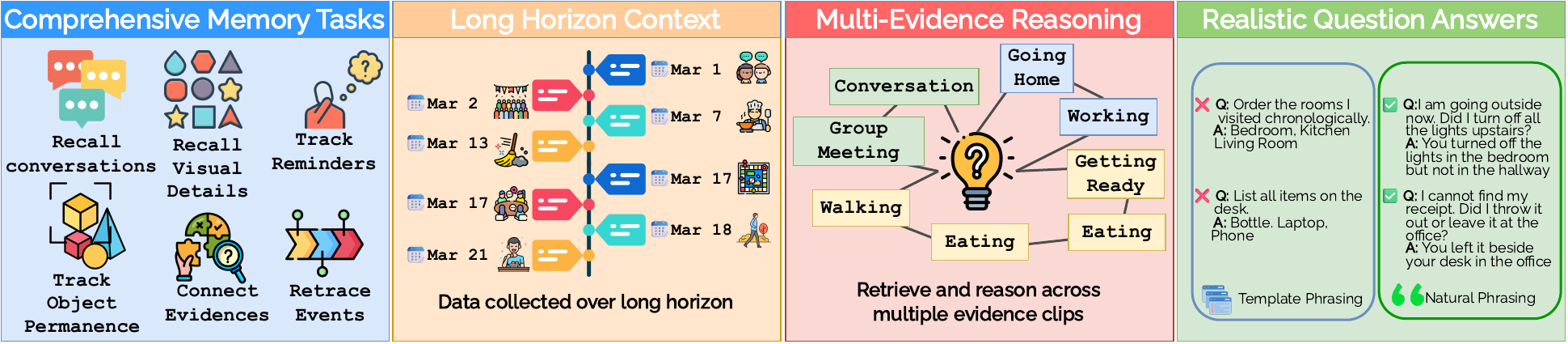}
    \vspace{-4mm}
    \caption{Overview of \textbf{SuperMemory-VQA}. \datasetName advances the state-of-the-arts in four dimensions: 1) \textbf{comprehensive memory tasks}: includes six user-evaluated and commonly encountered memory tasks; 2:) \textbf{long-horizon context}: data collected over a long horizon spanning days and weeks; 3) \textbf{multi-evidence reasoning}: requires retrieving and reasoning across multiple parts of recording; and 4) \textbf{realistic question answers}: employs natural, context-grounded phrasing rather than rigid, template-based questions and answers.
    }
    \label{fig:overview}
    \vspace{-0mm}
\end{figure}

In this work, we introduce \textbf{SuperMemory-VQA}, an egocentric visual question answering dataset designed around questions that people would realistically ask an AI memory assistant. \datasetName contains \datasetHours hours of everyday activities, recorded by participants wearing AI glasses~\citep{project_aria} that capture synchronized RGB video, spatial audio, eye gaze, IMU, and SLAM trajectories. As summarized in~\Cref{fig:overview}, \datasetName emphasizes four critical properties missing from prior benchmarks: comprehensive memory tasks, long-horizon context, multi-evidence reasoning, and realistic question answers.
Specifically, it covers six user-evaluated and commonly encountered memory tasks, including object and location memory, intent recall, visual scene recall, timeline reconstruction, conversational memory, and in-context retrieval, resulting in \numQA QAs in total.
A participant survey supports this motivation: participants judged questions generated from their own recordings as realistic, useful, and aligned with memory needs they might experience in daily life.

To construct the dataset, we developed a scalable, human-in-the-loop annotation pipeline. An agentic system first generates grounded descriptions and drafts question-answer pairs, followed by rigorous automated checks and final human verification and refinement.
Crucially, to evaluate both answer quality and hallucination robustness, each multiple-choice question includes ordered answer choices, including accurate, vague, incorrect, and unanswerable options.

We benchmarked two state-of-the-art agentic frameworks, Video-RAG~\citep{luo2024videorag} and EgoButler~\citep{yang2025egolife}, with leading open-source and proprietary Vision Language Models (VLMs). The results show that current systems remain far from reliable: they struggle with answerability detection, long temporal gaps, and evidence integration across multiple moments.
Our main contributions are summarized as follows:

\begin{itemize}
    \item We introduce \datasetName, a \datasetHours hours of multi-modal egocentric VQA dataset with 4,853 QAs for evaluating AI assistants on practical, long-horizon memory tasks.
    \item We develop a scalable human-in-the-loop pipeline for generating grounded, hallucination-resistant Q\&A pairs from continuous egocentric video.
    \item We support the perceived utility and practical relevance of \datasetName through a participant survey, showing that users view the generated questions as practical, relevant, and aligned with everyday AI memory needs.
    \item We benchmark state-of-the-art video understanding and RAG-based systems, exposing major gaps in long-horizon retrieval, grounded reasoning, and knowing when there is enough evidence to answer. We believe \datasetName will play an instrumental role in developing next-generation AI memory assistants to meet everyday memory needs.
\end{itemize}

\section{Related Work}

\begin{figure}[t]
    \centering
    \includegraphics[width=\linewidth]{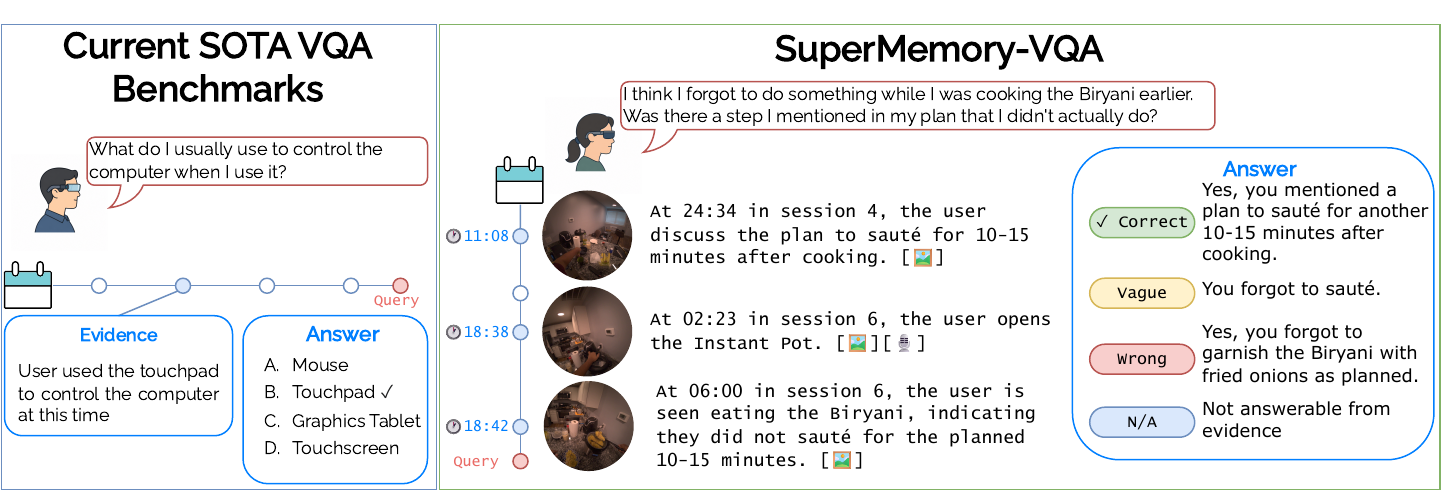}
    \caption{Comparison of \textbf{SuperMemory-VQA} with SOTA benchmarks. The example on the left is from EgoLife~\citep{yang2025egolife}: the question asks what the user usually uses to control a computer, and the answer is supported by a direct observation that the user used a touchpad at the shown time. In contrast, the \datasetName example on the right asks whether the user skipped a planned cooking step; answering requires linking a spoken plan from one clip with later visual evidence that the user opened the Instant Pot and ate the food without the planned saut\'e step. This illustrates how \datasetName emphasizes practical AI memory questions with long-horizon, multi-modal evidence and ordered answer choices.}
    \label{fig:teaser}
    \vspace{-0mm}
\end{figure}

\noindent \textbf{Egocentric Multimodal Datasets.} Egocentric vision captures situated human behavior. Early datasets used eye-tracking glasses for gaze and action annotations~\citep{fathi2012learning, gtea, egtea}, but were small and narrow. Later efforts scaled to hundreds or thousands of hours~\citep{epickitchens100,Damen_2018_ECCV,grauman2022ego4d}, though most still center on RGB video. Project Aria~\citep{project_aria} expanded the sensor suite with synchronized RGB, eye tracking, spatial audio, IMU, and 3D scene context across growing egocentric recordings~\citep{grauman2024egoexo4d, ma2024nymeria, lv2024aria, perrett2025hdepic}. Recent work~\citep{yang2025reading} demonstrated that gaze and other non-visual cues improve video understanding.
Despite this progress, existing benchmarks do not evaluate the longitudinal, multi-session memory that practical AI assistants require---recalling where objects were left across days, reconstructing daily timelines, or tracking conversational commitments.
The closest work to ours is EgoLife~\citep{yang2025egolife}, an egocentric dataset spanning a week. However, as illustrated in \Cref{fig:teaser}, it still lacks realistic, practical queries for human memory assistance.
In contrast, \datasetName fills this gap with extended multimodal egocentric data (video, gaze, audio, trajectory, IMU, and SLAM) and structured protocols for real-world memory tasks.

\noindent \textbf{Long Video Understanding.} Retrieval-Augmented Generation (RAG)~\citep{lewis2020retrieval, gao2023ragsurvey} retrieves external information beyond limited language model contexts. In video, captions often form a retrievable text corpus, while multimodal RAG indexes frames directly~\citep{reddy2025video, wan2025clamr, luo2024videorag}. Video-RAG~\citep{luo2024videorag} jointly retrieves frames, ASR, OCR, and detections to preserve visual details, but frame-level retrieval remains hard to index and scale~\citep{reddy2025video, wan2025clamr}. Structured variants include VideoRAG~\citep{ren2025videorag} for parallel text-, visual-, and graph-based clip retrieval; AdaVideoRAG~\citep{xue2025adavideorag} for adaptive retrieval; GraphVideoAgent~\citep{chu2025graphvideoagent}, which extends VideoAgent~\citep{wang2024videoagent} with caption-derived graphs; and VideoMindPalace~\citep{huang2025building}, whose room-level spatial graphs limit open-ended use. MemVid~\citep{yuan2025memory} adds planning and analysis agents for episodic video QA. These methods are strong baselines. However, our tasks---object tracking, conversation recall, and timeline reconstruction---require compositional reasoning over entity states across sessions and heterogeneous modalities (visual, auditory, spatial). \datasetName provides task categories and ground truth beyond single-video QA, covering multi-day episodic memory.

\section{SuperMemory-VQA Dataset}
\label{sec:dataset}

\begin{figure*}
    \centering
    \includegraphics[width=\linewidth]{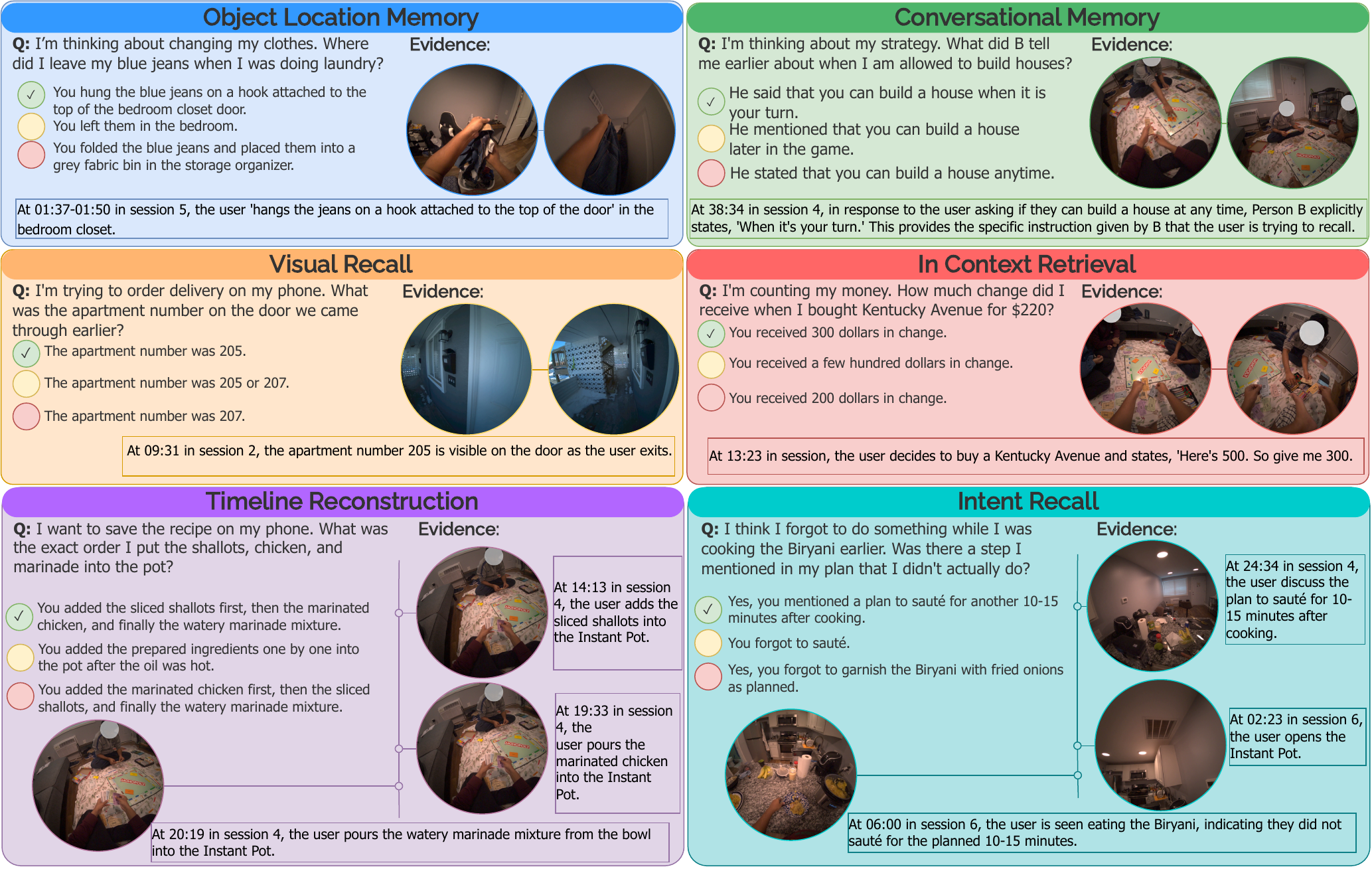}
    \caption{Example VQA pairs from each of the six task categories.}
    \label{fig:tasks}
    \vspace{-0mm}
\end{figure*}

\subsection{Overview}
\label{sec:methods.overview}

The \datasetName dataset contains \datasetHours hours of egocentric recordings of everyday activities collected from ten participants wearing \textup{Gen 1 Meta Aria Glasses}~\citep{project_aria}. 
The recordings are multimodal, including egocentric RGB video (1408x1408, 30fps), dual SLAM streams (640x480, 30fps), eye-tracking (320x240, 60fps), and 7-channel audio (48kHz).
%
Participants recorded data in both indoor and outdoor settings and followed a generic script, like following a cooking recipe or playing board games. Activities included household chores such as cooking, cleaning, and organizing, as well as group activities like playing games and having conversations.
Each participant contributed $3$ to $12$ hours of recordings across multiple sessions, including recordings from three participants spanning multiple days (up to two weeks). 
\Cref{sec:methods.protocol} describes the full hardware specification, protocol, anonymization procedure, and release details.

Besides the recordings, the \datasetName dataset also includes QA pairs that reflect what a human user would \emph{want} to remember for practical, personal, or social purposes. Specifically, \datasetName covers six commonly encountered memory tasks:
\vspace{-0mm}
\begin{itemize}
\item \textbf{Object \& Location Memory} recalls the last known position of an object and its trajectory across time and locations, based on the concept of \emph{episodic memory}~\citep{tulving1972episodic}.
\item \textbf{Conversational Memory} recalls spoken facts, commitments,
deferred answers, and mid-conversation corrections. Also known as Dialogue State Tracking (DST), this is a formal component within task-oriented conversational AI \citep{henderson2014word, wu2019transferable}.
\item \textbf{Visual Scene Recall} retrieves visual details such as text, screens, manuals, ingredients, or scene contents. While rooted in episodic memory, it concurrently evaluates the system's \emph{semantic memory}~\citep{tulving1972episodic, squire2004memory}.
\item \textbf{In-Context Retrieval} combines current context with prior facts. It evaluates the system's capacity for \emph{relational memory}~\citep{eichenbaum1997declarative, cohen1997memory} -- the cognitive ability to represent and navigate associations between independent elements of an experience.
\item \textbf{Timeline Reconstruction} chronologically sequences events, evaluating the temporal aspect of episodic memory, and \emph{procedural memory}~\citep{cohen1980preserved, squire2004memory}.
\item \textbf{Intent Recall} recovers stated or implied goals, reminders, and intended future actions, capturing~\emph{prospective memory}
\citep{einstein1990normal, brandimonte1996prospective}.
\end{itemize}

\Cref{fig:tasks} shows example QA pairs of all these tasks. We also asked participants to map their reasoning when answering questions about their own recordings to a defined set of cognitive memory strategies. \Cref{fig:skill_answer_sankey} shows this mapping and validates the taxonomy through participant-described memory strategies.


\subsection{Comparison to Existing Datasets}
\label{sec:comparison}

\begin{table*}[t]
\centering
\caption{Comparison of \textbf{SuperMemory-VQA} with existing egocentric benchmarks.}
\label{tab:dataset_comparison}

\begingroup
\footnotesize
\setlength{\tabcolsep}{2pt}
\renewcommand{\arraystretch}{1.08}
\arrayrulecolor{tblRule}
\newcommand{\tblb}[1]{{\fontsize{7.6pt}{8.4pt}\selectfont\bfseries #1}}
\newcommand{\tblbm}[1]{{\fontsize{7.6pt}{8.4pt}\selectfont\boldmath\bfseries #1}}
\begin{tabularx}{\textwidth}{@{}
  >{\raggedright\arraybackslash}m{0.160\textwidth}
  >{\raggedright\arraybackslash}m{0.140\textwidth}
  >{\centering\arraybackslash}m{0.045\textwidth}
  >{\centering\arraybackslash}m{0.074\textwidth}
  >{\centering\arraybackslash}m{0.060\textwidth}
  >{\centering\arraybackslash}m{0.092\textwidth}
  >{\centering\arraybackslash}m{0.072\textwidth}
  >{\raggedright\arraybackslash}X
@{}}
\specialrule{0.08em}{0pt}{0pt}
\rowcolor{tblHeader}[3pt][3pt]
\tblb{Dataset} & \tblb{Focus} & \tblb{Hrs} & \tblb{Context} & \tblb{QAs} & \tblb{\makecell{Multi-\\Evidence}} & \tblb{Natural} & \tblb{Evaluation} \\
\specialrule{0.05em}{0pt}{0pt}

\rowcolor{tblStripe}[3pt][3pt]
\textsc{EPIC-KITCHENS-100} \citep{epickitchens100}
& Action Rec.
& 100
& $\approx$8.5m
& --
& --
& No
& Verb-noun labels \& narrations \\

Ego4D \citep{grauman2022ego4d}
& Ego Activities
& 3{,}670
& $\approx$23m
& --
& --
& No
& Temporal/spatial localization labels \\

\rowcolor{tblStripe}[3pt][3pt]
EgoSchema \citep{mangalam2023_egoschema}
& Long Video QA
& $>250$
& 3m
& 5{,}063
& Single
& No
& 5-way MCQ over clips \\

HoloAssist \citep{wang2023holoassist}
& Assistance
& 166
& $\approx$5--10m
& --
& --
& No
& Mistakes, interventions \\

\rowcolor{tblStripe}[3pt][3pt]
AEA \citep{lv2024aria}
& Spatial
& $>7.5$
& $\approx$5m
& --
& --
& N/A
& No QA; sensor/MPS outputs \\

Nymeria \citep{ma2024nymeria}
& Motion
& 300
& $\approx$15m
& --
& --
& N/A
& Motion/action descriptions \\

\rowcolor{tblStripe}[3pt][3pt]
EgoLife \citep{yang2025egolife}
& Life Assistant
& 300
& $>$1h
& 6{,}000
& Limited
& No
& Generic MCQ w/ evidence timestamps \\

\specialrule{0.05em}{0pt}{0pt}
\rowcolor{tblOurs}[3pt][3pt]
\tblb{SuperMemory-VQA}\ \tblb{(Ours)}
& \tblb{SuperMemory{\raisebox{0.35ex}{\scriptsize *}}}
& \tblbm{\datasetHours}
& \tblbm{$>$1h}
& \tblb{4{,}853}
& \tblb{34\%}
& \tblb{Yes}
& \tblb{Ordered MCQs with time spans} \\

\specialrule{0.08em}{0pt}{0pt}
\end{tabularx}
\endgroup

\vspace{0.4ex}
\begin{minipage}{\textwidth}
\textbf{--}: no QA benchmark or QA evidence annotations. Multi-evidence means a QA requires more than one supporting timestamp or clip.\\
\textsuperscript{*} Encompasses multiple aspects of supermemory (\Cref{app:tasks}).
\end{minipage}
\vspace{-0mm}
\end{table*}

Recent egocentric and video benchmarks advance first-person activity understanding, long-form video QA, and multimodal life logging, including \textsc{EPIC-KITCHENS}~\citep{Damen_2018_ECCV}, Ego4D~\citep{grauman2022ego4d}, EgoSchema~\citep{mangalam2023_egoschema}, Video-MME~\citep{fu2025video}, AEA~\citep{lv2024aria}, Nymeria~\citep{ma2024nymeria}, and EgoLife~\citep{yang2025egolife}. As \Cref{tab:dataset_comparison} shows, these datasets mainly emphasize action recognition, general video comprehension, or generic template-based QAs. 
In contrast, \datasetName advances state-of-the-arts from the following perspectives:


\vspace{0.5mm}
\begin{itemize}
\item\textbf{Natural Phrasing.}
\datasetName uses conversational, context-dependent queries rather than template-style questions, requiring models to infer user intent, temporal reference, and memory context instead of relying on predictable syntax.

\item\textbf{Long-Horizon Context.}
Questions are grounded in recordings that last hours and sometimes span days, exceeding the practical context limits of current vision-language models~\citep{team2023gemini,team2024gemini1.5} and matching the long-tail temporal complexity of egocentric video~\citep{grauman2022ego4d}.

\item\textbf{Dense Multi-Evidence Retrieval.}
Many of our questions require linking sparse evidence across disjoint moments, such as a spoken plan in one session and actions performed much later, so systems need retrieval-augmented reasoning~\citep{lewis2020retrieval} and temporal abstraction to avoid ``lost-in-the-middle'' failures~\citep{liu2024lost}.

\item\textbf{Grounded Multi-modal Reasoning.}
Memory assistance from egocentric data requires aligning video, audio, gaze, motion, and spatial context to track actions, object states, and user intent over time. \datasetName also stresses efficient multi-modal evidence use: for example, \Cref{fig:teaser} shows how auditory cues of steam venting can localize the Instant Pot event before visual confirmation.

\item\textbf{Epistemic Calibration and Hallucination Robustness.}
Instead of standard independent multiple-choice questions~\citep{hendrycks2021measuring,srivastava2022beyond}, \datasetName uses ordered answer choices that distinguish correct, vague, wrong, and unanswerable responses. Inspired by reading-comprehension abstention settings~\citep{rajpurkar2018know}, this design tests whether models can avoid confident hallucinations when evidence is missing~\citep{lin2022truthfulqa,li2023halueval} and provides ranked feedback (Correct $>$ Vague $>$ Wrong) for alignment methods such as DPO~\citep{ouyang2022training,rafailov2023direct}.
\end{itemize}

\subsection{Data Collection Process}
\label{sec:methods.protocol}
We recruited a total of ten participants.
Under an IRB-approved protocol, these participants wore Gen 1 Meta Aria Glasses and completed guided 30--45 minute sessions in a simulated home environment, including calibration, exploration, and loosely scripted indoor and outdoor tasks using assigned pseudonyms for privacy. Participant demographics are withheld to maintain privacy.
Raw audio is withheld; instead, we provide WhisperX-transcribed text~\citep{bain2023whisperx} with sensitive information manually removed. Faces and license plates are obscured using EgoBlur~\citep{raina2023egoblurresponsibleinnovationaria}, and direct interactions with non-participants were excised.
\Cref{app:data_collection.protocol} details the protocol, recruitment, scripts, and follow-up sessions; \Cref{app:data_collection.privacy} details de-identification and non-participant handling.

\subsection{Annotation Pipeline}
\label{sec:methods.pipeline}
\begin{figure}[t]
    \centering
    \begin{subfigure}[b]{0.8\textwidth}
        \centering
        \includegraphics[width=\linewidth]{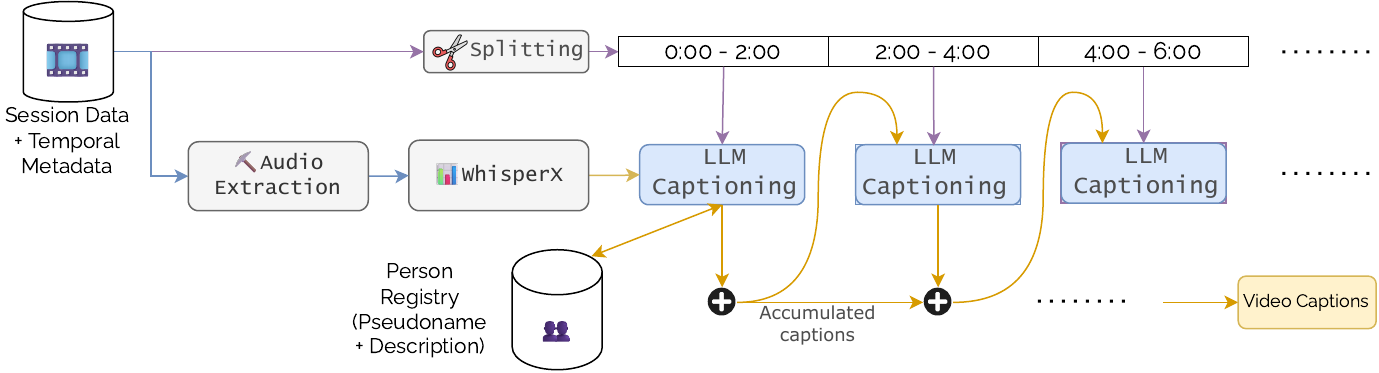}
        \caption{Phase 1: Dense Video Captioning}
        \label{fig:qa_pipeline_phase1}
    \end{subfigure}

    \vspace{1em}

    \begin{subfigure}[b]{\textwidth}
        \centering
        \includegraphics[width=0.8\linewidth]{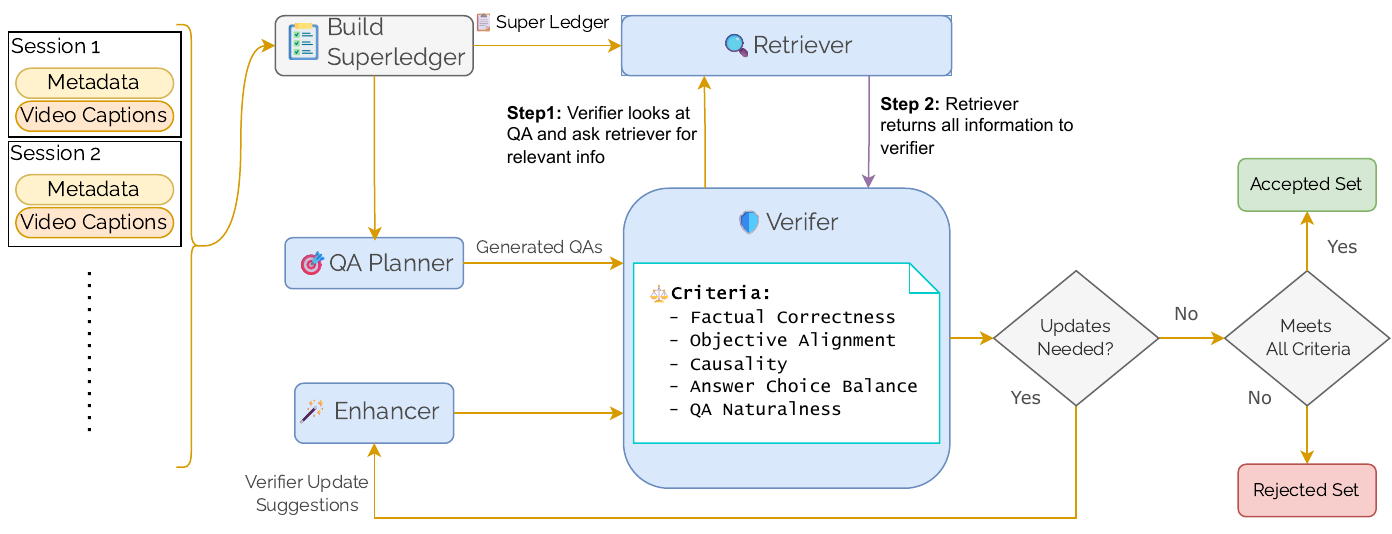}
        \caption{Phase 2: QA Generation}
        \label{fig:qa_pipeline_phase2}
    \end{subfigure}
    \caption{Illustration of the annotation pipeline.}
    \label{fig:qa_pipeline_complete}
    \vspace{-0mm}
\end{figure}

Besides the dataset, we have also developed an annotation pipeline for annonating the collected recordings. As shown in \Cref{fig:qa_pipeline_complete}, the annotation pipeline consists of two phases, each followed by human review.

\vspace{0.5mm}
\textbf{Phase 1: Dense Video Captioning.}
Video chunks and WhisperX audio transcriptions are processed by an LLM Captioning agent. Using a dynamic Person Registry, it extracts descriptions of visual actions, objects, and auditory events, aggregating them into consolidated Video Captions (\Cref{fig:qa_pipeline_phase1}).

\vspace{0.5mm}
\textbf{Phase 2: Agentic QA Generation and Human Review.}
Captions and metadata are merged into a unified ``Super Ledger''. A QA Planner proposes rationale-first~\citep{willard2023efficient}, instance-level chain-of-thought~\citep{xiao2024efficient} QA pairs targeting the dataset dimensions and tasks~\citep{wang2023selfinstruct}. A Verifier checks each pair against criteria such as factual correctness, causality, and naturalness by generating criterion-level rationales and scores~\citep{zheng2023judging} while querying a Retriever over the Super Ledger; an Enhancer iteratively refines pairs when needed. Approved pairs enter the Accepted Set for final human review (\Cref{fig:qa_pipeline_phase2}).
The Planner and Verifier agents use \modelname{gemini-3.1-pro-preview}, while the Captioning, Retriever, and Enhancer agents use \modelname{gemini-3-flash-preview}.
\Cref{app:annotation_pipeline,app:verification_annotation_format,app:human_review} detail the phases, verification schema, and human review workflow.
\vspace{-1mm}
\subsection{Dataset Statistics}
\label{sec:methods.stats}

\begin{figure}[t!]
    \centering
    \vspace{-2mm}
    \begin{minipage}[t]{0.42\linewidth}
        \centering
        \includegraphics[width=\linewidth]{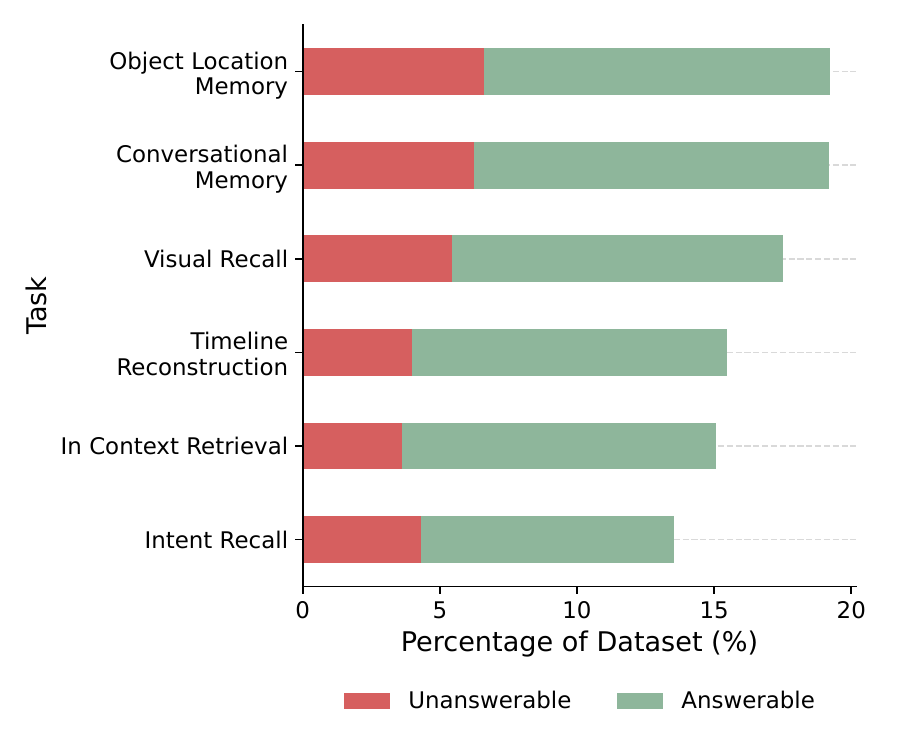}
        \captionof{figure}{Distribution of answerable and unanswerable questions per task category.}
        \label{fig:skill_distribution}
    \end{minipage}
    \hfill
    \begin{minipage}[t]{0.55\linewidth}
        \centering
        \includegraphics[width=\linewidth]{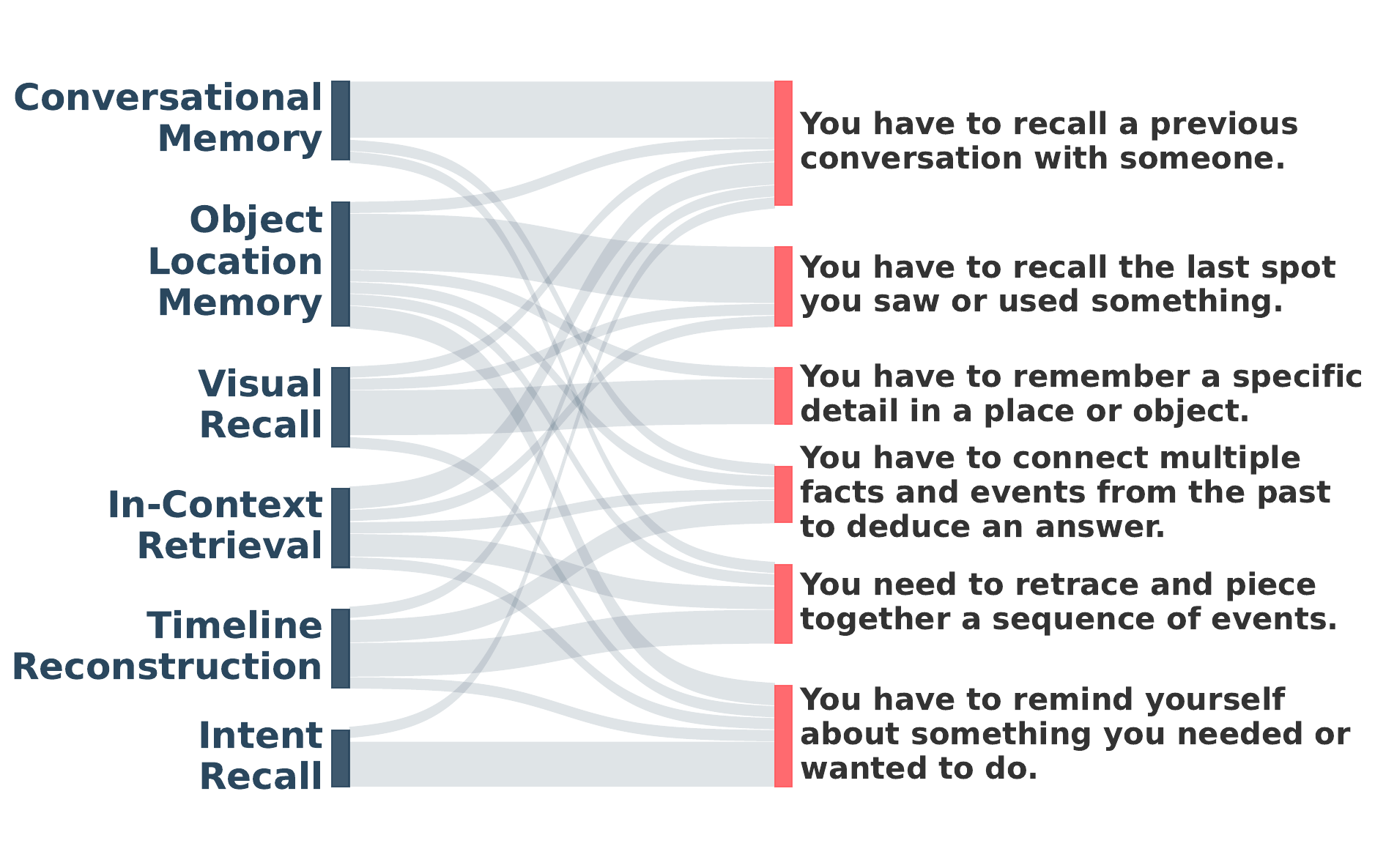}
        \captionof{figure}{Sankey diagram linking task categories to participant-described memory strategies.}
        \label{fig:skill_answer_sankey}
    \end{minipage}
    \vspace{-0mm}
\end{figure}

\datasetName includes \numQA QA pairs from \datasetHours hours of multimodal egocentric video, balanced across task types while preserving natural variation in question complexity, as shown in \Cref{fig:skill_distribution}. Object Location Memory ($\sim$19\%) and Conversational Memory ($\sim$18\%) are the most common, with Visual Recall, In Context Retrieval, Timeline Reconstruction, and Intent Recall making up the rest. \Cref{sec:data_stats} reports temporal gaps, evidence counts, and evidence-duration distributions.

\section{Experimental Setup}
\label{sec:experiment}

\subsection{Frameworks and Models}
\label{sec:experiment.baselines}
We evaluate two state-of-the-art frameworks for long-form video understanding on \texttt{SuperMemory-VQA}: Video-RAG \citep{luo2024videorag} and EgoButler \citep{yang2025egolife}, each illustrating a distinct strategy for managing long-horizon context\footnote{
We also evaluated VideoAgent~\citep{fan2024videoagent}. However, it used substantially more tokens while having worse results compared to Video-RAG and EgoButler. We report limited results on VideoAgent in \Cref{sec:additional_results.compare-videoagent}.
}.

\textbf{Video-RAG} \citep{luo2024videorag} is a training-free, single-turn retrieval-augmented framework that augments a VLM with auxiliary text extracted from the source video. Three parallel databases were precomputed per session: ASR transcripts (Whisper-X \citep{bain2023whisperx}), OCR text from sampled frames (EasyOCR \citep{easyocr}), and object detections on CLIP-selected keyframes (APE \citep{shen2024aligning}). At inference, the VLM decomposes the query into a retrieval request, each database is queried via FAISS \citep{johnson2019billion} over Contriever embeddings \citep{izacard2021unsupervised}, and the retrieved texts are concatenated with sampled frames as input to the VLM.

\textbf{EgoButler} \citep{yang2025egolife}, proposed alongside EgoLife, is the closest in spirit to \texttt{SuperMemory-VQA} as it explicitly targets ultra-long, multi-day egocentric QAs. It pairs EgoGPT, an omni-modal VLM that produces dense visual–audio captions, with EgoRAG, which recursively summarizes these captions into hour- and day-level digests to form a hierarchical memory bank. At query time, EgoRAG performs coarse-to-fine temporal localization, retrieving high-level summaries first, then narrowing to clips and passing the top-k clip captions to the VLM for answer generation.

\textbf{VLMs.}
We evaluate a diverse set of both open and closed-source VLMs under Video-RAG and EgoButler. 
The open-source VLMs we include for benchmarking are Qwen-3-VL 8B, 30B, InternVL-3.5 8B, 30B, Gemma-4-E4B IT, and Gemma-4 31B. The closed-source VLMs we include for benchmarking are Gemini-3-Flash, Gemini-3.1-Pro, GPT-5.4-mini and GPT-5.4.

\subsection{Implementation Details}

\textbf{VLMs.}
Open-source models are implemented on 4×A100 GPUs; closed-source models are accessed via official APIs. All answer-generation calls use greedy decoding (temperature 0), while internal planning and captioning calls retain each framework's defaults. 

\textbf{Video-RAG.}
Using the authors' code, session histories are partitioned into 30-minute shards with per-shard FAISS indices; at query time, the retrieval request fans out across all shards preceding the question, and the top-scoring auxiliary texts are merged. The VLM receives 32 uniformly sampled frames from the most relevant shard plus the merged texts.

\textbf{EgoButler.}
EgoGPT is first replaced with the VLMs we selected. Clip captions are generated over 30-second windows at 1 fps, using WhisperX transcripts for the audio channel. Hour and day-level summaries are produced by \modelname{gemini-3-flash-preview}.

Both Video-RAG and EgoButler receive the question and four answer choices; to enforce causality, we cut the video at the question end time and provide only preceding segments. \Cref{app:evaluation_compute} gives further compute and reproducibility details.

\subsection{Evaluation Metrics}
We benchmark the performance using three metrics: Ans-F1, QA-Acc, and QA-MRR. 
Specifically, Ans-F1 is the F1 score for the binary decision of whether a question is answerable from the available evidence or should receive the unanswerable option. 
QA-Acc is four-way multiple-choice accuracy, where only the ground-truth correct option receives credit; vague, wrong, and incorrect abstention choices are counted as incorrect. 
Lastly, QA-MRR is the mean reciprocal rank induced by the model's ordered answer scores over the answer choices, rewarding models that rank the correct answer above vague, wrong, and unsupported alternatives.

\section{Benchmarking Results}
\label{sec:results}

\subsection{Main Results}
\label{sec:results.main}

\begin{table*}[t]
\centering
\caption{Performance of open- and closed-source VLMs under Video-RAG and EgoButler.}
\label{tab:supermemory_results}
\vspace{-1mm}
\small
\begin{tblr}{
  width = \textwidth,
  colspec = {
    X[1.55,l]
    X[0.62,c]
    X[0.62,c]
    X[0.62,c]
    X[0.62,c]
    X[0.62,c]
    X[0.62,c]
  },
  rows = {
    fg=resText,
    valign=m,
    abovesep=0.55pt,
    belowsep=0.55pt
  },
  row{1} = {
    bg=resHeader,
    fg=resHeaderFg,
    font=\bfseries,
    halign=c,
    abovesep=0.7pt,
    belowsep=0.7pt
  },
  row{2} = {
    bg=resSubHeader,
    font=\bfseries,
    halign=c,
    abovesep=0.45pt,
    belowsep=0.45pt
  },
  row{3} = {
    bg=resOpenGroup,
    font=\bfseries,
    abovesep=0.45pt,
    belowsep=0.45pt
  },
  row{4-9} = {bg=resOpenRow},
  row{10} = {
    bg=resClosedGroup,
    font=\bfseries,
    abovesep=0.45pt,
    belowsep=0.45pt
  },
  row{11-14} = {bg=resClosedRow},
  column{1} = {cmd=\RaggedRight},
  cell{1}{2} = {c=3}{bg=resHeader,fg=resHeaderFg,font=\bfseries},
  cell{1}{5} = {c=3}{bg=resHeader,fg=resHeaderFg,font=\bfseries},
  cell{3}{1} = {c=7}{l},
  cell{10}{1} = {c=7}{l},
  hline{1,15} = {-}{0.06em,resRule},
  hline{2} = {2-7}{0.04em,resRule},
  hline{3,4,10,11} = {-}{0.03em,resRule},
}
Model
& Video-RAG & &
& EgoButler & & \\

& Ans-F1 & Acc. & MRR
& Ans-F1 & Acc. & MRR \\

Open-source models
& & & & & & \\

Qwen-3-VL 8B
& 75.0 & 41.8 & 63.8
& 44.5 & 38.8 & 61.0 \\

Qwen-3-VL 30B
& 56.6 & 45.5 & 65.7
& 44.2 & 39.1 & 61.8 \\

InternVL-3.5 8B
& 81.7 & 41.0 & 63.3
& 61.4 & 39.8 & 61.8 \\

InternVL-3.5 30B
& 77.7 & 42.3 & 63.7
& 28.5 & 27.3 & 53.4 \\

Gemma-4-E4B IT
& 40.3 & 35.3 & 58.2
& 30.9 & 36.4 & 58.2 \\

Gemma-4 31B
& 67.2 & 45.6 & 65.5
& 43.9 & 41.5 & 62.2 \\

Closed-source models
& & & & & & \\

Gemini-3-Flash
& \SetCell{bg=resBest}\textbf{83.9}
& \SetCell{bg=resBest}\textbf{61.0}
& \SetCell{bg=resBest}\textbf{76.0}
& 71.2
& \SetCell{bg=resBest}\textbf{54.1}
& \SetCell{bg=resBest}\textbf{71.6} \\

Gemini-3.1-Pro
& 67.4 & 53.2 & 70.7
& 43.5 & 42.6 & 64.2 \\

GPT-5.4-mini
& 77.6 & 47.8 & 67.4
& \SetCell{bg=resBest}\textbf{75.0}
& 46.0 & 66.1 \\

GPT-5.4
& 78.3 & 52.3 & 69.5
& 71.7 & 48.0 & 67.2 \\

\end{tblr}

\vspace{0.2ex}
\vspace{-3mm}
\end{table*}

\Cref{tab:supermemory_results} reports the performance of the baseline agentic systems. We evaluate each framework using the same set of open-source and closed-source vision-language models as underlying reasoning models. 
%
Overall, the results show that \datasetName challenges systems in both retrieving relevant evidence and deciding whether there is sufficient information to answer. Even the strongest configuration, Gemini-3-Flash with Video-RAG, reaches only 61.0\% QA-Acc, despite achieving 83.9\% Ans-F1 and 76.0\% QA-MRR. This gap indicates that detecting whether a question is answerable is only the first hurdle: models must also retrieve the precise multimodal evidence, interpret it correctly, distinguish the correct answer from plausible distractors, and abstain only when the evidence is insufficient. These results demonstrate that long-form, real-world memory assistance remains a significant open challenge.

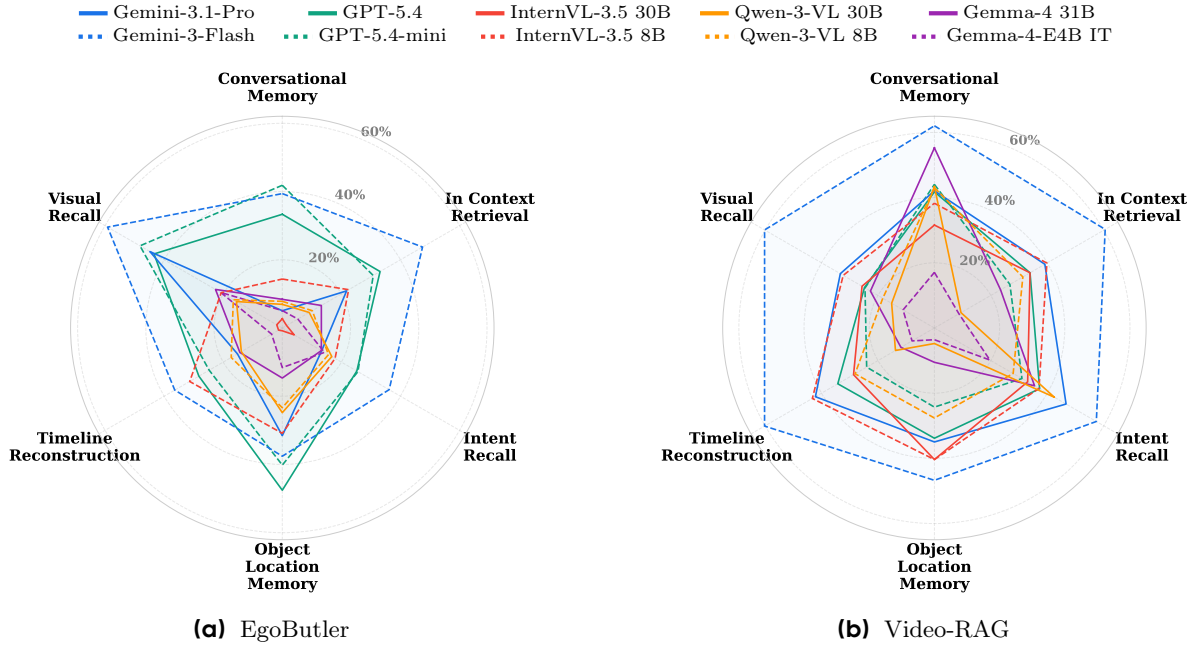
\begin{figure*}[t!]
    \centering

    {\scriptsize
    \setlength{\tabcolsep}{5pt}
    \renewcommand{\arraystretch}{1.1}
    \begin{tabular}{ccccc}
        \begin{tikzpicture}[baseline=-0.6ex]
            \draw[colorGemini, ultra thick] (0,0) -- (0.38,0);
        \end{tikzpicture}\,Gemini-3.1-Pro
        &
        \begin{tikzpicture}[baseline=-0.6ex]
            \draw[colorGPT, ultra thick] (0,0) -- (0.38,0);
        \end{tikzpicture}\,GPT-5.4
        &
        \begin{tikzpicture}[baseline=-0.6ex]
            \draw[colorInternVL, ultra thick] (0,0) -- (0.38,0);
        \end{tikzpicture}\,InternVL-3.5 30B
        &
        \begin{tikzpicture}[baseline=-0.6ex]
            \draw[colorQwen, ultra thick] (0,0) -- (0.38,0);
        \end{tikzpicture}\,Qwen-3-VL 30B
        &
        \begin{tikzpicture}[baseline=-0.6ex]
            \draw[colorGemma, ultra thick] (0,0) -- (0.38,0);
        \end{tikzpicture}\,Gemma-4 31B
        \\
        \begin{tikzpicture}[baseline=-0.6ex]
            \draw[colorGemini, ultra thick, dotted] (0,0) -- (0.38,0);
        \end{tikzpicture}\,Gemini-3-Flash
        &
        \begin{tikzpicture}[baseline=-0.6ex]
            \draw[colorGPT, ultra thick, dotted] (0,0) -- (0.38,0);
        \end{tikzpicture}\,GPT-5.4-mini
        &
        \begin{tikzpicture}[baseline=-0.6ex]
            \draw[colorInternVL, ultra thick, dotted] (0,0) -- (0.38,0);
        \end{tikzpicture}\,InternVL-3.5 8B
        &
        \begin{tikzpicture}[baseline=-0.6ex]
            \draw[colorQwen, ultra thick, dotted] (0,0) -- (0.38,0);
        \end{tikzpicture}\,Qwen-3-VL 8B
        &
        \begin{tikzpicture}[baseline=-0.6ex]
            \draw[colorGemma, ultra thick, dotted] (0,0) -- (0.38,0);
        \end{tikzpicture}\,Gemma-4-E4B IT
    \end{tabular}
    }

    \vspace{0.5em}

    \begin{subfigure}{0.48\textwidth}
        \centering
        \includegraphics[width=0.9\linewidth]{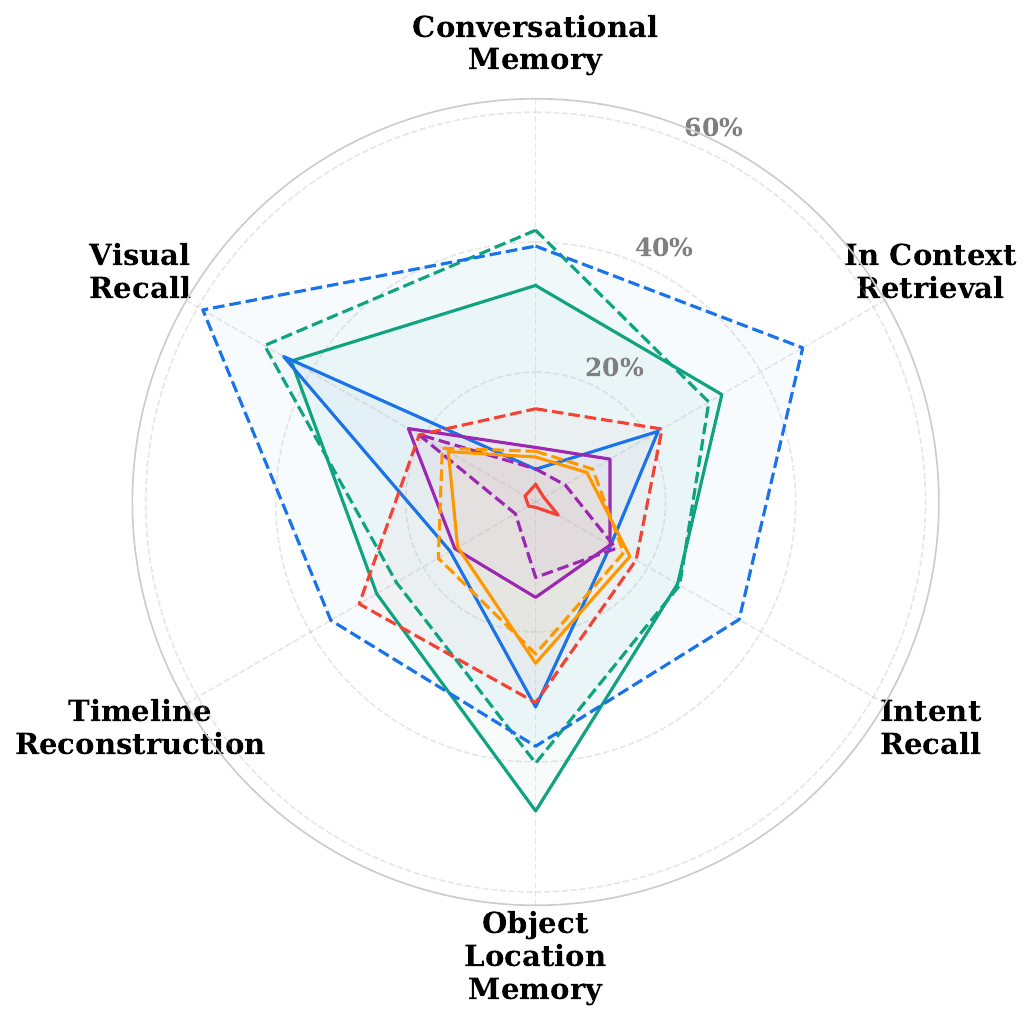}
        \caption{EgoButler}
        \label{fig:radar_egobutler}
    \end{subfigure}
    \hfill
    \begin{subfigure}{0.48\textwidth}
        \centering
        \includegraphics[width=0.9\linewidth]{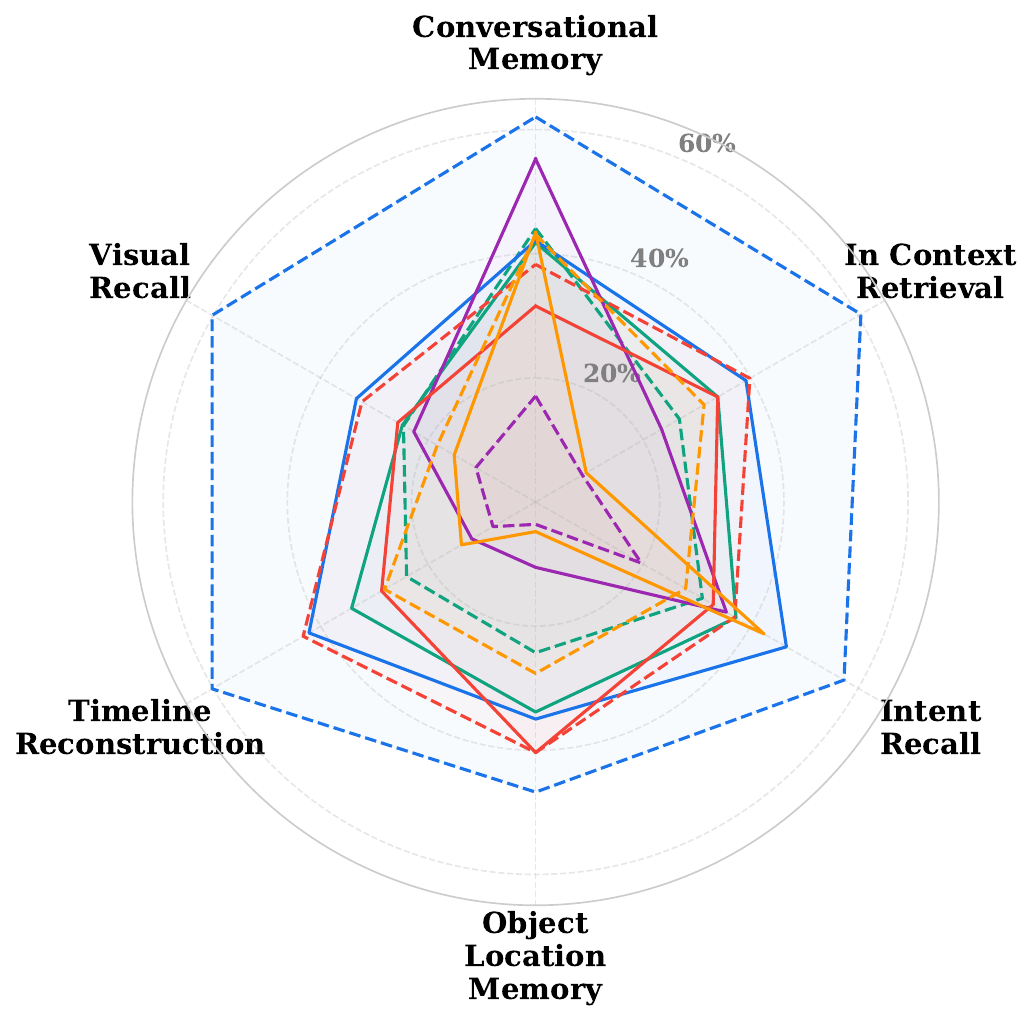}
        \caption{Video-RAG}
        \label{fig:radar_videorag}
    \end{subfigure}

    \caption{Task-category breakdowns across model families and agentic systems.}
    \label{fig:skill_radars}
    \vspace{-3mm}
\end{figure*}

Across models, Video-RAG outperforms EgoButler on most metrics. On average, it improves Ans-F1 from 51.5\% to 70.5\%, QA-Acc from 41.4\% to 46.6\%, and QA-MRR from 62.8\% to 66.4\%. The largest gain is in Ans-F1, indicating that retrieval-augmented evidence construction is especially useful for deciding whether a question is grounded in the recorded memory. Video-RAG achieves higher Ans-F1 for every model and higher QA-MRR for nine of ten models, with one tie. Although QA-Acc remains mixed, Video-RAG still obtains the best accuracy, reaching 61.0\% with Gemini-3-Flash.

\subsection{Detailed Analysis}
\label{sec:results.analysis}

\vspace{1mm}
\textbf{Closed-source models are stronger, but performance is not tied to model size.}
Closed-source models generally outperform open-source models under both frameworks. Under Video-RAG, they average 76.8\% Ans-F1, 53.6\% QA-Acc, and 70.9\% QA-MRR, compared with 66.4\%, 41.9\%, and 63.4\% for open-source models. However, performance is not monotonic with model size. Gemini-3-Flash performed the best across all Video-RAG metrics and also leads EgoButler in QA-Acc and QA-MRR, outperforming Gemini-3.1-Pro across all reported metrics. Flash appears better matched to retrieved evidence, more often committing when support is present, whereas Gemini-3.1-Pro is more conservative under noisy or partial long-horizon evidence; the gap persists with partial credit for vague choices (see \Cref{tab:vague_choice_accuracy_simple}), and hence Gemini-3.1-Pro's lower performance is not solely due to selecting more vague-but-related answers. Similarly, GPT-5.4-mini achieves the best EgoButler Ans-F1 at 75.0\%, exceeding GPT-5.4. These results suggest that long-horizon egocentric memory QA depends not only on base model scale but also on retrieval format, evidence quality, and deciding when to answer versus abstain.

Similarly, larger variants do not consistently improve performance in open-source models. Qwen-3-VL 30B underperforms Qwen-3-VL 8B on Video-RAG Ans-F1 by a wide margin (56.6\% vs 75.0\%), even though its QA-Acc and QA-MRR are slightly higher, indicating that the larger model abstains or hedges more often when the retrieved evidence is noisy. The contrast is sharper for InternVL-3.5: under EgoButler, the 30B variant collapses to 28.5\% Ans-F1 and 27.3\% QA-Acc, compared with 61.4\% and 39.8\% for the 8B variant, suggesting that the larger model fails to exploit EgoButler's caption-based memory format. Gemma exhibits the opposite trend under Video-RAG, where the 31B model improves substantially over the E4B IT variant on every metric (Ans-F1 40.3\% → 67.2\%, QA-Acc 35.3\% → 45.6\%, QA-MRR 58.2\% → 65.5\%).

\vspace{1mm}
\textbf{Task-level results highlight the value of structured retrieval.}
\Cref{fig:skill_radars} shows that EgoButler produces more irregular task profiles, with models struggling on in-context retrieval, conversational memory, intent recall, and timeline reconstruction. Video-RAG yields more balanced coverage across memory tasks, especially for tasks requiring evidence linked across time. This supports the need for structured retrieval over temporally distributed memory, rather than isolated frame-level reasoning.

\begin{figure}[t!]
    \centering
    \vspace{-4mm}

    \begin{minipage}[t]{0.52\linewidth}
        \centering
        \includegraphics[width=\linewidth]{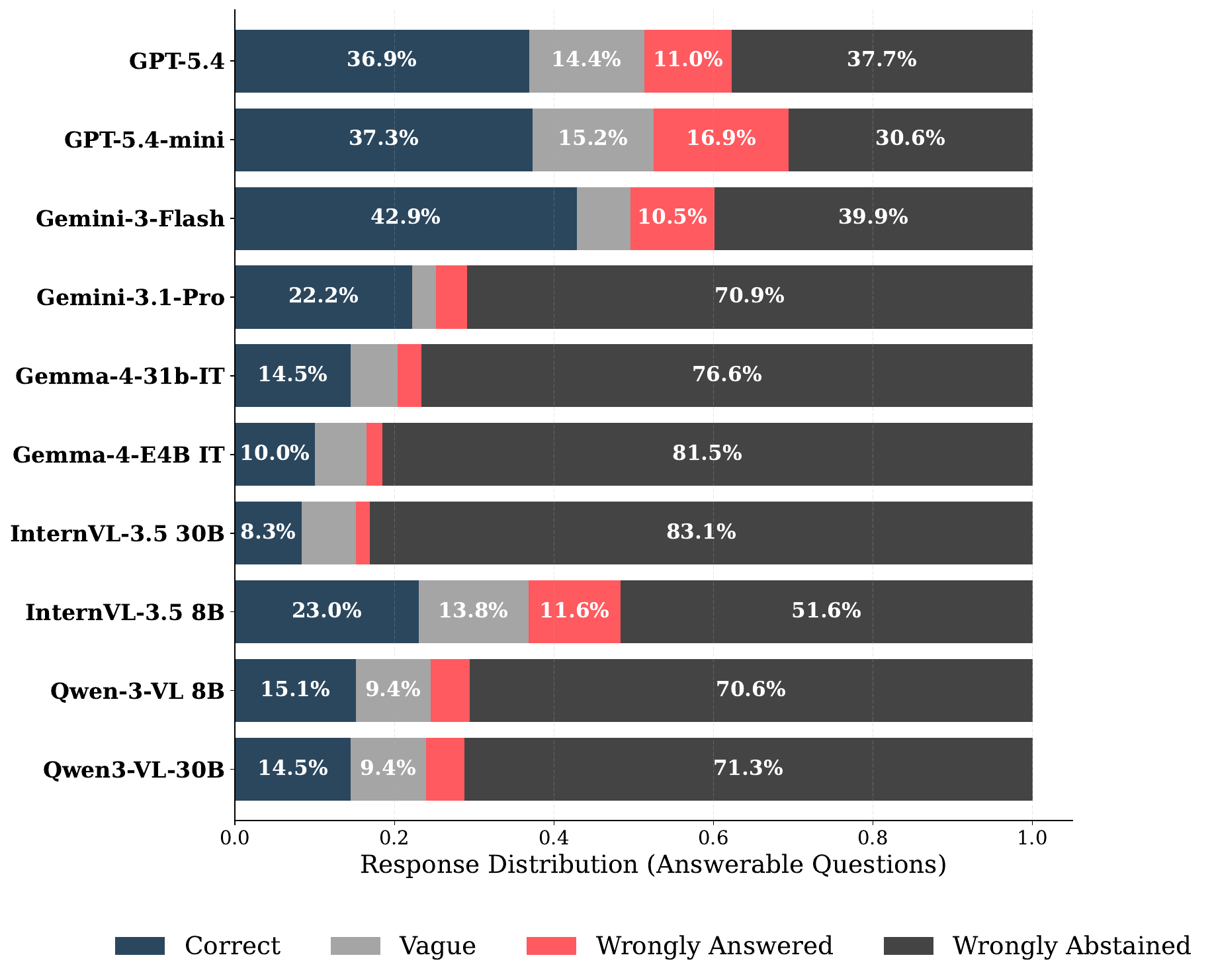}
        \captionof{figure}{Response reliability on answerable QAs.}
        \label{fig:reliability}
    \end{minipage}
    \hfill
    \begin{minipage}[t]{0.45\linewidth}
        \centering
        \includegraphics[width=\linewidth]{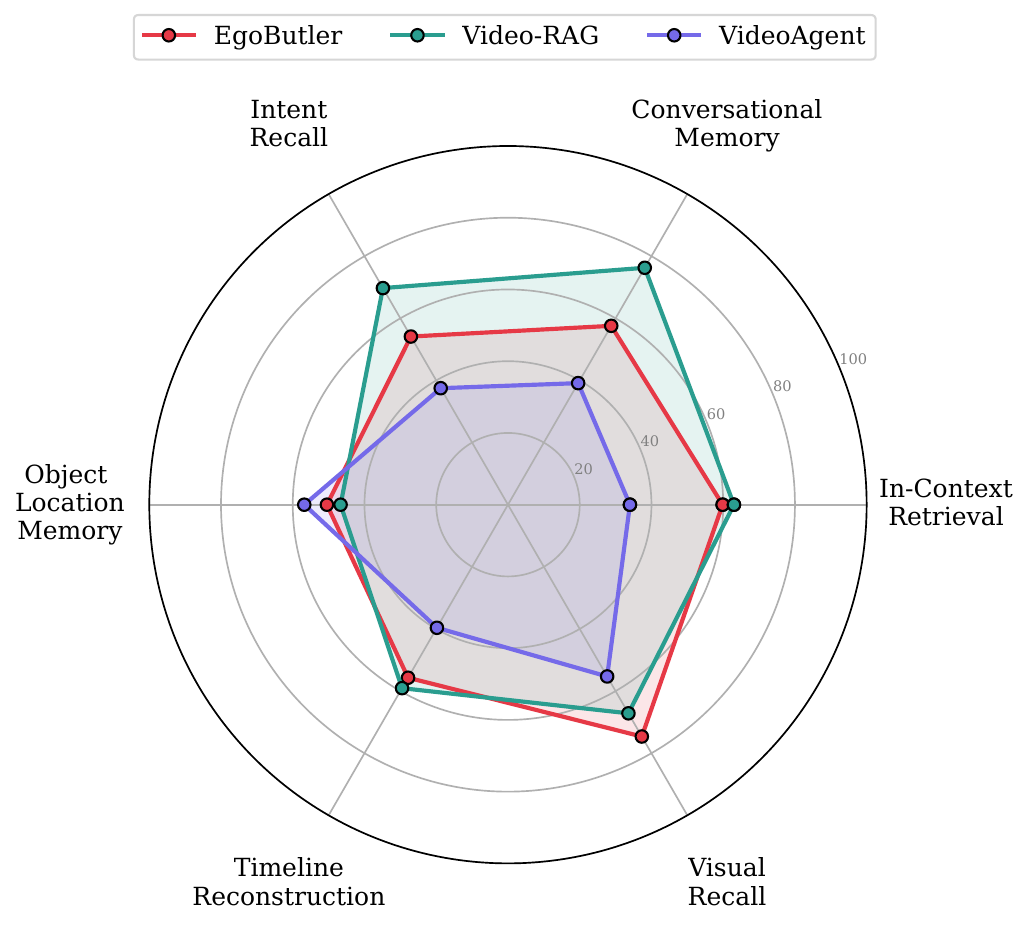}
        \captionof{figure}{Performance of EgoButler, Video-RAG, and VideoAgent using gemini-3-flash model across six \datasetName task categories.}
        \label{fig:radar_3_baseline_gemini}
    \end{minipage}

    \vspace{-4mm}
\end{figure}

\vspace{1mm}
\textbf{Reliability failures are dominated by excessive abstention.}
\Cref{fig:reliability} shows that a major failure mode on answerable questions is not choosing the wrong answer, but abstaining when evidence is present. Several open-source models wrongly abstain on more than 70\% of answerable questions. Even Gemini-3-Flash, the strongest model, answers correctly on only 42.9\% of answerable cases and wrongly abstains on 39.9\%. This behavior is more pronounced for Gemini-3.1-Pro, which has a 70.9\% abstention rate, with GPT-5.4 and GPT-5.4-mini showing similar traits. Thus, \datasetName effectively tests whether models can calibrate their predictions: they must avoid hallucinating when evidence is absent, while avoiding excessive abstention when support is available.

In summary, while Video-RAG improves stability and answerability detection, and Gemini-3-Flash provides the strongest baseline performance, the persistent gap between Ans-F1 and QA accuracy highlights a major opportunity for future architectures to better couple long-horizon retrieval with grounded reasoning. \Cref{sec:additional_results.qualitative} provides detailed qualitative case studies of these success and failure modes.

\subsubsection{Comparison with VideoAgent}
\label{sec:additional_results.compare-videoagent}
We additionally evaluate VideoAgent~\citep{fan2024videoagent} on \datasetName. Due to its iterative agentic loop, VideoAgent consumes substantially more tokens and computation than EgoButler and Video-RAG, while yielding less competitive results. We therefore report VideoAgent using Gemini-3-Flash only, which offers a favorable cost-performance trade-off for this comparison.
As shown in \Cref{fig:radar_3_baseline_gemini}, VideoAgent underperforms both EgoButler and Video-RAG across most dimensions, with the largest gaps on Conversational Memory, Intent Recall, and Timeline Reconstruction. The three methods perform comparably on Object Location Memory, indicating that this dimension is less sensitive to the choice of retrieval or agentic strategy. Overall, the added computational cost of VideoAgent does not translate into improved performance on \datasetName, motivating its exclusion from the main comparison.

\subsubsection{Blind Text-Only LLM Evaluation}
\label{sec:blind}

\noindent
A common concern in multiple-choice video QA benchmarks is \emph{information leakage}: questions may be answerable from linguistic priors alone, without genuine grounding in visual, auditory, or temporal evidence. Surface cues in question phrasing, implausible distractors, or world knowledge encoded in large language models can yield non-trivial accuracy and inflate apparent benchmark difficulty. To quantify this risk, we conduct a \emph{blind} evaluation where a strong open-weight LLM receives only the question text and answer options.

\noindent
We evaluate Qwen3-8B~\citep{yang2025qwen3} in a text-only configuration. Each prompt contains a brief instruction, the question stem, and four options labeled A--D; no frames, audio, transcripts, or metadata are provided. We report the overall accuracy and accuracy by task category.
\Cref{tab:blind} reports results on Person~1 ($N=1{,}017$ questions). Overall accuracy is $23.8\%$, slightly below the $25\%$ four-way chance baseline. Per-category accuracy ranges from $20.0\%$ (\texttt{object\_location\_memory}) to $27.0\%$ (\texttt{intent\_recall}). No category is meaningfully or statistically above chance, indicating that information leakage is minimal and that answer choices are well balanced across options.

\begin{table}[t]
\centering
\small
\caption{Blind text-only evaluation with Qwen3-8B on Person~1 ($N=1{,}017$ questions). The model receives only question text and options; no visual, audio, or temporal evidence is provided. Chance is $25\%$.}
\label{tab:blind}
\begin{tabular}{lccc}
\toprule
\textbf{Category} & \textbf{Correct} & \textbf{Total} & \textbf{Accuracy (\%)} \\
\midrule
\rowcolor{gray!20}\multicolumn{4}{l}{\textit{Overall}} \\
All questions & 242 & 1{,}017 & 23.8 \\
\midrule
\rowcolor{gray!20} \multicolumn{4}{l}{\textit{By task category}} \\
conversational\_memory   & 47 & 182 & 25.8 \\
in\_context\_retrieval   & 40 & 166 & 24.1 \\
intent\_recall           & 40 & 148 & 27.0 \\
object\_location\_memory & 41 & 205 & 20.0 \\
timeline\_reconstruction & 30 & 141 & 21.3 \\
visual\_recall           & 44 & 175 & 25.1 \\
\bottomrule
\end{tabular}
\end{table}

\noindent
The blind baseline confirms two points. First, our QA set exhibits minimal information leakage: a capable 8B model performs at chance and well below multimodal systems with full egocentric input (\Cref{tab:supermemory_results}), confirming that answer choices are balanced and that questions cannot be solved through linguistic priors alone. Second, this baseline establishes a floor against which gains from visual, auditory, and temporal evidence can be attributed to genuine multimodal grounding rather than dataset artifacts.

\subsubsection{Per-video and per-person Evaluation}
\label{sec:per-video-person}
\Cref{tab:model_accuracy_per_video_person} reports model accuracy on the \datasetName benchmark across two evaluation granularities. \textbf{Per-Video} accuracy is computed per video clip: for each video, we calculate the mean QA accuracy over all questions associated with that clip, then report the mean and standard deviation of these per-video scores across all videos in the benchmark.
\textbf{Per-Person} accuracy aggregates at the participant level: for each person, we compute the mean QA accuracy over all questions attributed to that individual, then report the mean and standard deviation across all participants. Both metrics are reported for two retrieval baselines, Video-RAG and EgoButler, spanning open-source and closed-source vision-language models.

\begin{table}[t]
\centering
\caption{Model accuracy on \datasetName across two evaluation granularities.
}
\vspace{-1mm}
\label{tab:model_accuracy_per_video_person}
\small
\begin{tblr}{
  width = \textwidth,
  colspec = {
    X[1.55,l]
    X[1.0,c]
    X[1.0,c]
  },
  rows = {
    fg=resText,
    valign=m,
    abovesep=0.7pt,
    belowsep=0.7pt
  },
  row{1} = {
    bg=resHeader,
    fg=resHeaderFg,
    font=\bfseries,
    halign=c,
    abovesep=0.8pt,
    belowsep=0.8pt
  },
  row{2} = {
    bg=resSubHeader,
    font=\bfseries,
    halign=c,
    abovesep=0.55pt,
    belowsep=0.55pt
  },
  row{3} = {
    bg=resOpenGroup,
    font=\bfseries,
    abovesep=0.55pt,
    belowsep=0.55pt
  },
  row{4-8} = {bg=resOpenRow},
  row{9} = {
    bg=resClosedGroup,
    font=\bfseries,
    abovesep=0.55pt,
    belowsep=0.55pt
  },
  row{10-13} = {bg=resClosedRow},
  column{1} = {cmd=\RaggedRight},
  cell{1}{2} = {c=2}{bg=resHeader,fg=resHeaderFg,font=\bfseries},
  cell{1}{4} = {c=2}{bg=resHeader,fg=resHeaderFg,font=\bfseries},
  cell{3}{1} = {c=5}{l},
  cell{9}{1} = {c=5}{l},
  hline{1,14} = {-}{0.06em,resRule},
  hline{2} = {2-5}{0.04em,resRule},
  hline{3,4,9,10} = {-}{0.03em,resRule},
}
Model
& Per-Video &
& Per-Person & \\

& Video-RAG & EgoButler
& Video-RAG & EgoButler \\

Open-source models
& & & & \\

Qwen-3-VL 8B
& $42.4_{\pm15.3}$ & $40.6_{\pm16.9}$
& $42.9_{\pm5.4}$ & $40.7_{\pm6.2}$ \\

Qwen-3-VL 30B
& $45.5_{\pm17.6}$ & $40.5_{\pm17.5}$
& $47.2_{\pm8.4}$ & $41.1_{\pm6.7}$ \\

InternVL-3.5 8B
& $40.2_{\pm12.4}$ & $38.6_{\pm14.1}$
& $41.7_{\pm3.5}$ & $41.2_{\pm5.8}$ \\

Gemma-4-E4B IT
& $34.9_{\pm15.1}$ & $36.7_{\pm16.6}$
& $36.8_{\pm3.7}$ & $38.1_{\pm5.2}$ \\

Gemma-4 31B
& $48.6_{\pm12.3}$ & $43.5_{\pm18.2}$
& $49.0_{\pm5.2}$ & $45.2_{\pm7.9}$ \\

Closed-source models
& & & & \\

Gemini-3-Flash
& \SetCell{bg=resBest}$\mathbf{61.3}_{\pm15.7}$
& \SetCell{bg=resBest}$\mathbf{55.3}_{\pm15.1}$
& \SetCell{bg=resBest}$\mathbf{62.1}_{\pm6.3}$
& \SetCell{bg=resBest}$\mathbf{56.5}_{\pm7.3}$ \\

Gemini-3.1-Pro
& $52.3_{\pm17.3}$ & $43.4_{\pm17.2}$
& $54.9_{\pm8.8}$ & $44.6_{\pm6.9}$ \\

GPT-5.4-mini
& $49.7_{\pm19.8}$ & $45.9_{\pm14.5}$
& $47.4_{\pm5.8}$ & $47.4_{\pm5.7}$ \\

GPT-5.4
& $52.3_{\pm19.5}$ & $49.0_{\pm16.7}$
& $52.8_{\pm7.9}$ & $50.0_{\pm7.6}$ \\

\end{tblr}

\end{table}

\Cref{tab:vague_choice_accuracy_simple} reports a complementary accuracy variant that assigns partial credit to vague choices. This diagnostic tests whether model differences are driven mainly by choosing vague but related answers instead of fully correct answers. Gemini-3-Flash remains ahead of Gemini-3.1-Pro under both frameworks, indicating that the closed-source scaling gap in the main results is not explained solely by Gemini-3.1-Pro selecting more vague answers.

\begin{table}[t]
\centering
\caption{QA average accuracy over all videos when assigning vague choices partial credit of 0.5, along with the absolute change ($\Delta$) from standard QA accuracy where vague choices receive zero credit.}
\vspace{-1mm}
\label{tab:vague_choice_accuracy_simple}
\small
\begin{tblr}{
  width = \textwidth,
  colspec = {
    X[1.45,l]
    X[0.9,c]
    X[0.9,c]
    X[0.9,c]
    X[0.9,c]
  },
  rows = {
    fg=resText,
    valign=m,
    abovesep=0.7pt,
    belowsep=0.7pt
  },
  row{1} = {
    bg=resHeader,
    fg=resHeaderFg,
    font=\bfseries,
    halign=c,
    abovesep=0.8pt,
    belowsep=0.8pt
  },
  row{2} = {
    bg=resOpenGroup,
    font=\bfseries,
    abovesep=0.55pt,
    belowsep=0.55pt
  },
  row{3-7} = {bg=resOpenRow},
  row{8} = {
    bg=resClosedGroup,
    font=\bfseries,
    abovesep=0.55pt,
    belowsep=0.55pt
  },
  row{9-12} = {bg=resClosedRow},
  column{1} = {cmd=\RaggedRight},
  cell{2}{1} = {c=3}{l},
  cell{8}{1} = {c=3}{l},
  hline{1,13} = {-}{0.06em,resRule},
  hline{2,3,8,9} = {-}{0.03em,resRule},
}
Model & Video-RAG & EgoButler \\

Open-source models
& & \\

Qwen-3-VL 8B
& 50.8 {\scriptsize\color{txtGreen}[$\uparrow$9.0\%]}
& 42.1 {\scriptsize\color{txtGreen}[$\uparrow$3.3\%]} \\
Qwen-3-VL 30B
& 49.6 {\scriptsize\color{txtGreen}[$\uparrow$4.1\%]}
& 42.3 {\scriptsize\color{txtGreen}[$\uparrow$3.2\%]} \\
InternVL-3.5 8B
& 49.1 {\scriptsize\color{txtGreen}[$\uparrow$8.1\%]}
& 44.6 {\scriptsize\color{txtGreen}[$\uparrow$4.8\%]} \\
Gemma-4-E4B IT
& 38.8 {\scriptsize\color{txtGreen}[$\uparrow$3.4\%]}
& 38.7 {\scriptsize\color{txtGreen}[$\uparrow$2.2\%]} \\
Gemma-4-31b
& 54.4 {\scriptsize\color{txtGreen}[$\uparrow$2.9\%]}
& 44.2 {\scriptsize\color{txtGreen}[$\uparrow$1.5\%]} \\
Closed-source models
& & \\
Gemini-3-Flash
& \SetCell{bg=resBest}\textbf{64.4} {\scriptsize\color{txtGreen}[$\uparrow$3.5\%]}
& \SetCell{bg=resBest}\textbf{56.4} {\scriptsize\color{txtGreen}[$\uparrow$2.3\%]} \\
Gemini-3.1-Pro
& 55.2 {\scriptsize\color{txtGreen}[$\uparrow$2.0\%]}
& 43.6 {\scriptsize\color{txtGreen}[$\uparrow$1.1\%]} \\
GPT-5.4-mini
& 55.2 {\scriptsize\color{txtGreen}[$\uparrow$7.3\%]}
& 51.4 {\scriptsize\color{txtGreen}[$\uparrow$5.4\%]} \\
GPT-5.4
& 58.3 {\scriptsize\color{txtGreen}[$\uparrow$6.1\%]}
& 53.1 {\scriptsize\color{txtGreen}[$\uparrow$5.0\%]} \\

\end{tblr}

\end{table}

\subsubsection{Qualitative Results}
\label{sec:additional_results.qualitative}

\setcounter{topnumber}{3}
\setcounter{totalnumber}{4}
\renewcommand{\topfraction}{0.95}
\renewcommand{\textfraction}{0.05}
\renewcommand{\floatpagefraction}{0.8}
\setlength{\floatsep}{6pt plus 2pt minus 2pt}
\setlength{\textfloatsep}{8pt plus 2pt minus 2pt}

\newtcolorbox{qualcard}[2][]{
    enhanced,
    colback=schemaBg,
    colframe=schemaFrame,
    colbacktitle=tblHeader,
    coltitle=schemaText,
    fonttitle=\bfseries\scriptsize,
    title={#2},
    boxrule=0.35pt,
    arc=1mm,
    left=0.65mm,
    right=0.65mm,
    top=0.55mm,
    bottom=0.55mm,
    toptitle=0.35mm,
    bottomtitle=0.35mm,
    boxsep=0pt,
    #1
}

\newenvironment{qualcardpair}{%
    \begin{minipage}[t]{0.492\linewidth}\vspace{0pt}%
    \renewcommand{\textbf}[1]{{\bfseries\sffamily##1}}%
}{%
    \end{minipage}%
}

\newcommand{\qualframe}[3]{%
    \begin{minipage}[t]{#1}
        \centering
        \includegraphics[width=\linewidth]{#2}
        \par\vspace{0.25mm}
        {\tiny\color{tblText}#3}
    \end{minipage}%
}

\newcommand{\qualchip}[4]{%
    {\setlength{\fboxsep}{1.2pt}%
    \fcolorbox{#2}{#1}{%
        \begin{minipage}[t]{0.265\linewidth}
            \scriptsize\textbf{#3}\par #4
        \end{minipage}%
    }}%
}

\newcommand{\qualresultrow}[3]{%
    \par\vspace{0.7mm}
    \noindent
    \qualchip{qaNABg}{qaNABorder}{Truth}{#1}\hfill
    \qualchip{qaCorrectBg}{qaCorrectBorder}{Video-RAG}{#2}\hfill
    \qualchip{qaWrongBg}{qaWrongBorder}{EgoButler}{#3}
}

\newcommand{\qualanswers}[2]{%
    \par\vspace{0.35mm}
    \begin{tcolorbox}[
        enhanced,
        colback=white,
        colframe=schemaFrame,
        boxrule=0.2pt,
        arc=0.8mm,
        left=0.45mm,
        right=0.45mm,
        top=0.35mm,
        bottom=0.35mm,
        boxsep=0pt
    ]
        {\small\textbf{Answer choices}\par #1}
        \vspace{0.5mm}
        {\tiny #2}
    \end{tcolorbox}
}

\newcommand{\qualchoice}[2]{%
    \hangindent=4.9em\hangafter=1\noindent{\tiny #1~#2\par}%
}

\newcommand{\qualselect}[3]{%
    \resizebox{\linewidth}{!}{\textbf{GT:}~#1\quad\textbf{Video-RAG:}~#2\quad\textbf{EgoButler:}~#3}%
}

\newcommand{\qualmodelselect}[3]{%
    \resizebox{\linewidth}{!}{\textbf{GT:}~#1\quad\textbf{Gemini-3-Flash:}~#2\quad\textbf{Gemini-3.1-Pro:}~#3}%
}

To complement the aggregate benchmark results, we inspect a small set of curated examples drawn from the released QA annotations. These examples are intended to explain the concrete reasoning operations that make long-horizon egocentric memory difficult. The main pattern is that failures rarely arise from a single missing capability. Successful answers require the system to retrieve the right moment, preserve the relevant detail through summarization, compare it against the current query, and decide whether the available evidence is sufficient. A model can succeed at one stage and still fail downstream if it loses precision, over-compresses the memory trace, or treats plausible context as evidence.

The inspected examples fall into six recurring qualitative patterns. The figures show the concrete question text, answer choices, model selections, and evidence frames; the discussion below focuses on what each pattern reveals about the benchmark rather than repeating the figure content.

\begin{figure}[t!]
    \centering
    \begin{qualcardpair}
    \begin{qualcard}{Pinpoint dialogue retrieval}
        \qualframe{0.315\linewidth}{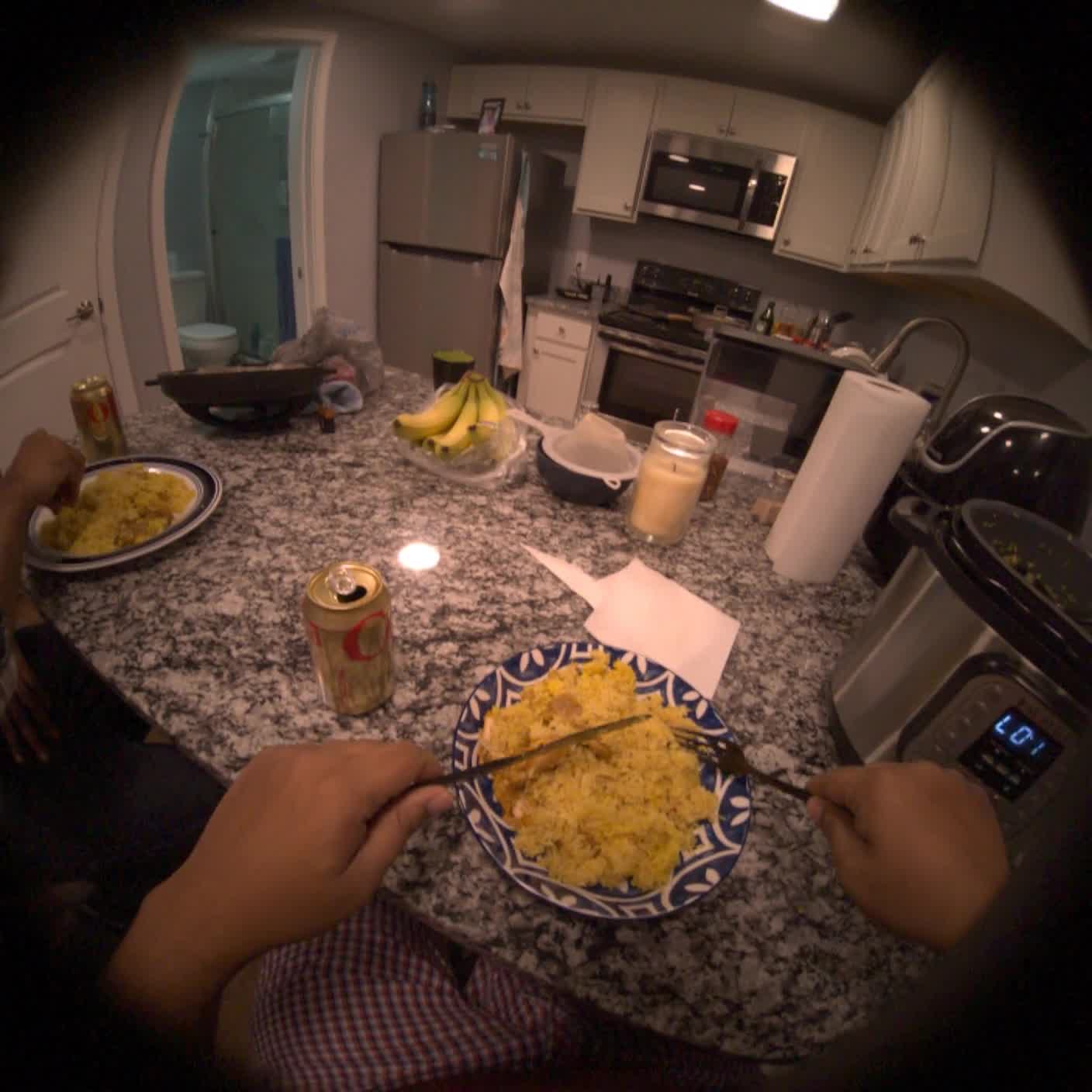}{E1}\hfill
        \qualframe{0.315\linewidth}{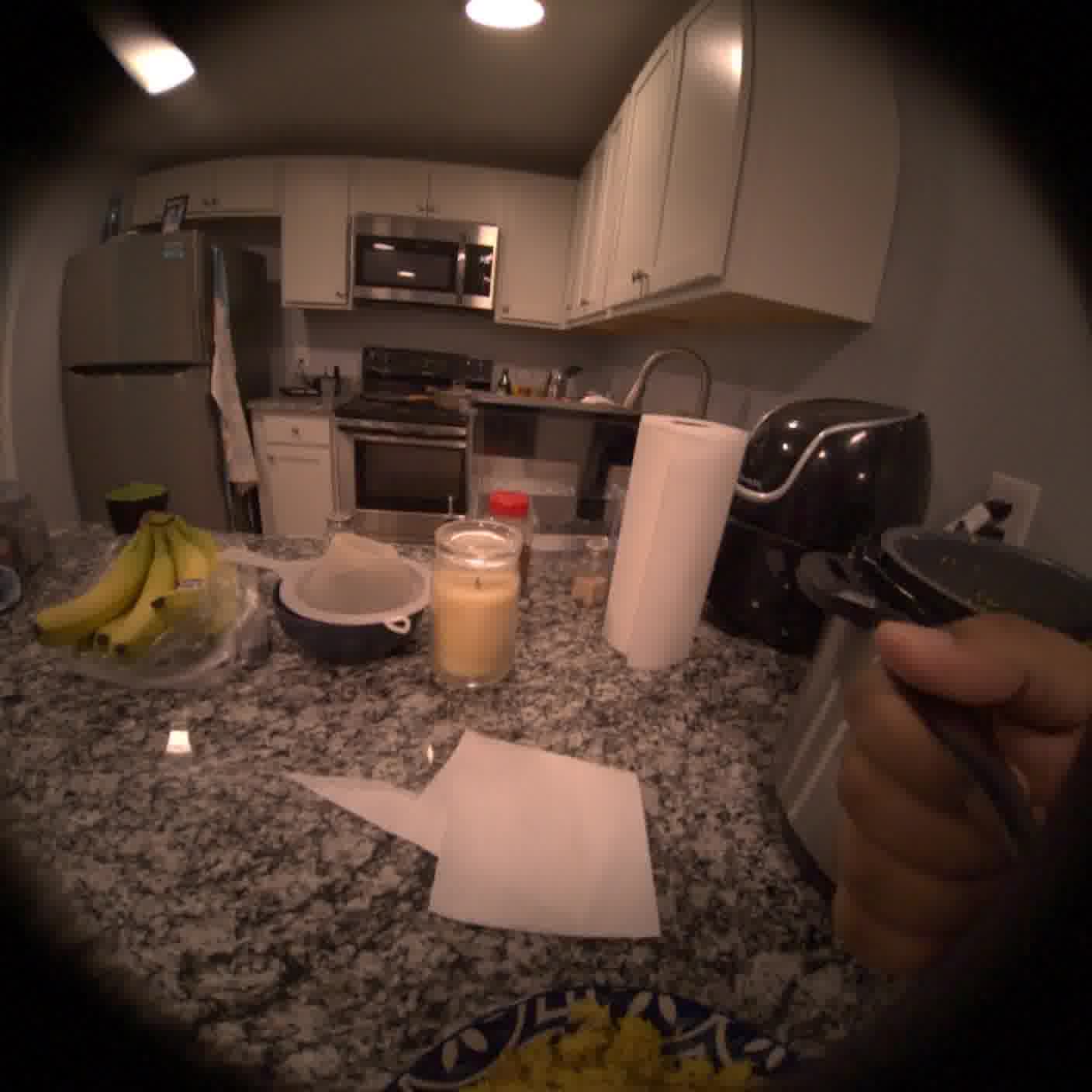}{E2}\hfill
        \qualframe{0.315\linewidth}{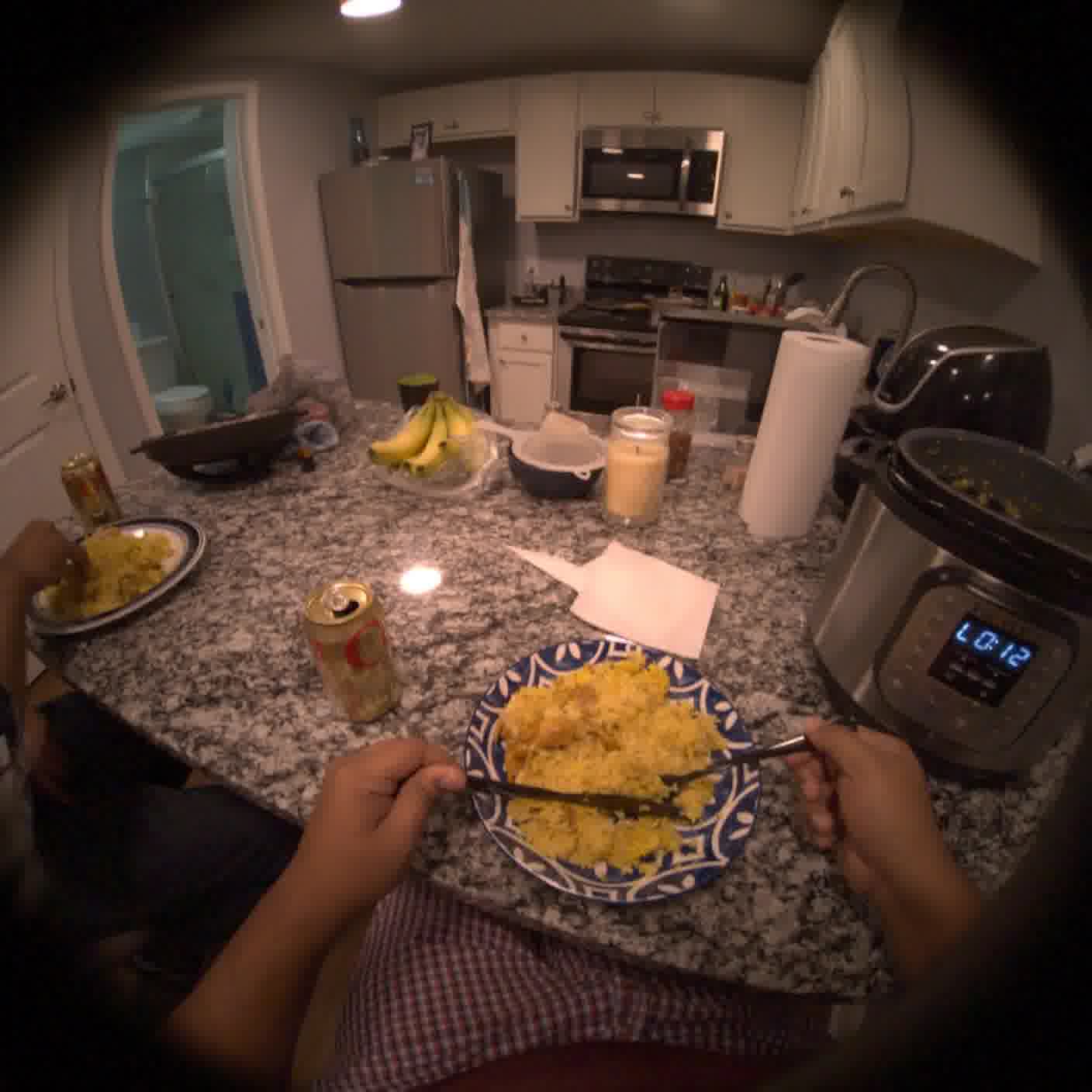}{E3}
        \vspace{0.35mm}

        {\tiny\textbf{Q:} I'm thinking about cooking meat. What did B say he uses his Instant Pot to cook?}
        \qualanswers{
            \qualchoice{\CorrectTag}{He said he uses it to cook beef.}
            \qualchoice{\VagueTag}{He said he uses it to cook meat.}
            \qualchoice{\WrongTag}{He said he uses it to cook chicken.}
            \qualchoice{\NATag}{This question cannot be answered.}
        }{\qualselect{\CorrectTag}{\CorrectTag}{\WrongTag}}
    \end{qualcard}
    \end{qualcardpair}\hfill
    \begin{qualcardpair}

    \begin{qualcard}{Board-game event retrieval}
        \qualframe{0.315\linewidth}{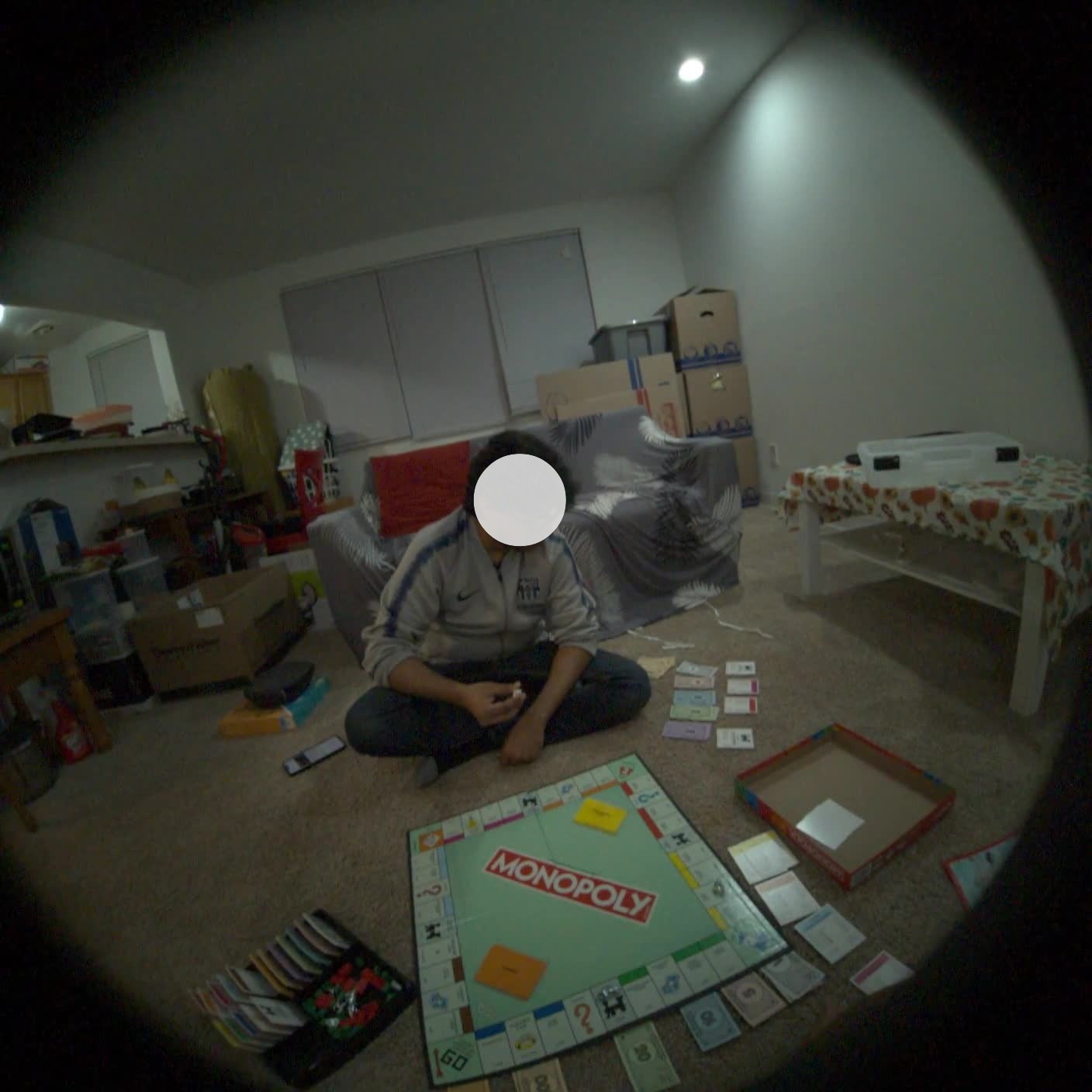}{e1}\hfill
        \qualframe{0.315\linewidth}{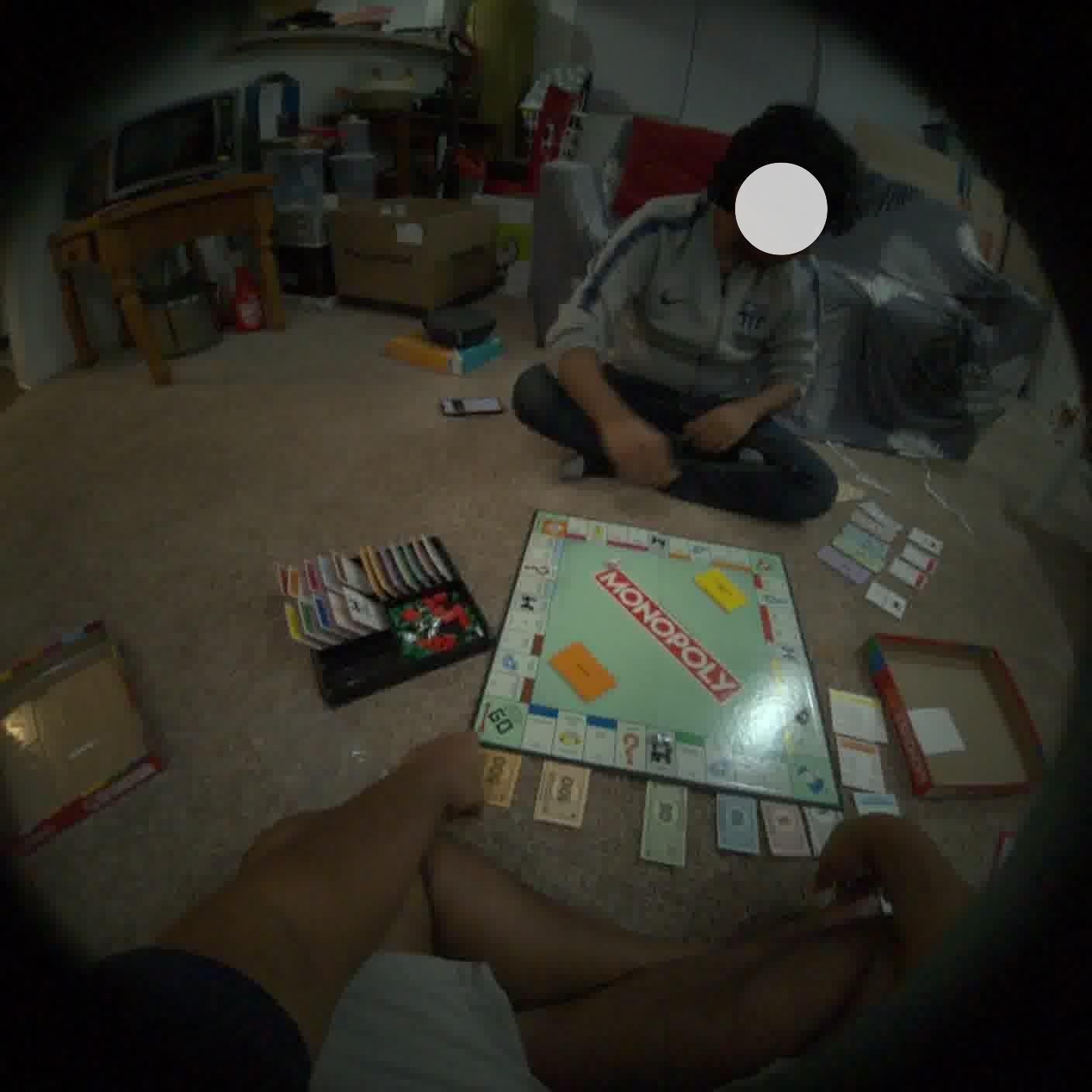}{e2}\hfill
        \qualframe{0.315\linewidth}{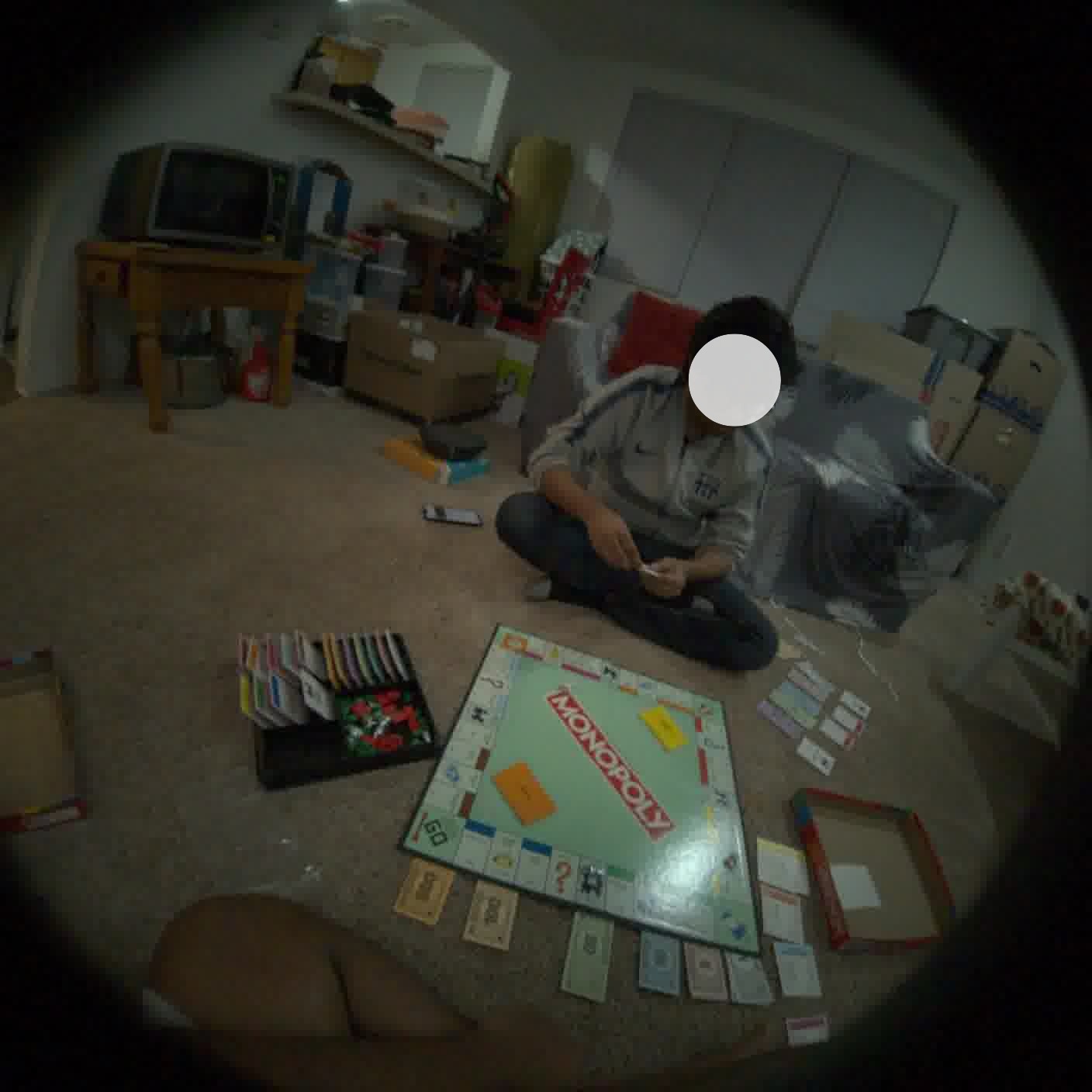}{e3}
        \vspace{0.35mm}

        {\tiny\textbf{Q:} I was just thinking about our Monopoly game the other day. After B landed on my St. Charles Place, what was the next property he ended up buying, and how much did he pay?}
        \qualanswers{
            \qualchoice{\CorrectTag}{He bought Water Works for \$150.}
            \qualchoice{\VagueTag}{He paid \$150.}
            \qualchoice{\WrongTag}{He bought Short Line for \$200.}
            \qualchoice{\NATag}{This question cannot be answered.}
        }{\qualselect{\CorrectTag}{\CorrectTag}{\NATag}}
    \end{qualcard}
    \end{qualcardpair}
    \caption{Retrieval cases where brief evidence is sufficient once the relevant moment is found.}
    \label{fig:qualitative_retrieval_cases}
\end{figure}

\textbf{Pinpoint retrieval of sparse evidence.}
When the decisive evidence is a short recoverable moment, retrieval can answer precisely. \Cref{fig:qualitative_retrieval_cases} shows two cases where the needed information is localized to a few frames or a brief conversational exchange. These examples clarify why retrieval quality matters even when the final reasoning step is simple: once the right evidence is surfaced, the answer is almost direct. The contrast between Video-RAG and EgoButler also suggests that hierarchical summaries can sometimes blur or omit isolated details that are not globally salient, while direct retrieval over lower-level traces can preserve them.

At the same time, these are the easiest qualitative successes: the evidence is sparse but not conceptually complex. They therefore provide a useful positive control for the benchmark. A system that cannot solve these cases is failing at basic memory lookup, whereas a system that can solve them may still struggle on cases that require disambiguation, counting, temporal ordering, or premise checking.

\begin{figure}[t!]
    \centering
    \begin{qualcardpair}
    \begin{qualcard}{Small-object tracking failure}
        \qualframe{0.315\linewidth}{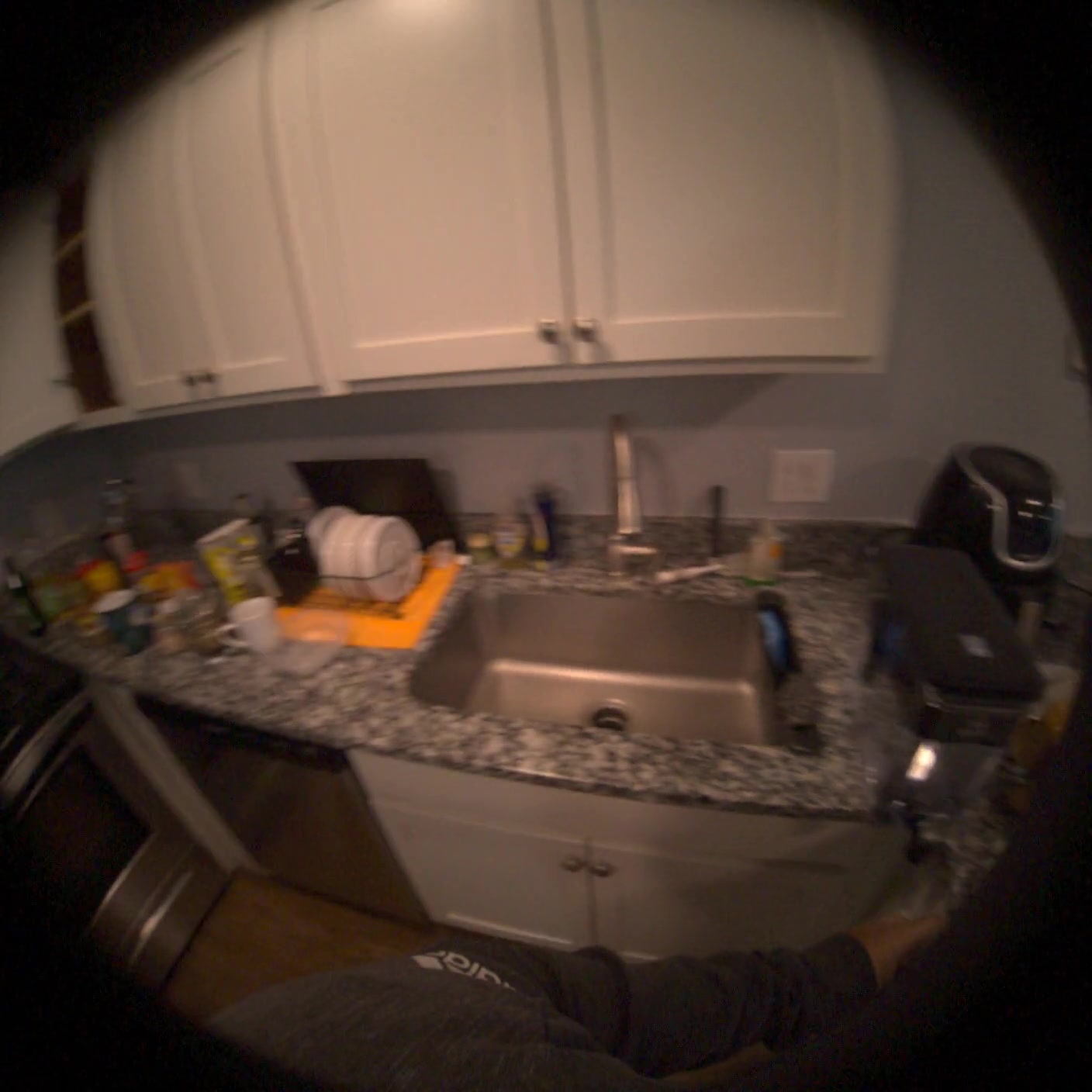}{e1}\hfill
        \qualframe{0.315\linewidth}{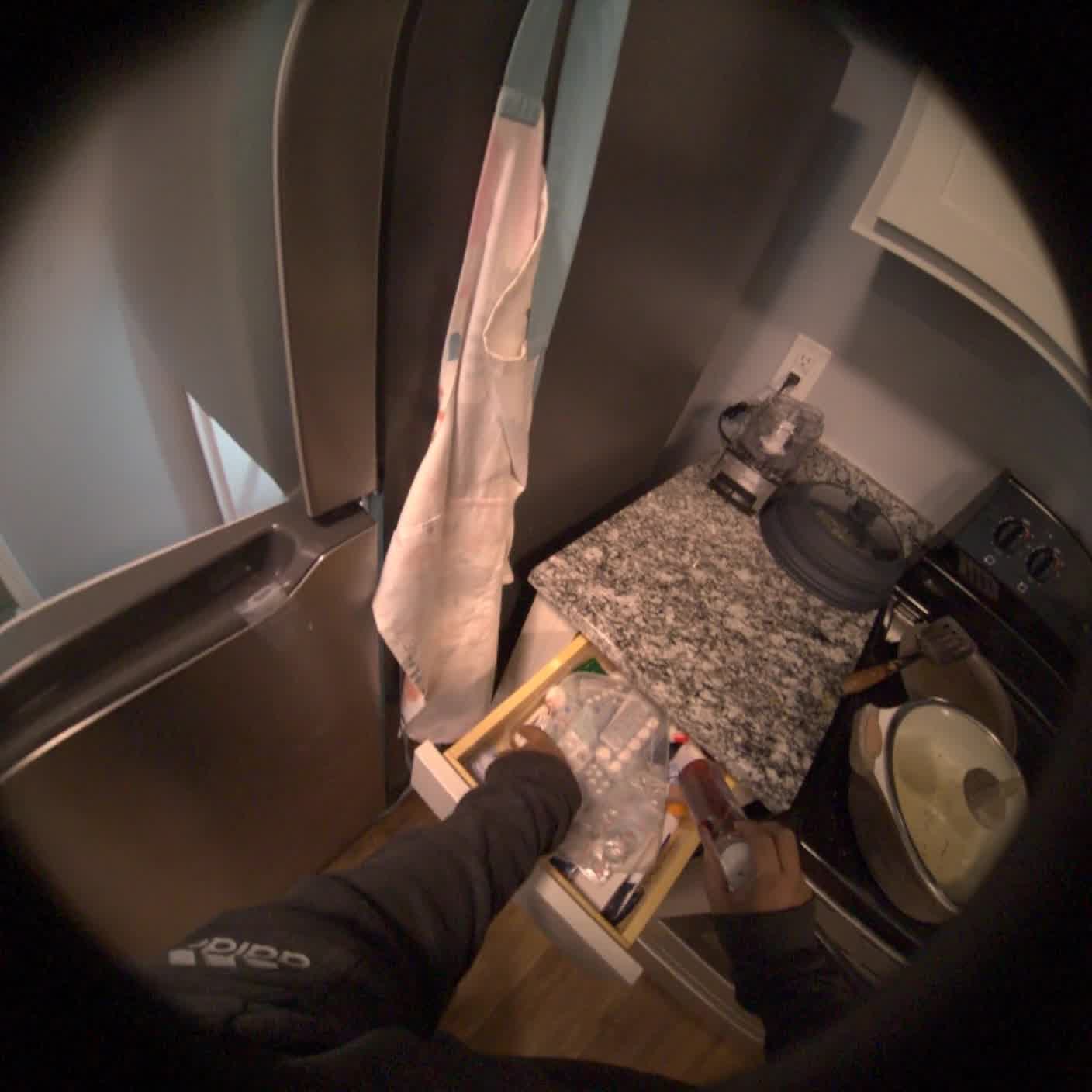}{e2}\hfill
        \qualframe{0.315\linewidth}{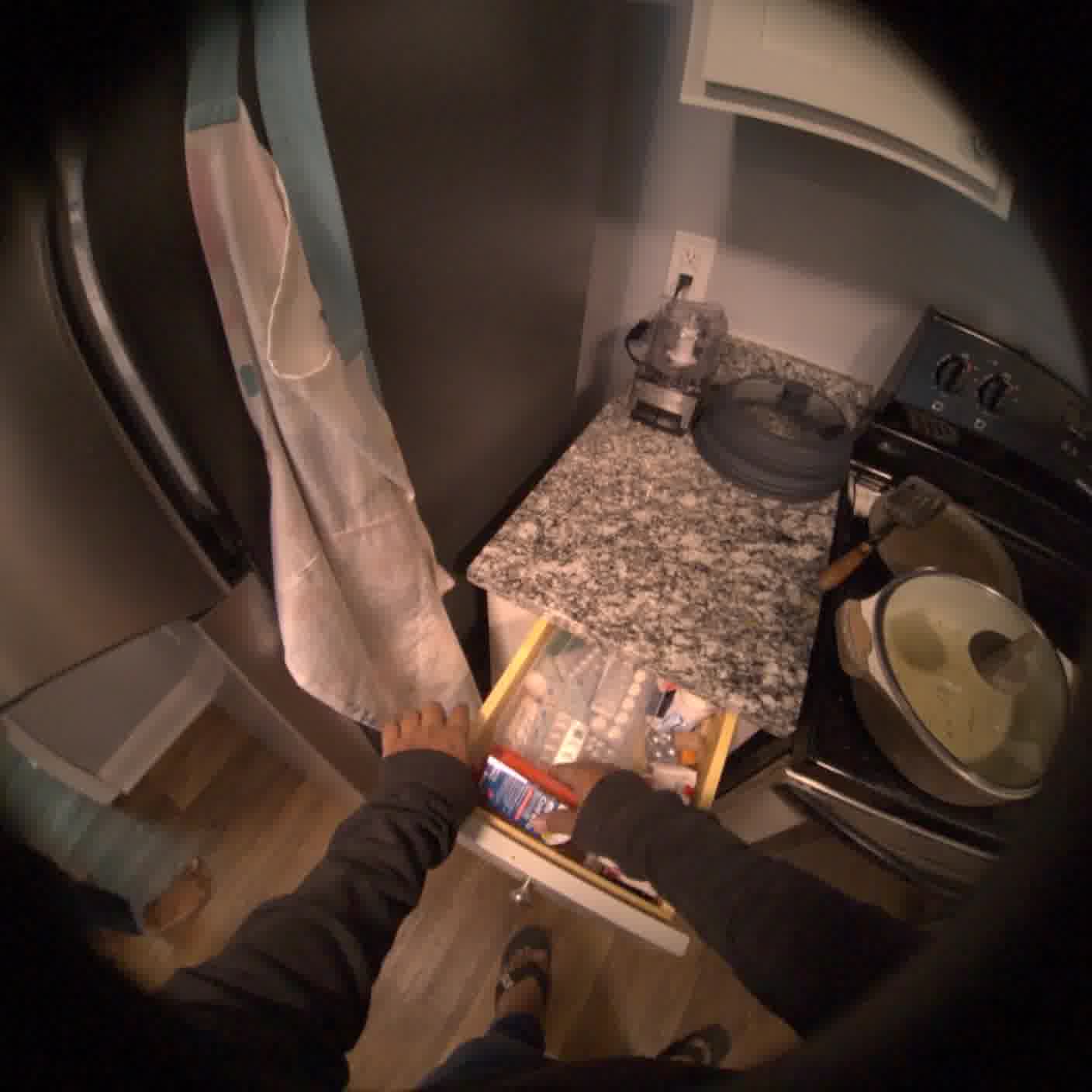}{e3}
        \vspace{0.35mm}

        {\tiny\textbf{Q:} I need to take my medicine before I start cooking. Where did I leave that small white bottle?}
        \qualanswers{
            \qualchoice{\CorrectTag}{You left it in the kitchen drawer next to the stove.}
            \qualchoice{\VagueTag}{You left it in one of the drawers in the kitchen area.}
            \qualchoice{\WrongTag}{You left it in the lower cabinet under the kitchen island.}
            \qualchoice{\NATag}{This question cannot be answered.}
        }{\qualselect{\CorrectTag}{\NATag}{\NATag}}
    \end{qualcard}
    \end{qualcardpair}\hfill
    \begin{qualcardpair}

    \begin{qualcard}{Numerical OCR failure}
        \qualframe{0.315\linewidth}{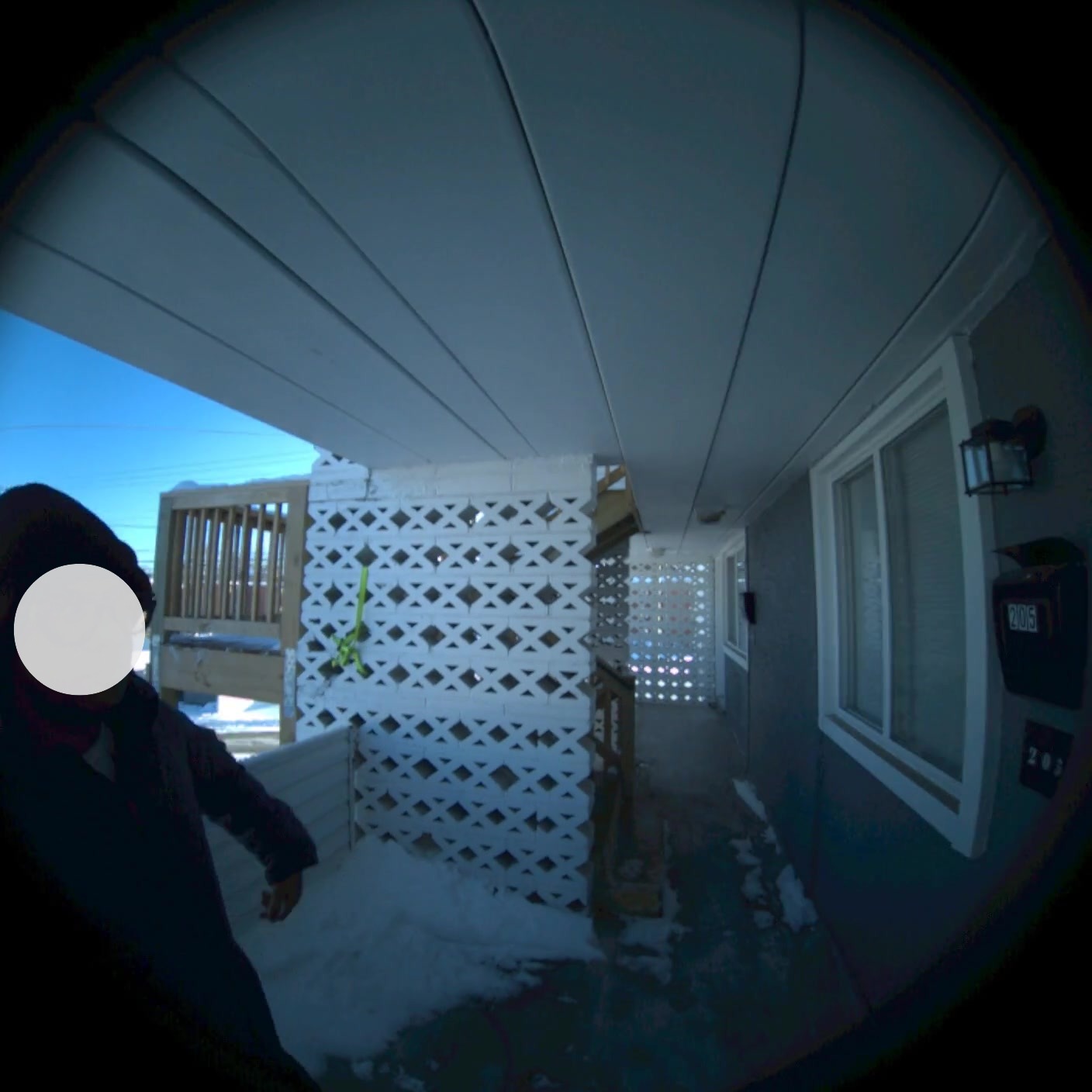}{e1}\hfill
        \qualframe{0.315\linewidth}{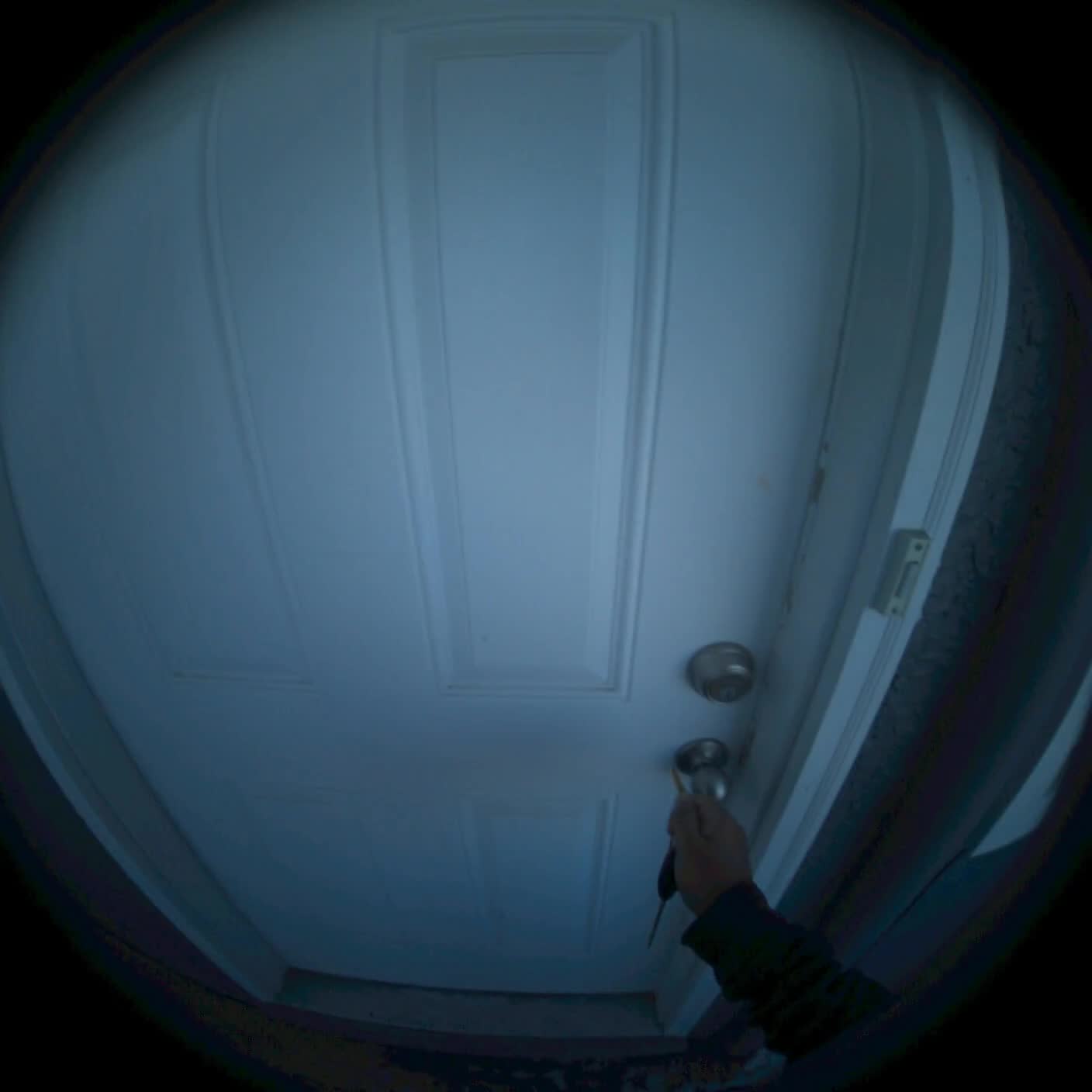}{e2}\hfill
        \qualframe{0.315\linewidth}{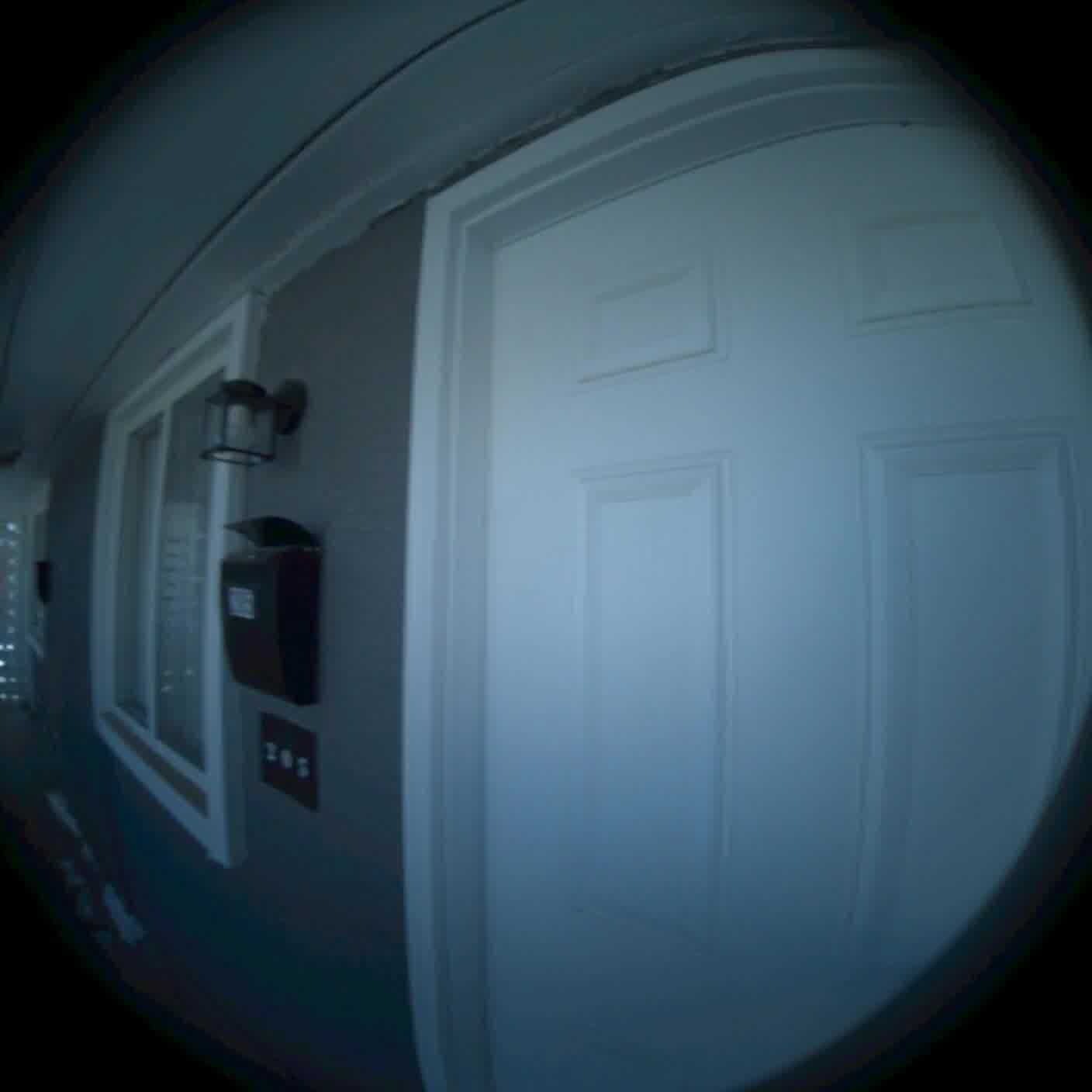}{e3}
        \vspace{0.35mm}

        {\tiny\textbf{Q:} I'm trying to order delivery on my phone. What was the apartment number on the door we came through earlier?}
        \qualanswers{
            \qualchoice{\CorrectTag}{The apartment number printed next to the door was 205.}
            \qualchoice{\VagueTag}{The apartment number was 205 or 207.}
            \qualchoice{\WrongTag}{The apartment number printed next to the door was 207.}
            \qualchoice{\NATag}{This question cannot be answered.}
        }{\qualselect{\CorrectTag}{\WrongTag}{\NATag}}
    \end{qualcard}
    \end{qualcardpair}
    \caption{Small visual evidence cases where the decisive cue is either a brief object observation or fine-grained text.}
    \label{fig:qualitative_visual_failures}
\end{figure}

\begin{figure}[t!]
    \centering
    \begin{qualcardpair}
    \begin{qualcard}{Temporal ordering failure}
        \qualframe{0.24\linewidth}{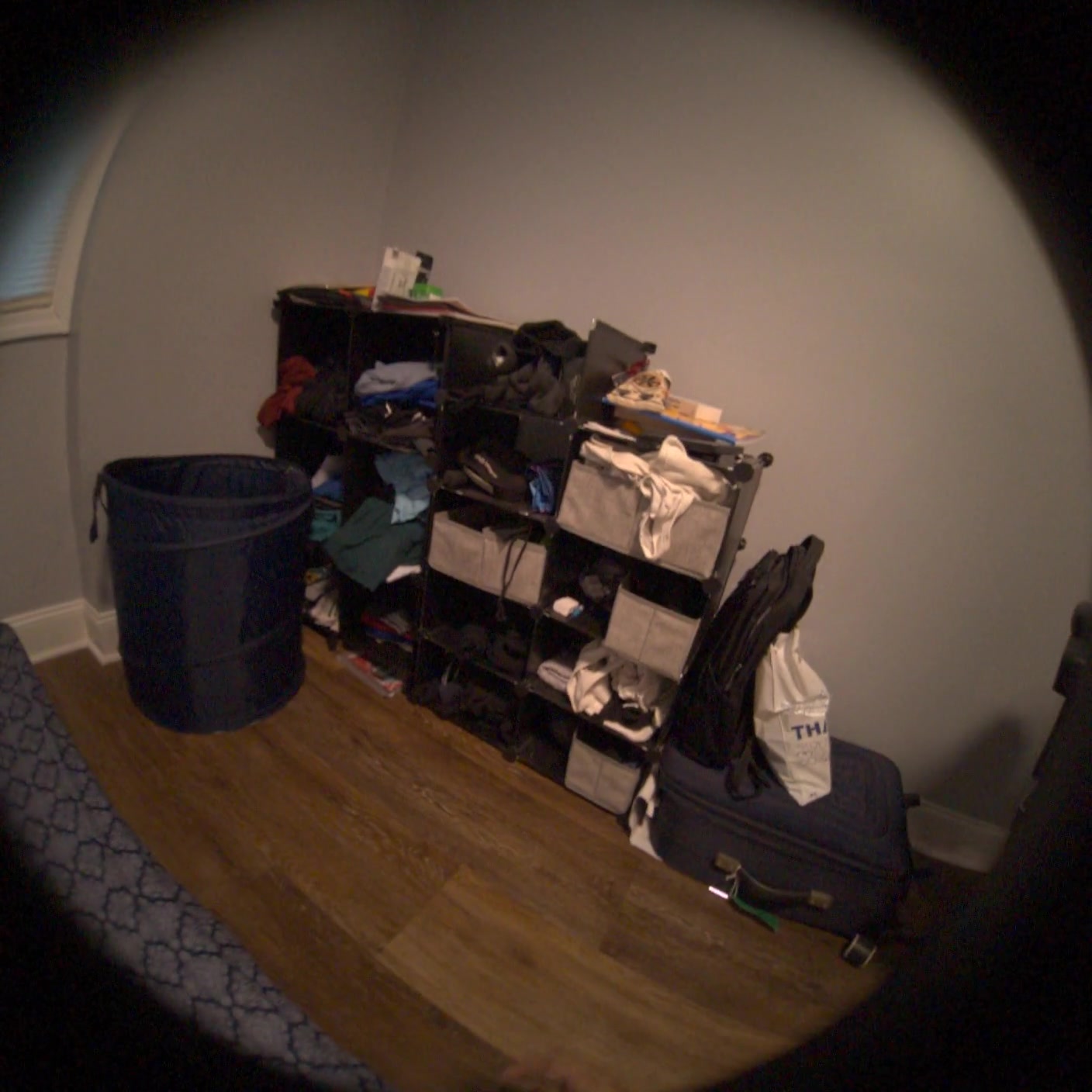}{socks}\hfill
        \qualframe{0.24\linewidth}{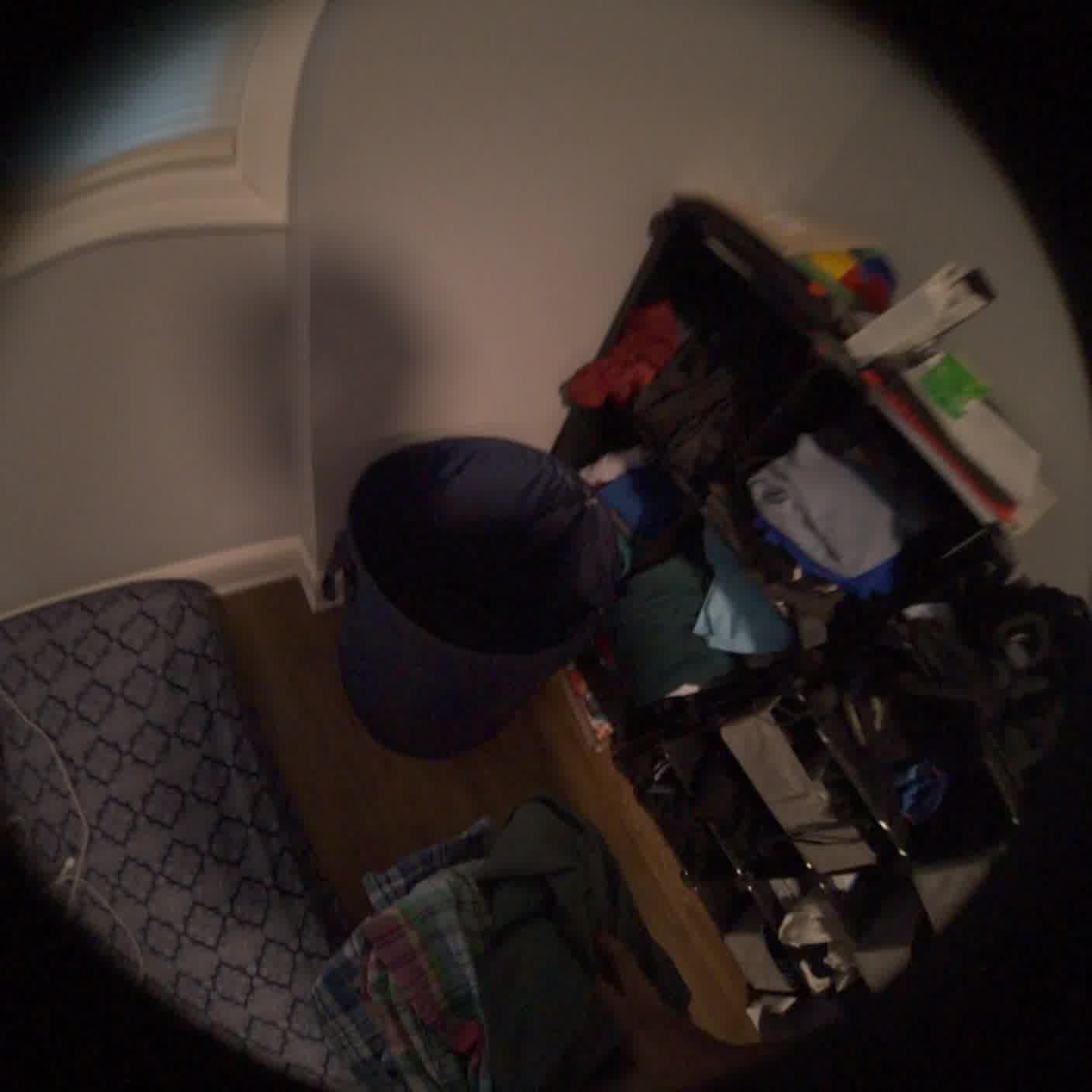}{shorts}\hfill
        \qualframe{0.24\linewidth}{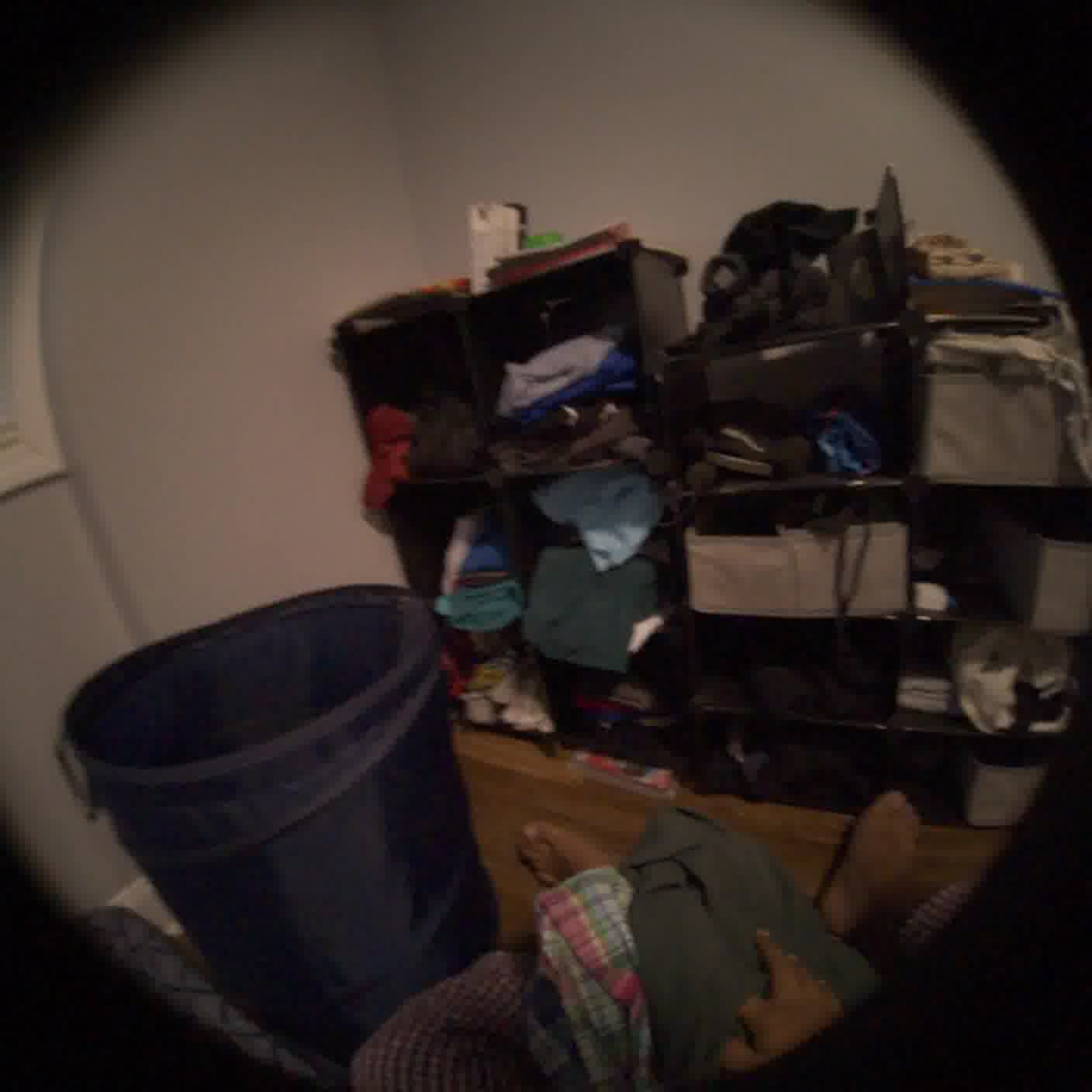}{shirt}\hfill
        \qualframe{0.24\linewidth}{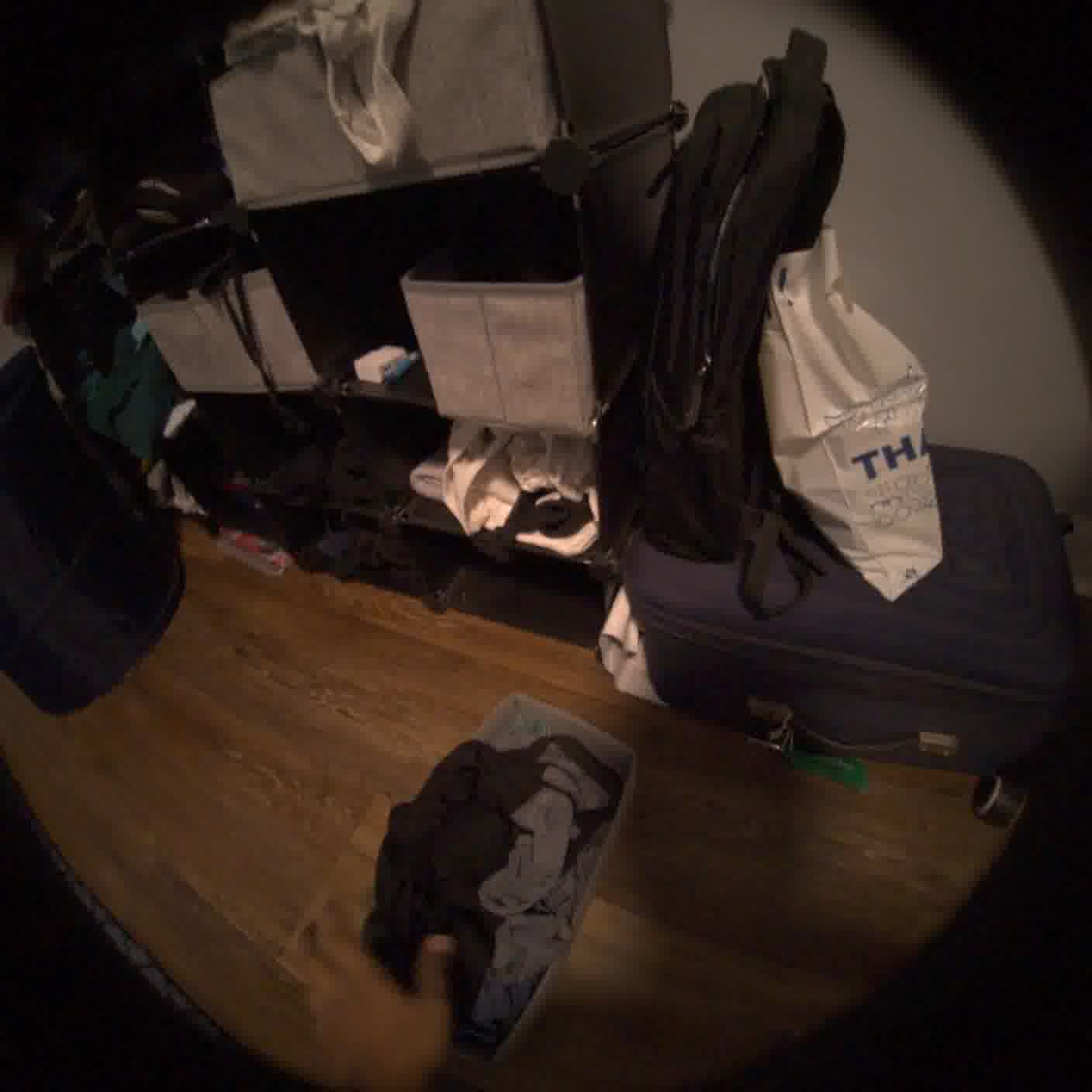}{later}
        \vspace{0.35mm}

        {\tiny\textbf{Q:} I'm trying to remember how I organized my laundry. In what order did I place the black socks, grey t-shirt, and blue patterned shorts into the storage organizer?}
        \qualanswers{
            \qualchoice{\NATag}{This question cannot be answered.}
            \qualchoice{\VagueTag}{You put the socks away before the shirts and shorts.}
            \qualchoice{\CorrectTag}{You placed the black socks in first, followed by the blue patterned shorts, and finally the grey t-shirt.}
            \qualchoice{\WrongTag}{You placed the grey t-shirt in first, followed by the black socks, and finally the blue patterned shorts.}
        }{\qualselect{\CorrectTag}{\NATag}{\NATag}}
    \end{qualcard}
    \end{qualcardpair}\hfill
    \begin{qualcardpair}

    \begin{qualcard}{State-change tracking success}
        \qualframe{0.24\linewidth}{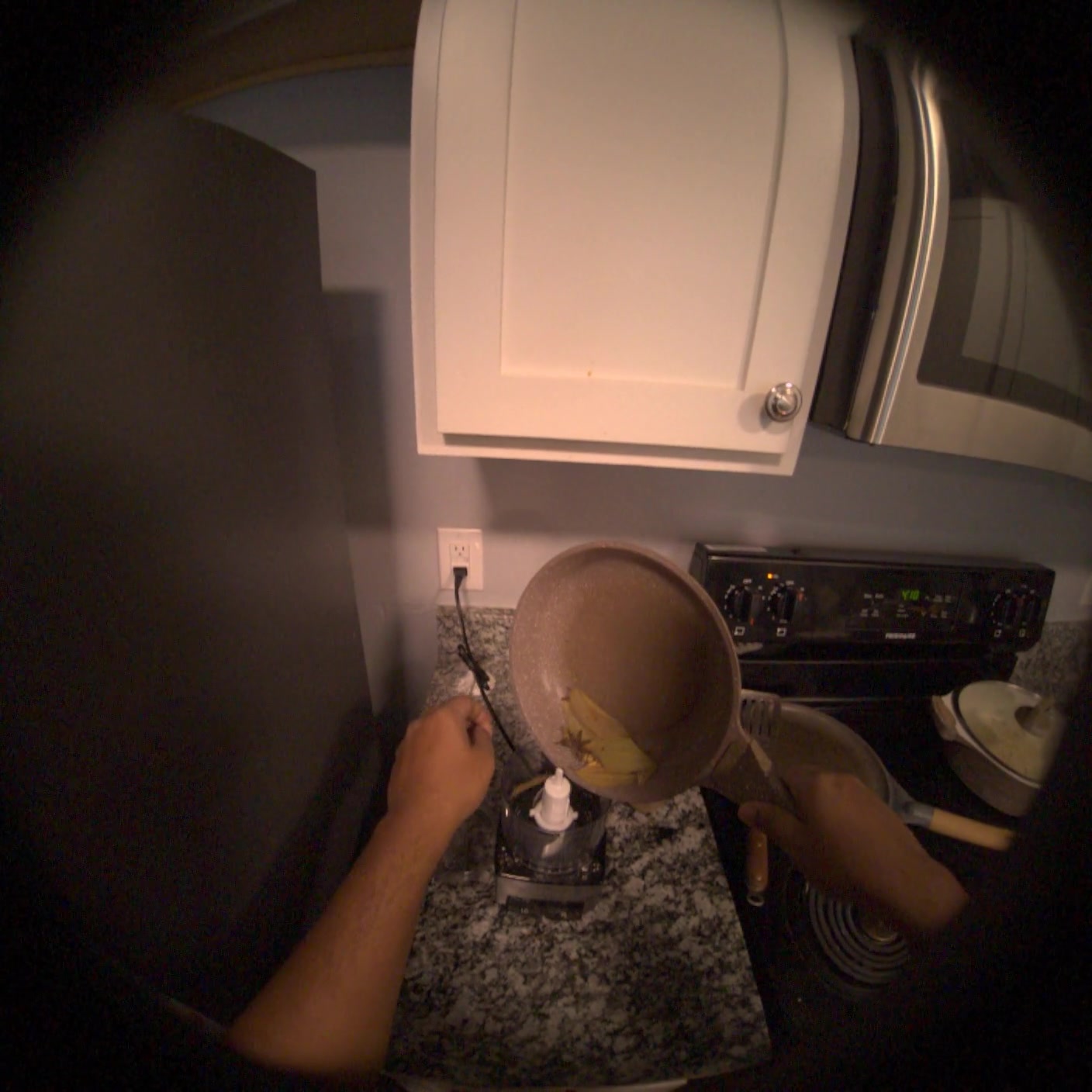}{used}\hfill
        \qualframe{0.24\linewidth}{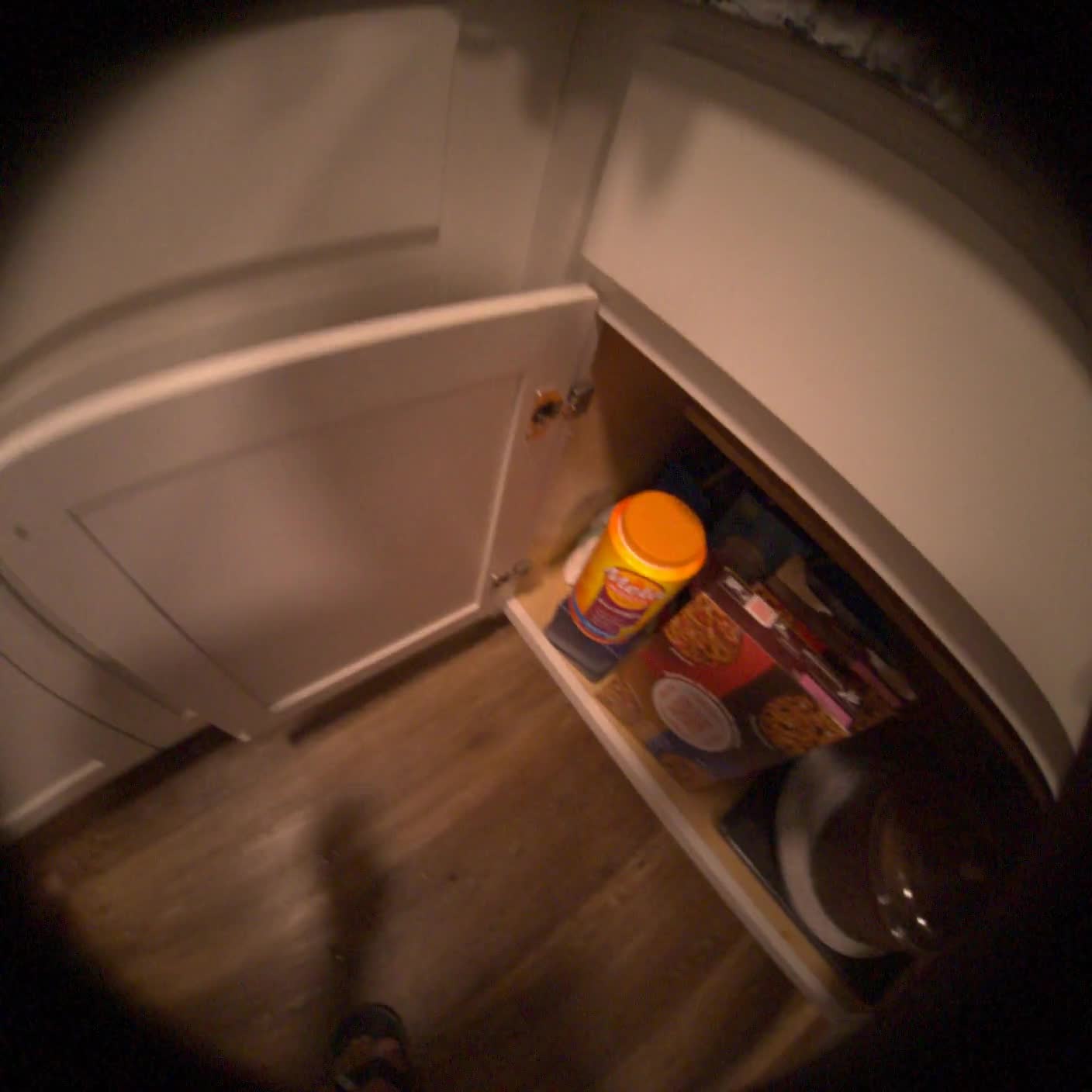}{washed}\hfill
        \qualframe{0.24\linewidth}{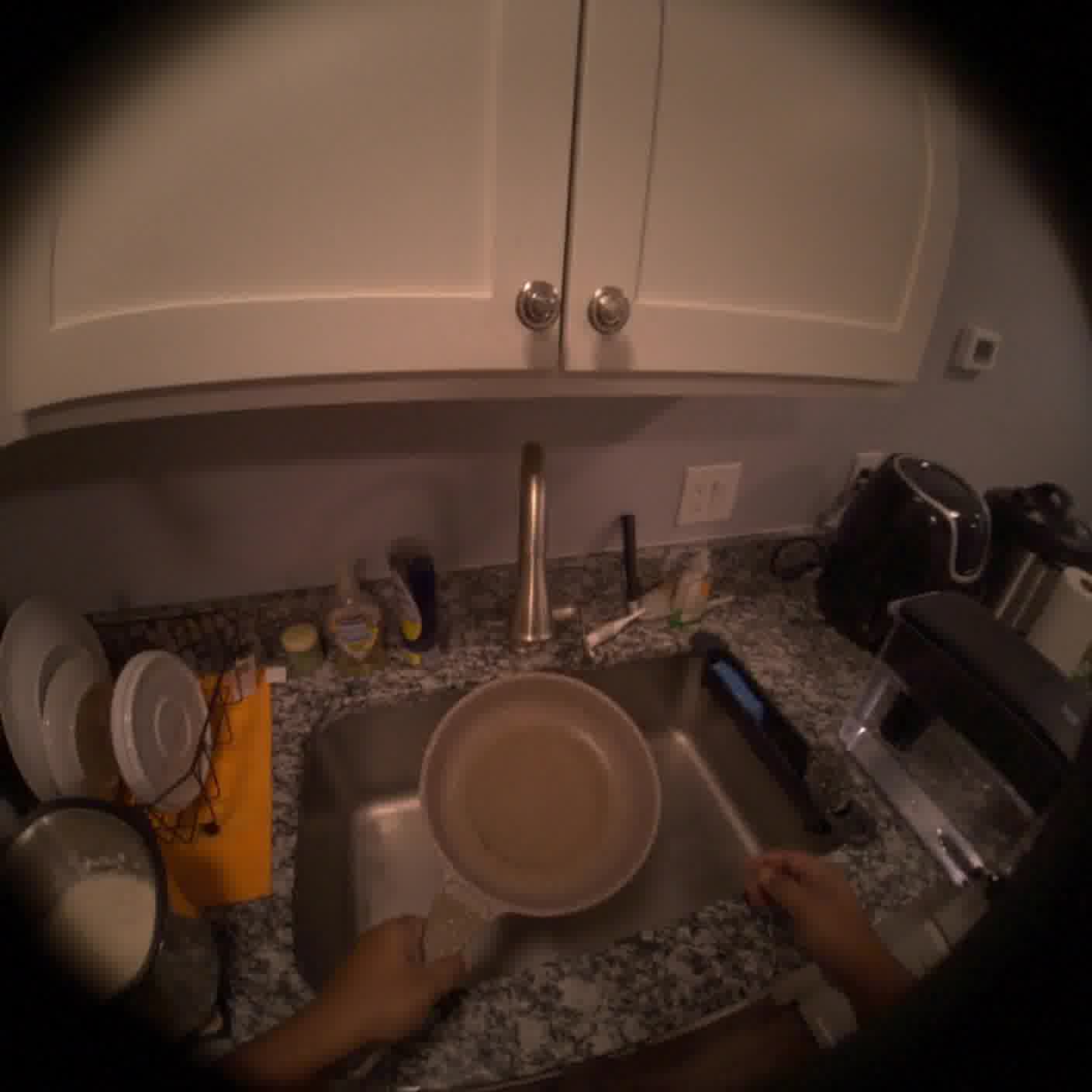}{stored}\hfill
        \qualframe{0.24\linewidth}{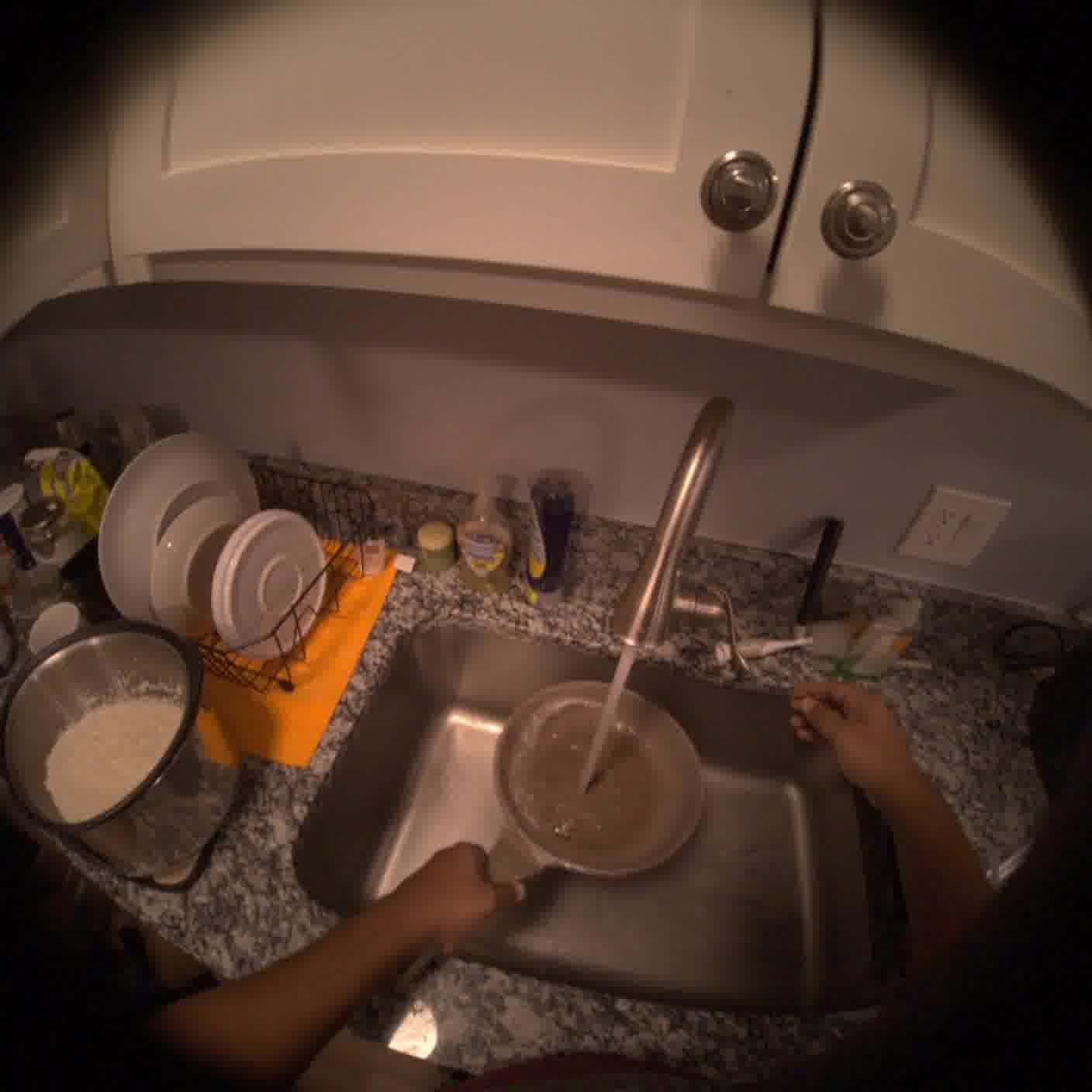}{later}
        \vspace{0.35mm}

        {\tiny\textbf{Q:} I'm loading up the dishwasher right now. Did I already wash the frying pan I used to roast the spices, or do I still need to track it down and put it in here?}
        \qualanswers{
            \qualchoice{\CorrectTag}{You already hand-washed the frying pan in the sink and put it away in a lower kitchen cabinet.}
            \qualchoice{\VagueTag}{You already cleaned the frying pan earlier today and put it away in one of the cabinets.}
            \qualchoice{\WrongTag}{You loaded the dirty frying pan into the bottom rack of the dishwasher, so it is already in there.}
            \qualchoice{\NATag}{This question cannot be answered.}
        }{\qualselect{\CorrectTag}{\CorrectTag}{\NATag}}
    \end{qualcard}
    \end{qualcardpair}
    \caption{Temporal reasoning examples where the answer depends on preserving action order or object state across evidence frames.}
    \label{fig:qualitative_temporal_reasoning}
\end{figure}

\begin{figure}[t!]
    \centering
    \begin{qualcardpair}
    \begin{qualcard}{Evidence unavailable: abstain}
        \qualframe{0.315\linewidth}{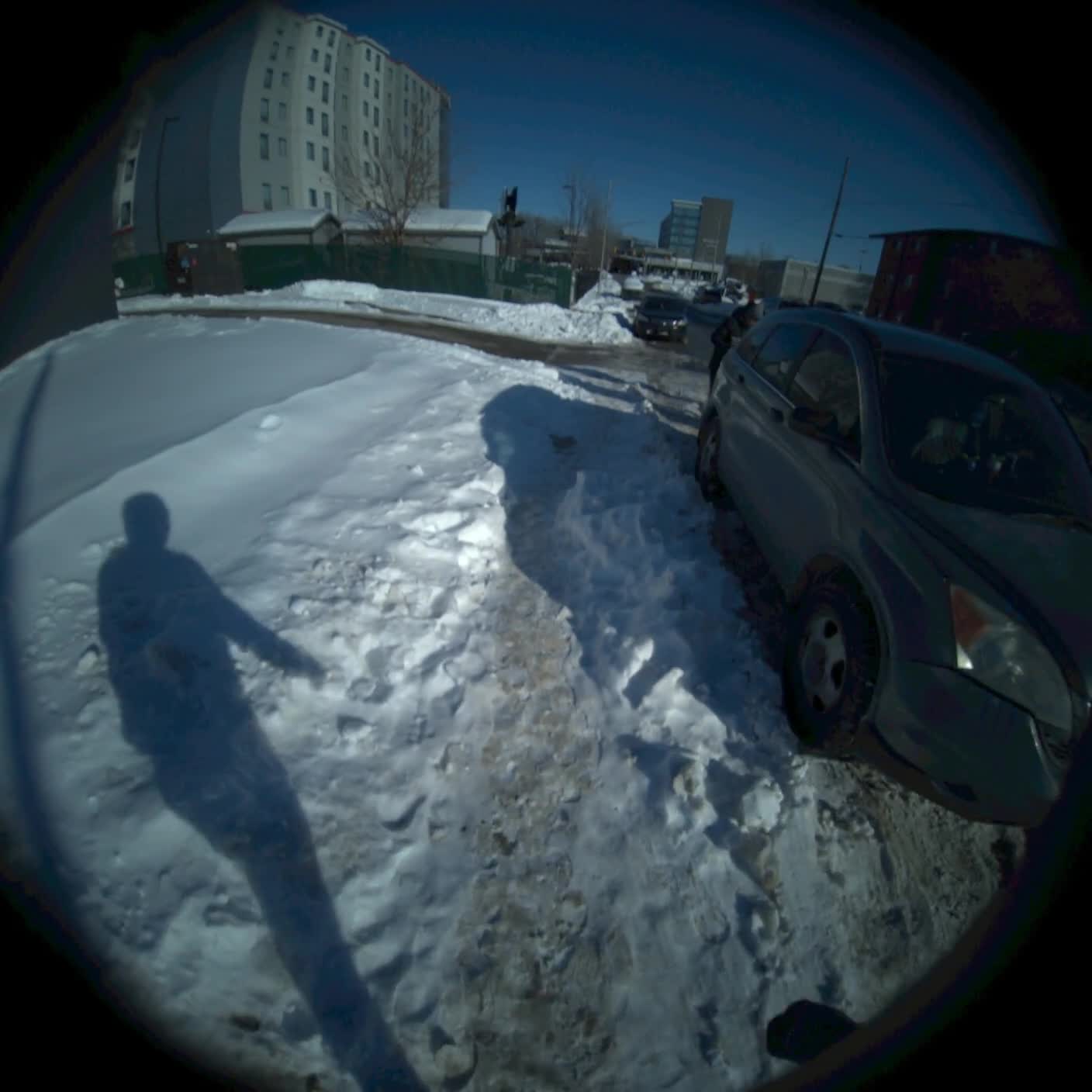}{e1}\hfill
        \qualframe{0.315\linewidth}{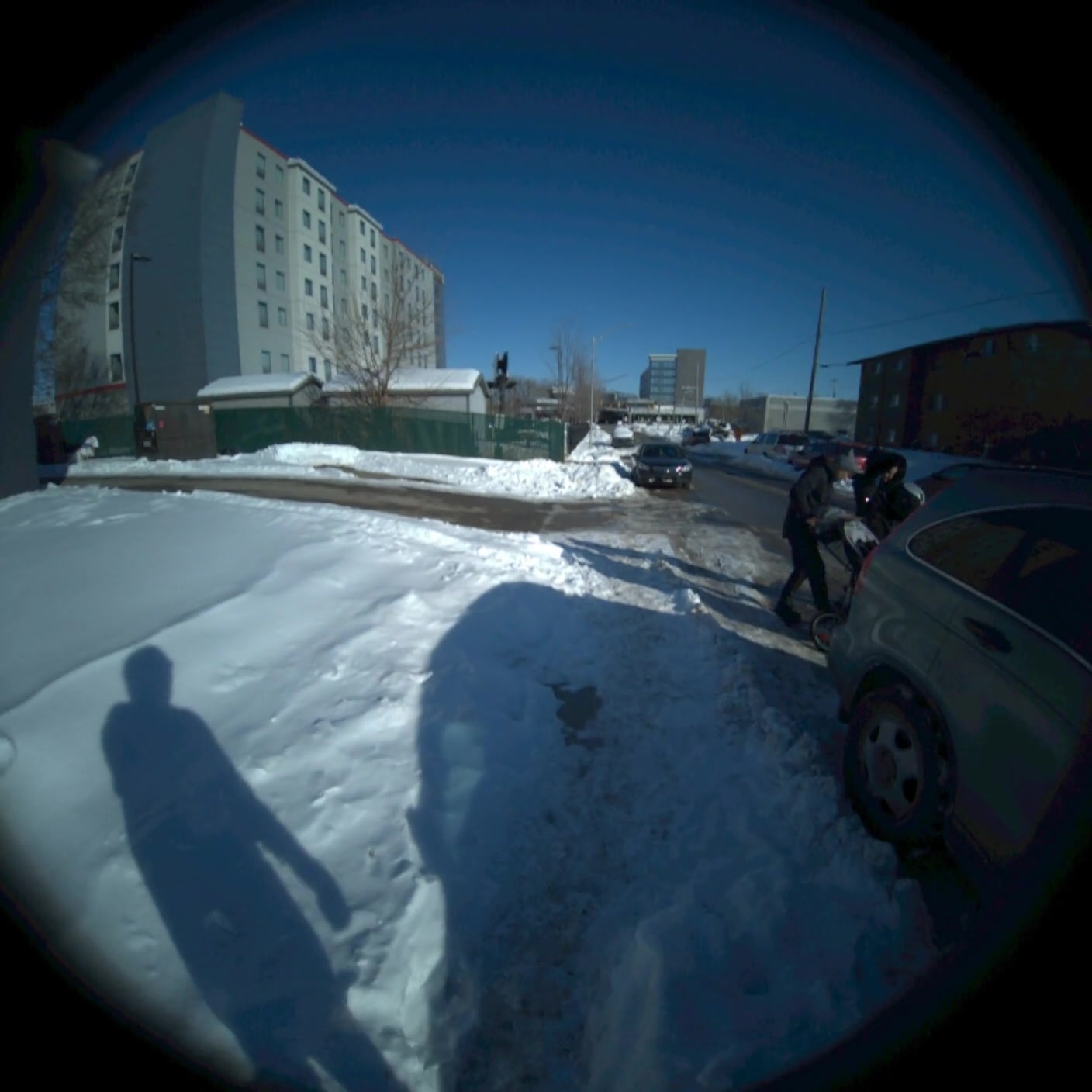}{e2}\hfill
        \qualframe{0.315\linewidth}{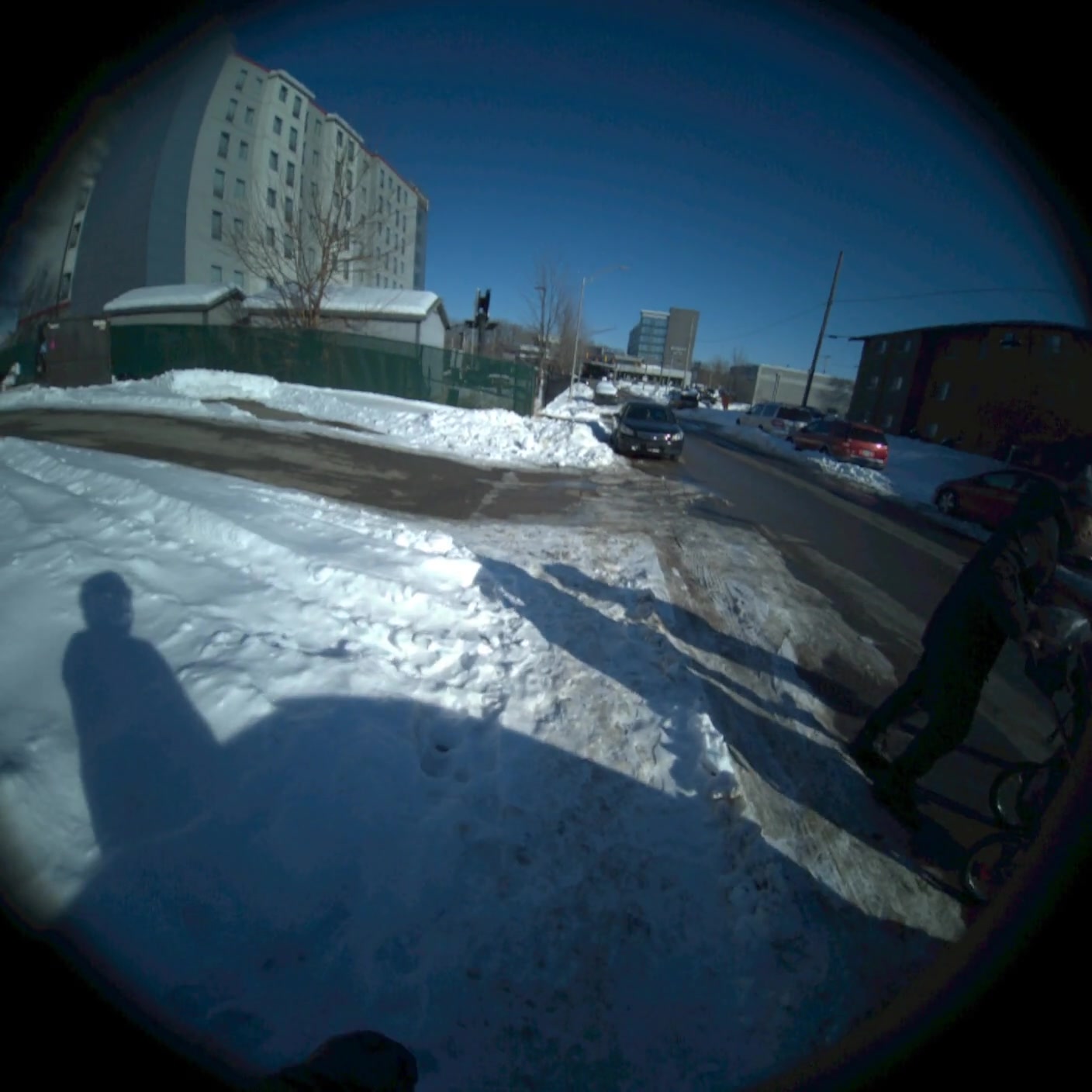}{e3}
        \vspace{0.35mm}

        {\tiny\textbf{Q:} I'm thinking about the walk we took earlier. What was the brand name printed on the stroller we passed?}
        \qualanswers{
            \qualchoice{\WrongTag}{The brand name printed on the stroller was Bugaboo.}
            \qualchoice{\WrongTag}{The brand name printed on the stroller was UPPAbaby.}
            \qualchoice{\WrongTag}{The brand name printed on the stroller was Graco.}
            \qualchoice{\NATag}{This question cannot be answered.}
        }{\qualselect{\NATag}{\NATag}{\NATag}}
    \end{qualcard}
    \end{qualcardpair}\hfill
    \begin{qualcardpair}

    \begin{qualcard}{False premise: abstain}
        \qualframe{0.315\linewidth}{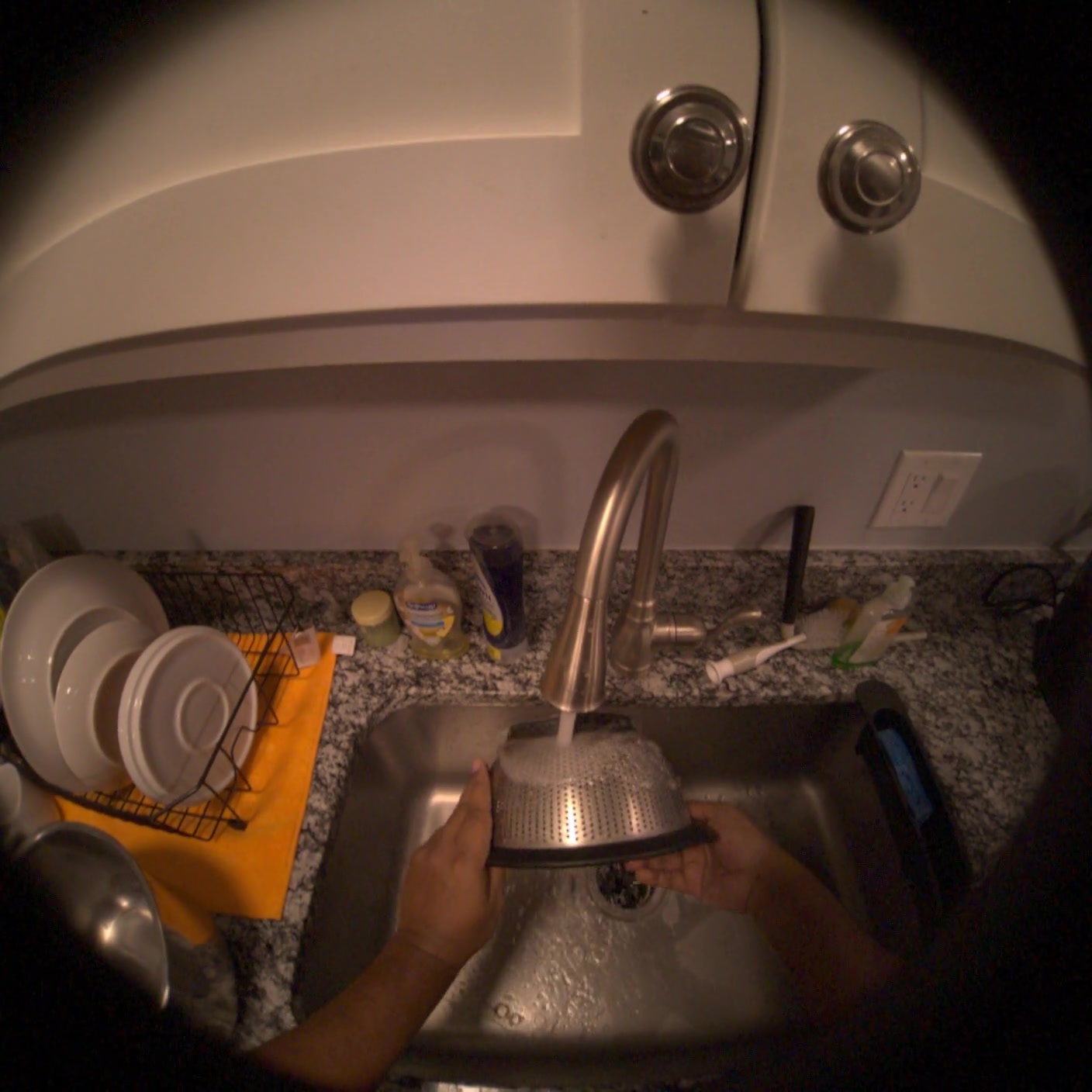}{e1}\hfill
        \qualframe{0.315\linewidth}{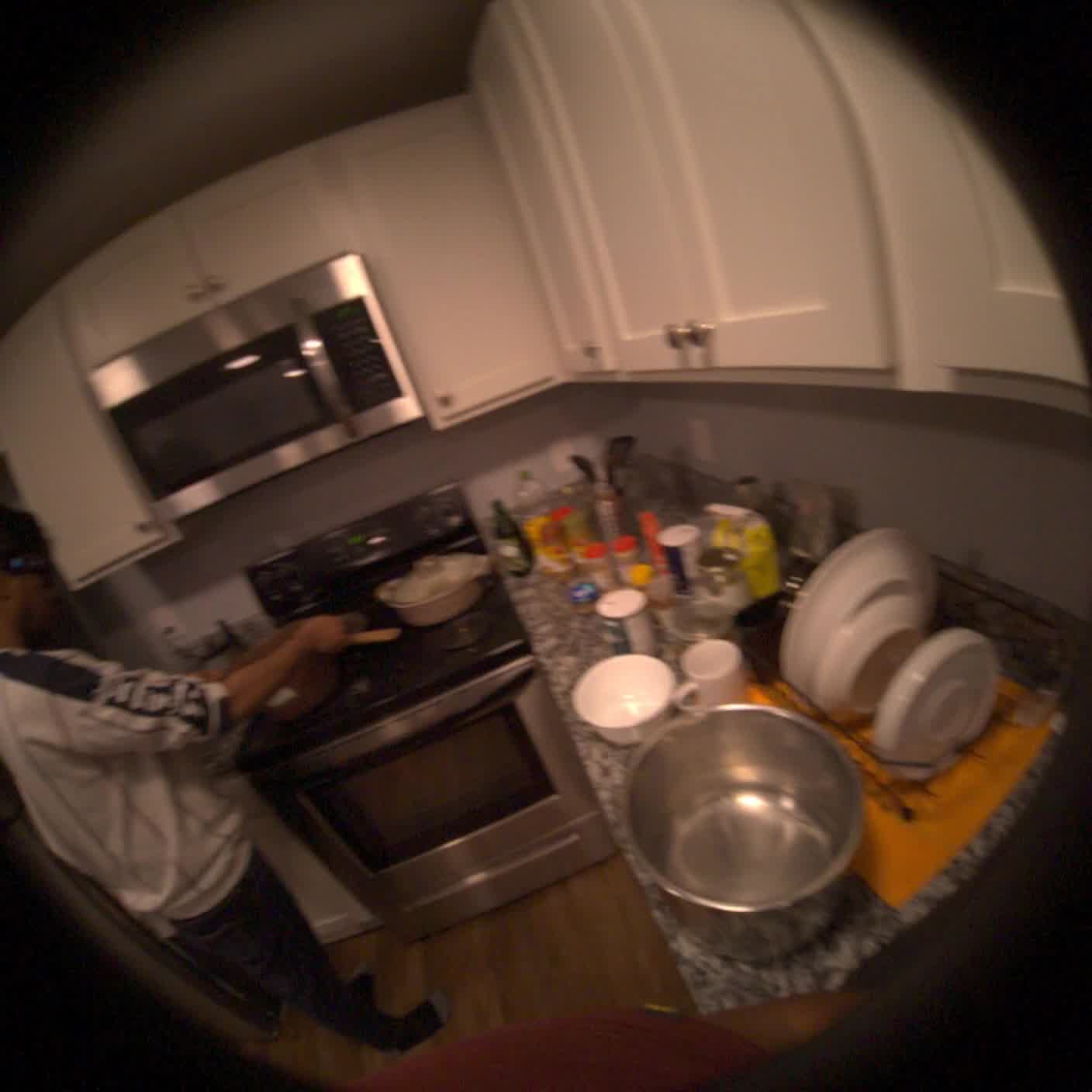}{e2}\hfill
        \qualframe{0.315\linewidth}{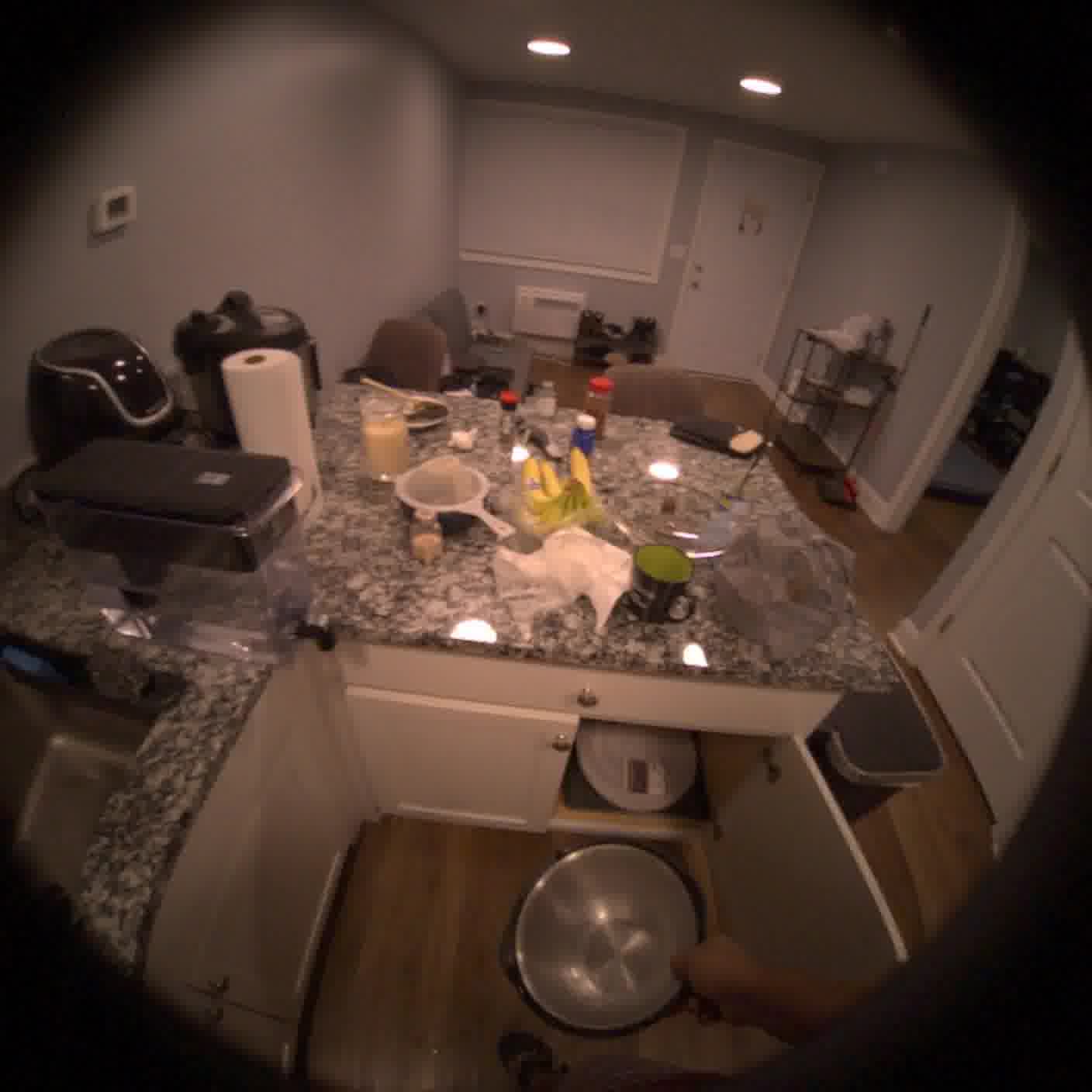}{e3}
        \vspace{0.35mm}

        {\tiny\textbf{Q:} I'm getting ready to cook. Where did I store the large metal colander after washing it with the blue sponge?}
        \qualanswers{
            \qualchoice{\WrongTag}{You placed the large metal colander on the top shelf of the upper kitchen cabinet, right next to the glass mixing bowls.}
            \qualchoice{\WrongTag}{You stored the large metal colander in the walk-in pantry on the bottom shelf, tucked away behind the large flour bags.}
            \qualchoice{\WrongTag}{You left the large metal colander to dry on the dish drying rack next to the sink, positioned behind the dinner plates.}
            \qualchoice{\NATag}{This question cannot be answered.}
        }{\qualselect{\NATag}{\WrongTag}{\NATag}}
    \end{qualcard}
    \end{qualcardpair}
    \caption{Abstention and premise-validation examples where related evidence exists but the requested fact is unsupported.}
    \label{fig:qualitative_abstention}
\end{figure}

\begin{figure}[t!]
    \centering
    \begin{qualcardpair}
    \begin{qualcard}{Multi-step arithmetic retrieval}
        \qualframe{0.315\linewidth}{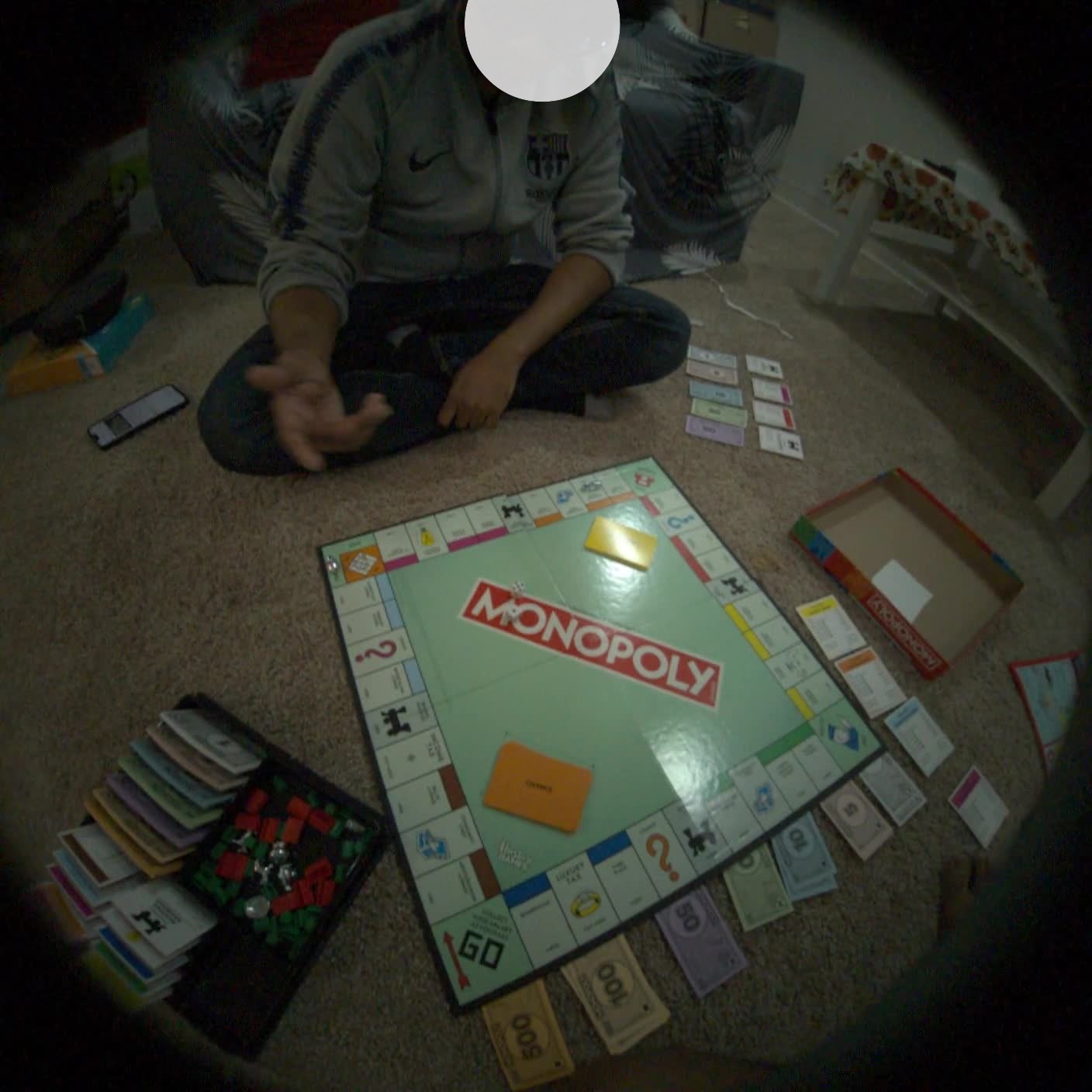}{rent}\hfill
        \qualframe{0.315\linewidth}{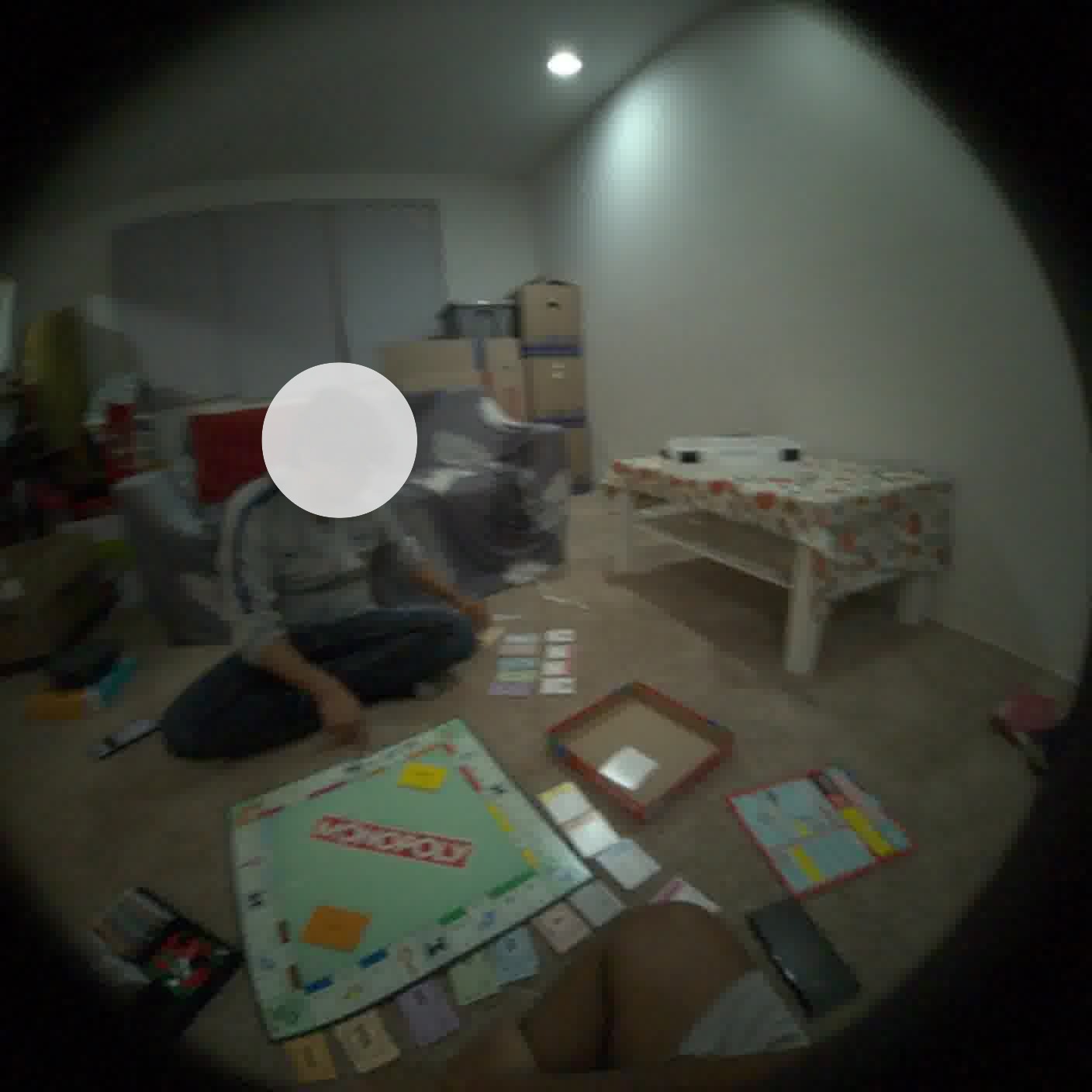}{later}\hfill
        \qualframe{0.315\linewidth}{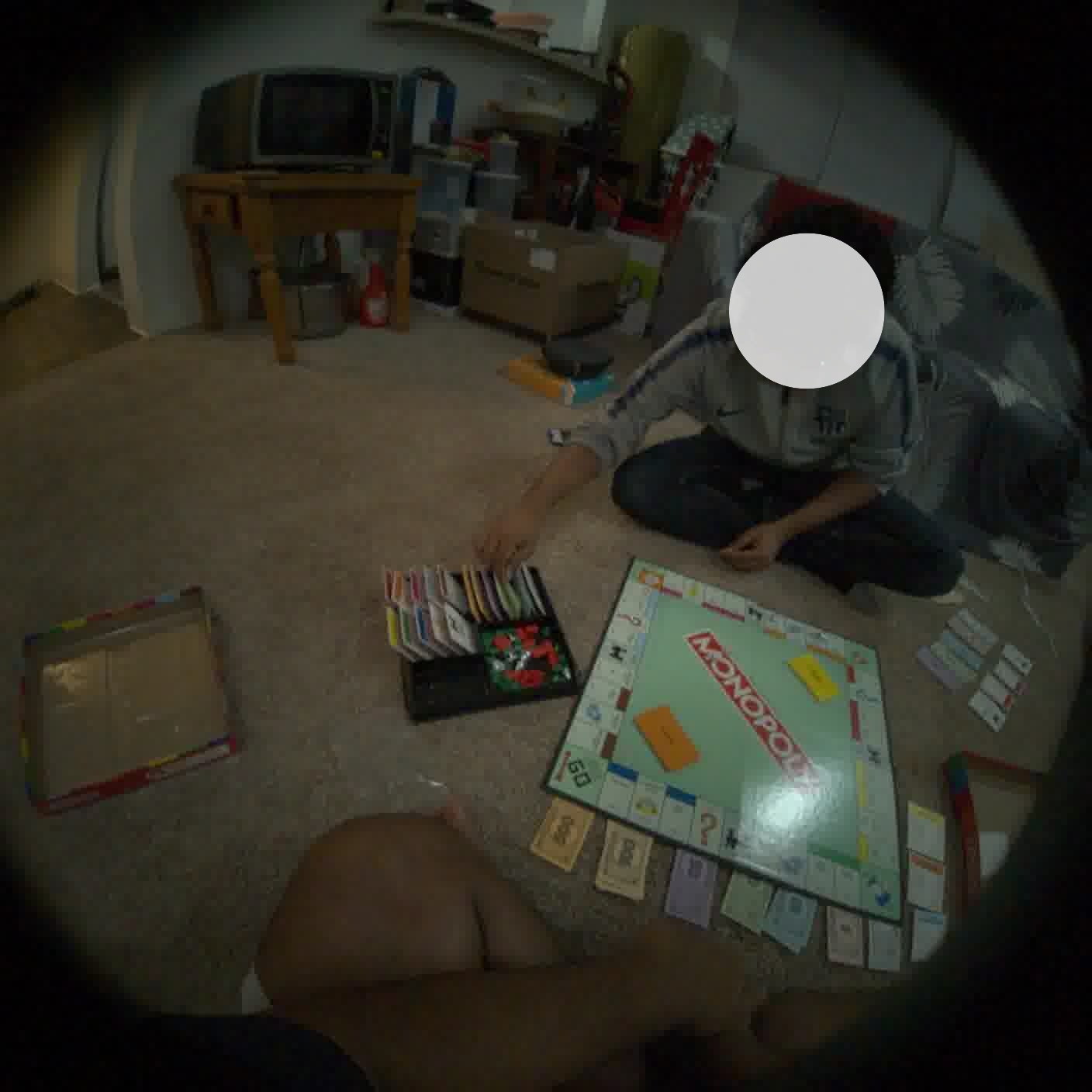}{calc}
        \vspace{0.35mm}

        {\tiny\textbf{Q:} Hey Aria, thinking back to our Monopoly game, what was the difference between that \$200 rent I paid B and the rent he paid me for St. Charles Place?}
        \qualanswers{
            \qualchoice{\CorrectTag}{You paid B \$186 more than he paid you for St. Charles Place.}
            \qualchoice{\VagueTag}{You paid B significantly more than he paid you for St. Charles Place.}
            \qualchoice{\WrongTag}{You paid B \$150 more than he paid you for St. Charles Place.}
            \qualchoice{\NATag}{This question cannot be answered.}
        }{\qualmodelselect{\CorrectTag}{\CorrectTag}{\NATag}}
    \end{qualcard}
    \end{qualcardpair}\hfill
    \begin{qualcardpair}

    \begin{qualcard}{Small-text reading}
        \qualframe{0.315\linewidth}{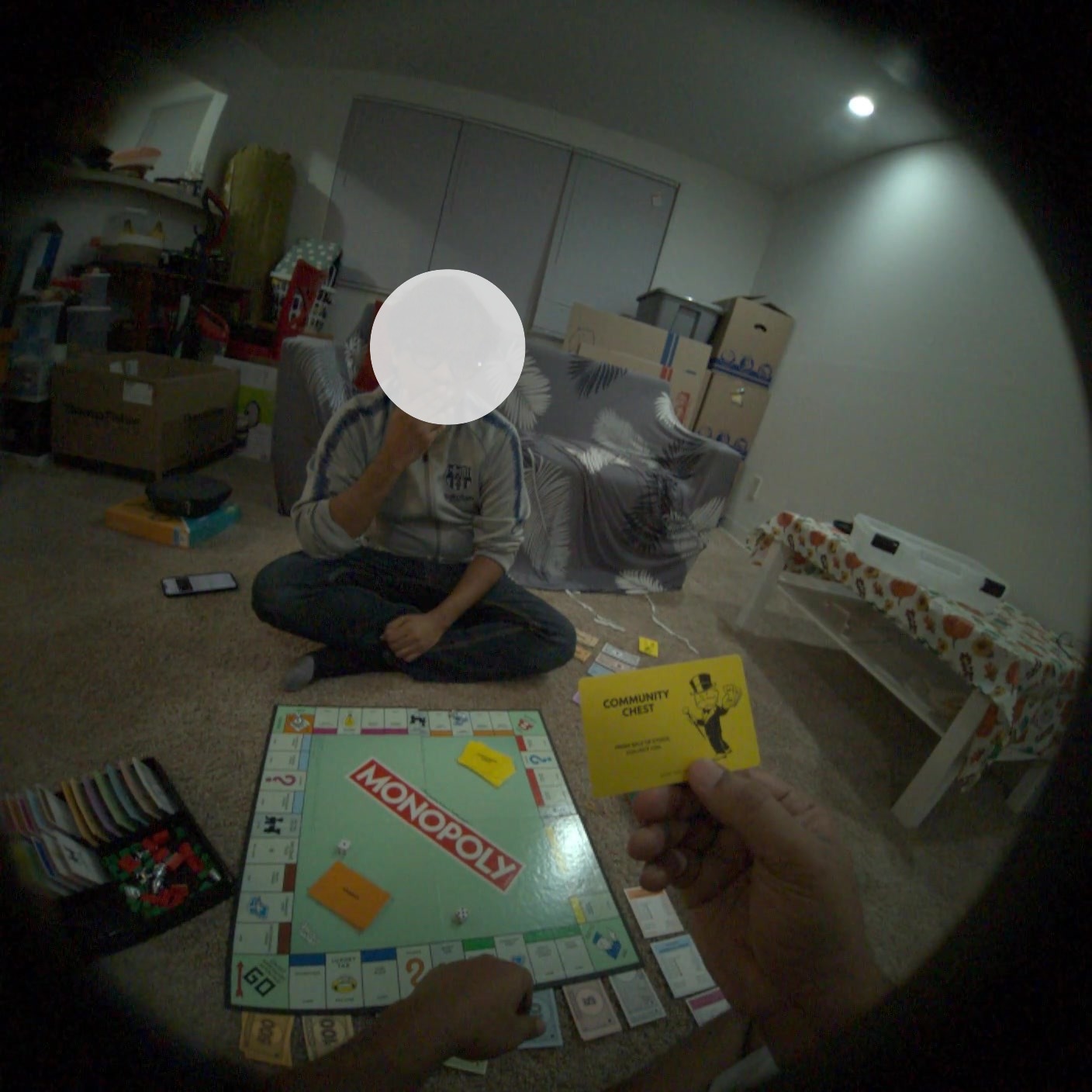}{card}\hfill
        \qualframe{0.315\linewidth}{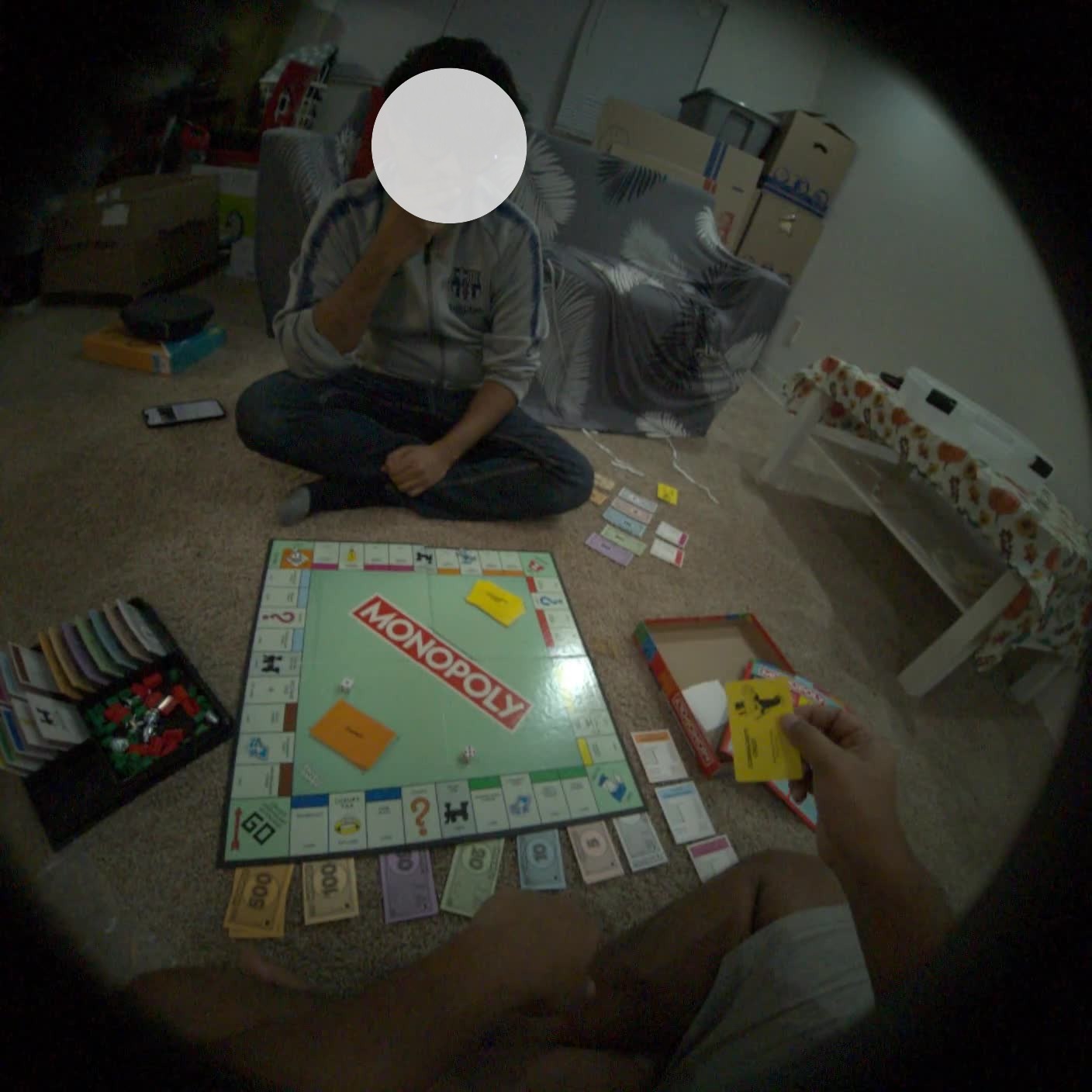}{text}\hfill
        \qualframe{0.315\linewidth}{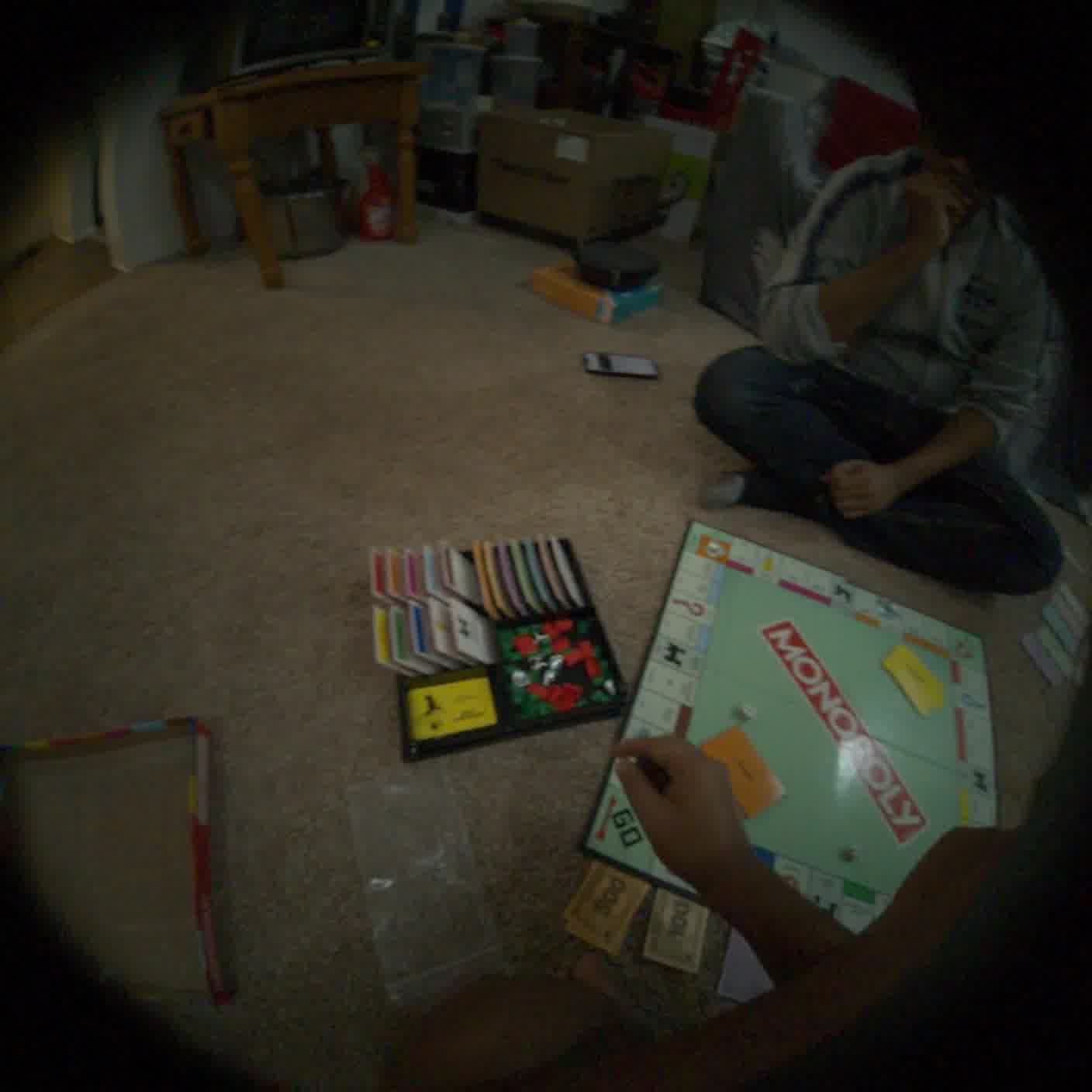}{later}
        \vspace{0.35mm}

        {\tiny\textbf{Q:} I'm texting my friend to settle a debate about our Monopoly game from the other day. What was the exact text on that Community Chest card I drew that gave me 50 dollars?}
        \qualanswers{
            \qualchoice{\CorrectTag}{The exact text on the card was ``FROM SALE OF STOCK YOU GET \$50.''}
            \qualchoice{\VagueTag}{The text on the card mentioned receiving \$50 from a stock sale.}
            \qualchoice{\WrongTag}{The exact text on the card was ``BANK ERROR IN YOUR FAVOR GET \$50.''}
            \qualchoice{\NATag}{This question cannot be answered.}
        }{\qualmodelselect{\CorrectTag}{\CorrectTag}{\NATag}}
    \end{qualcard}
    \end{qualcardpair}
    \caption{Answerable Flash and Gemini-3.1-Pro comparisons under Video-RAG. Flash commits to retrieved evidence, while Gemini-3.1-Pro abstains on multi-step arithmetic and small-text recall.}
    \label{fig:qualitative_flash_pro_answerable}
\end{figure}

\begin{figure}[t!]
    \centering
    \begin{qualcardpair}
    \begin{qualcard}{False object premise}
        \qualframe{0.315\linewidth}{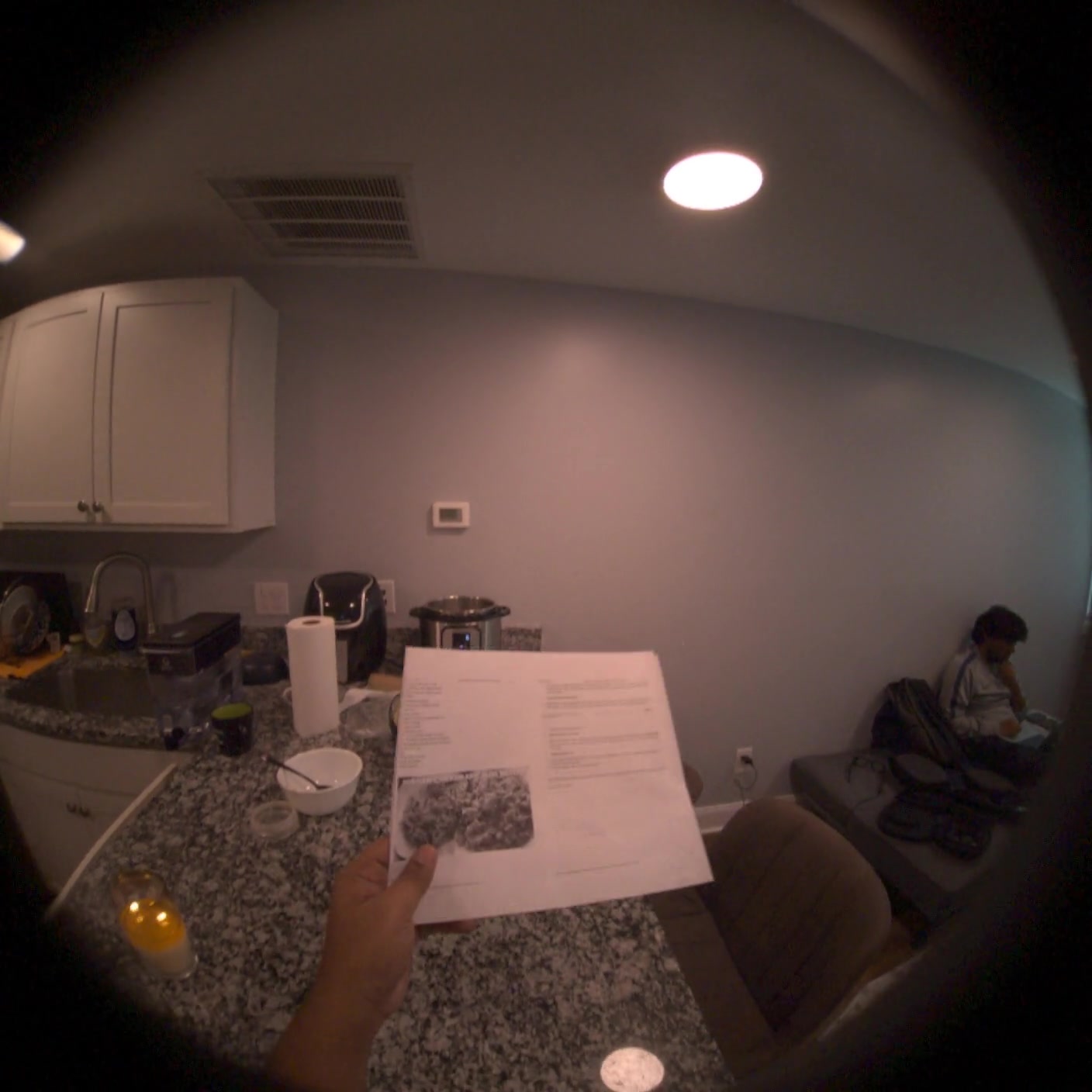}{recipe}\hfill
        \qualframe{0.315\linewidth}{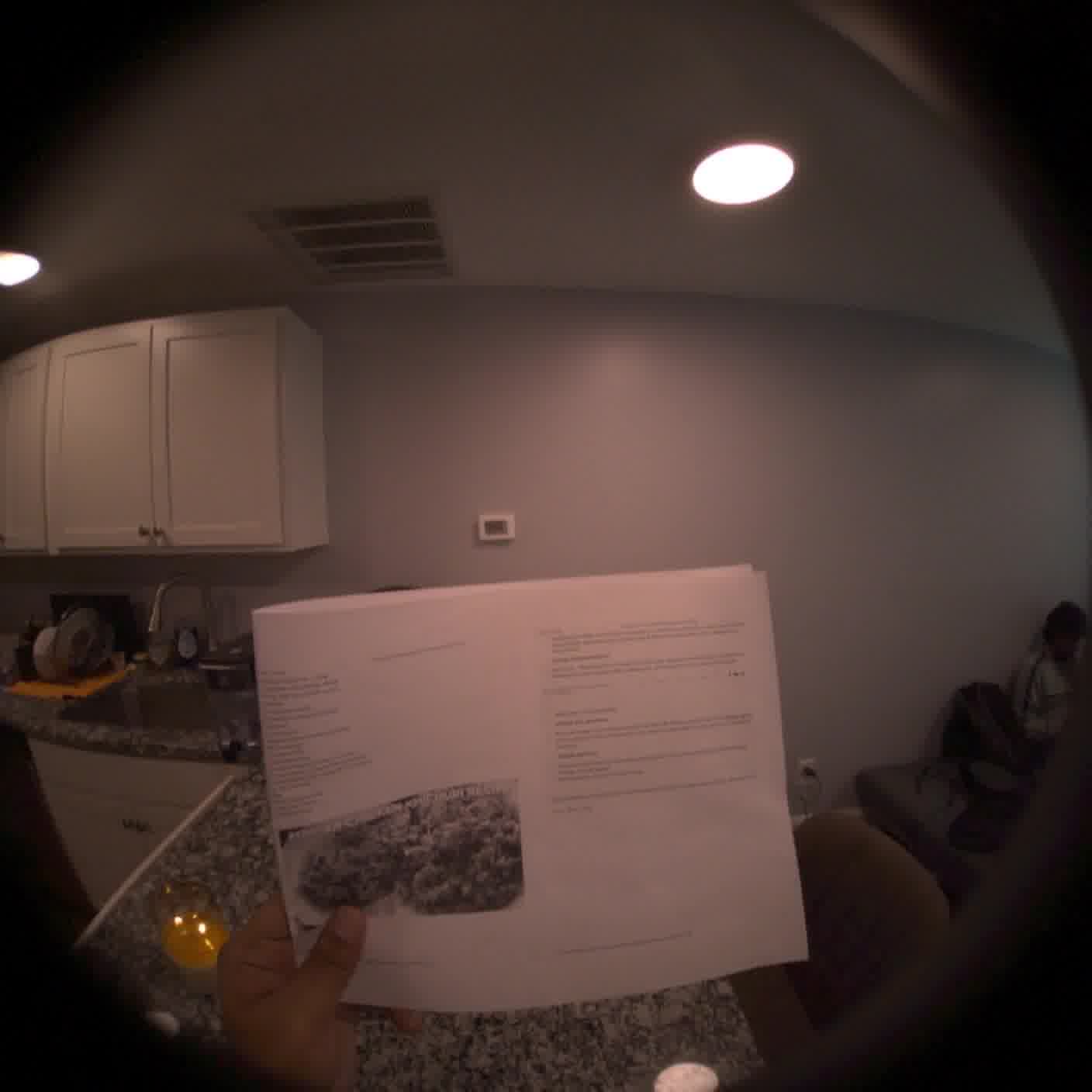}{counter}\hfill
        \qualframe{0.315\linewidth}{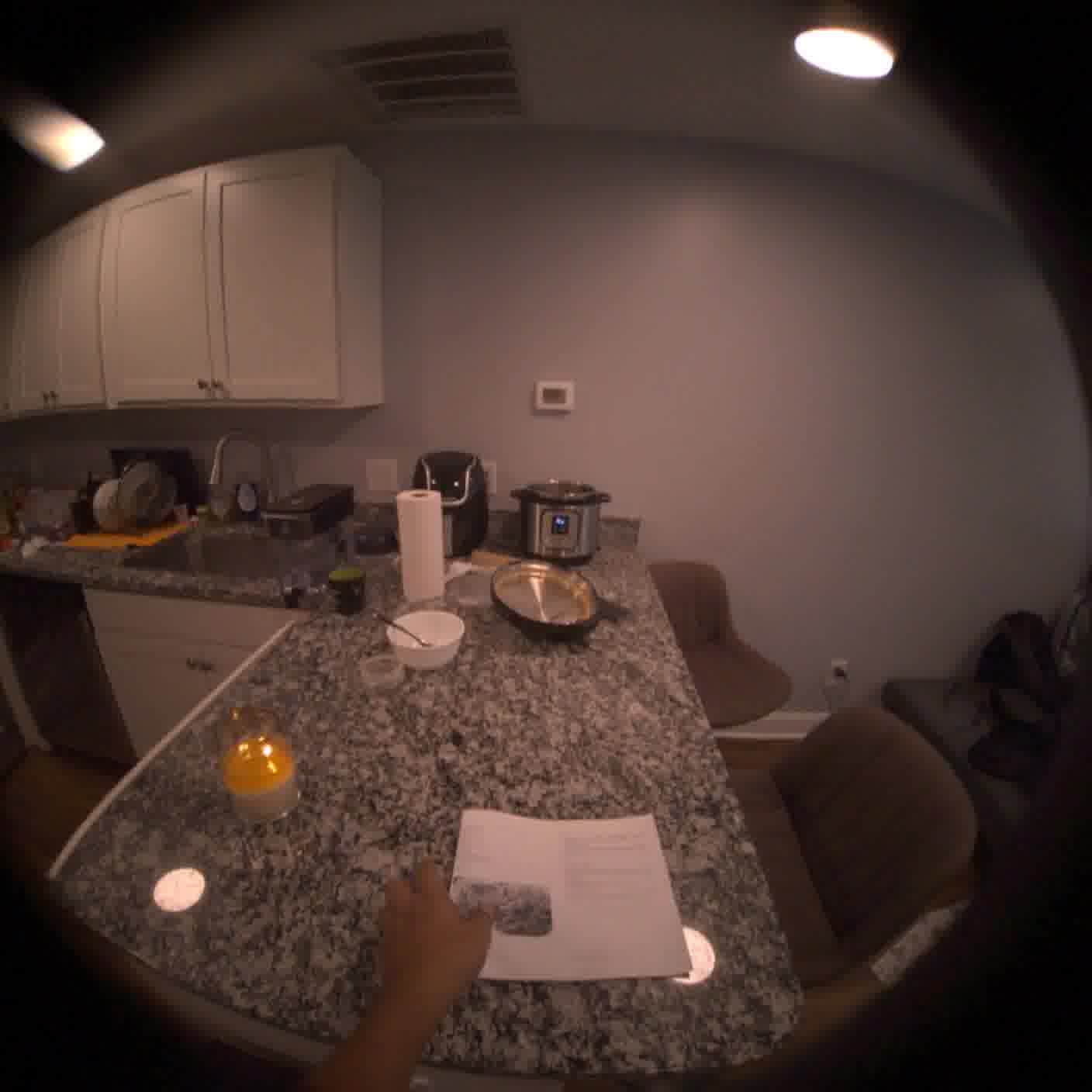}{later}
        \vspace{0.35mm}

        {\tiny\textbf{Q:} I'm looking for a tool to help with this recipe. Where did I leave the blue measuring cup?}
        \qualanswers{
            \qualchoice{\WrongTag}{The blue measuring cup is inside the dishwasher, which is currently closed.}
            \qualchoice{\WrongTag}{The blue measuring cup is in the microwave cabinet, tucked behind some plates.}
            \qualchoice{\WrongTag}{The blue measuring cup is on the counter, hidden behind the dish drying rack.}
            \qualchoice{\NATag}{This question cannot be answered.}
        }{\qualmodelselect{\NATag}{\WrongTag}{\NATag}}
    \end{qualcard}
    \end{qualcardpair}\hfill
    \begin{qualcardpair}

    \begin{qualcard}{False temporal premise}
        \qualframe{0.315\linewidth}{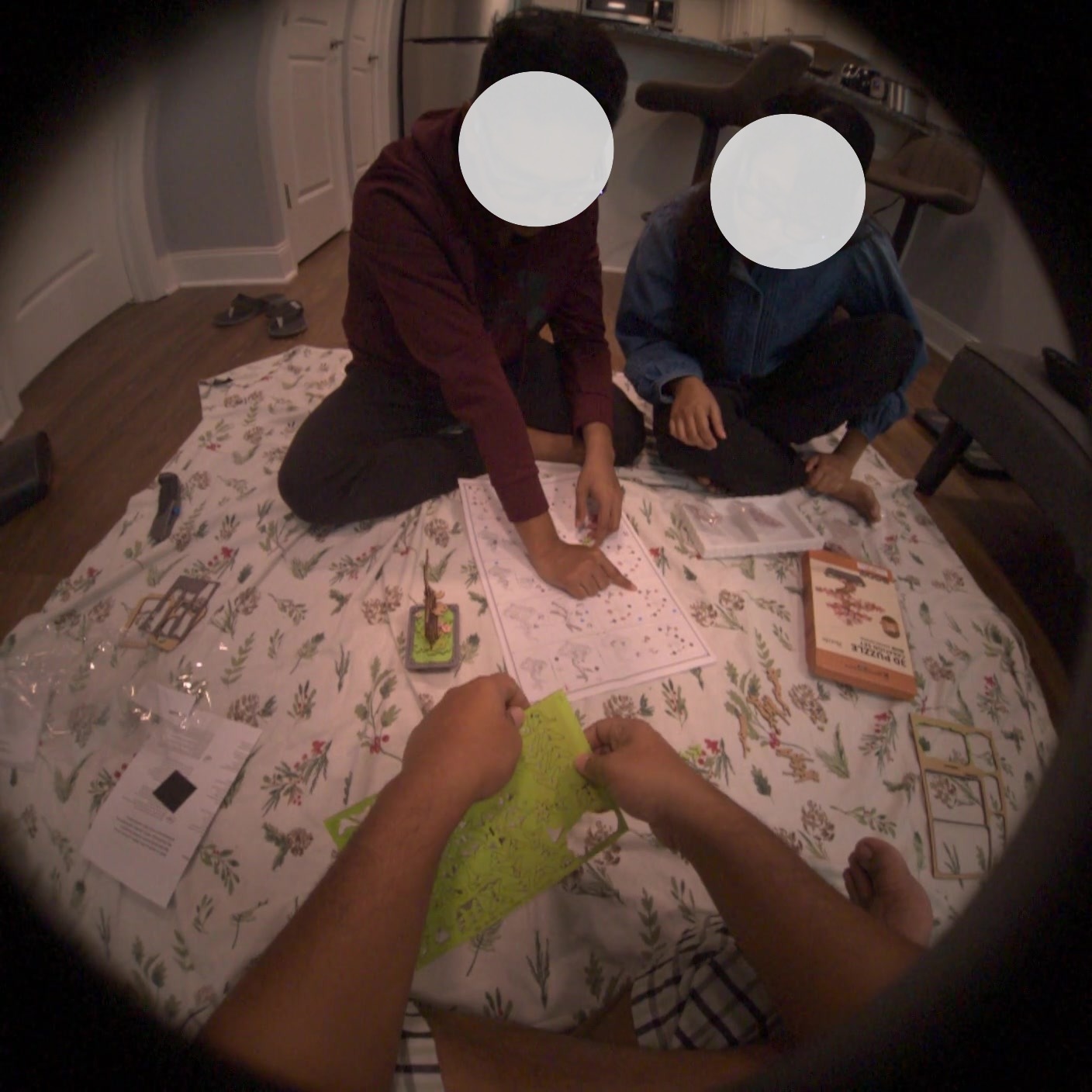}{game}\hfill
        \qualframe{0.315\linewidth}{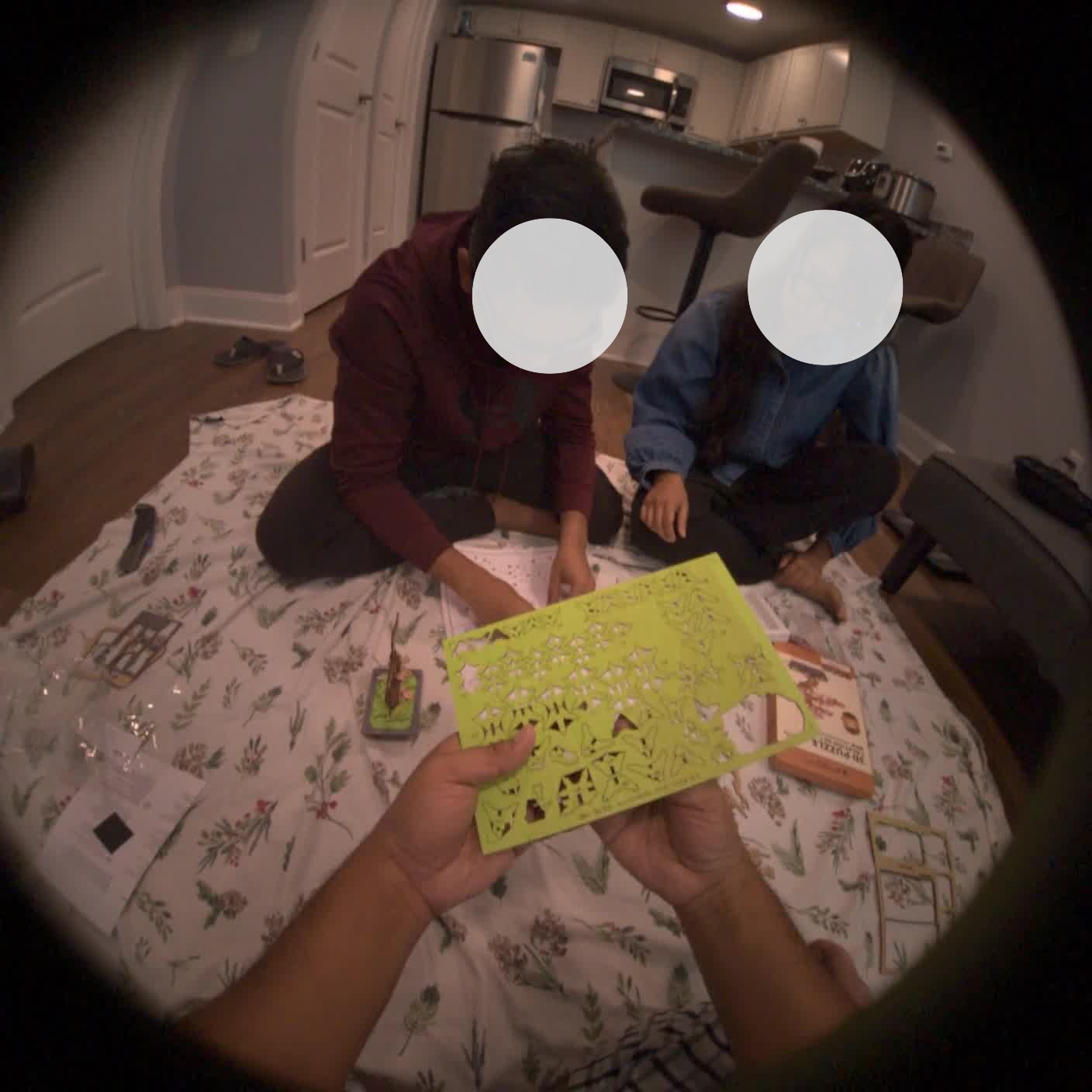}{floor}\hfill
        \qualframe{0.315\linewidth}{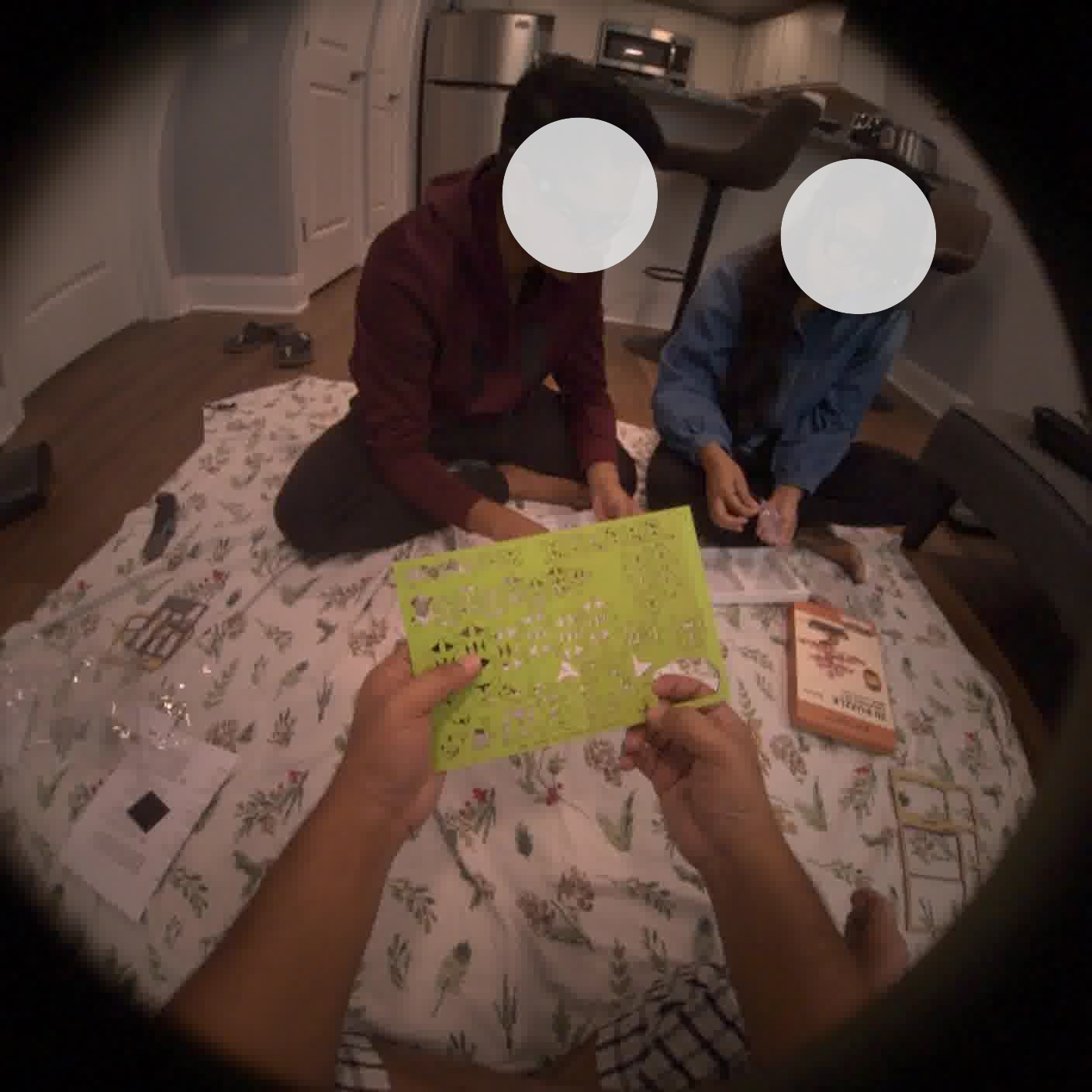}{later}
        \vspace{0.35mm}

        {\tiny\textbf{Q:} Now that we're sitting down to play Monopoly, did I actually pick up the Monopoly Deal box before I suggested we play it, or was it the other way around?}
        \qualanswers{
            \qualchoice{\WrongTag}{You picked up the Monopoly Deal card game box first, and then suggested playing it later.}
            \qualchoice{\WrongTag}{You suggested playing Monopoly Deal first, and then picked up the game box to show it.}
            \qualchoice{\WrongTag}{You picked up the box and suggested playing it while clearing the living room floor.}
            \qualchoice{\NATag}{This question cannot be answered.}
        }{\qualmodelselect{\NATag}{\WrongTag}{\NATag}}
    \end{qualcard}
    \end{qualcardpair}
    \caption{Unanswerable Flash and Gemini-3.1-Pro comparisons under Video-RAG. Gemini-3.1-Pro correctly rejects unsupported object and temporal premises, while Flash gives plausible but fabricated details.}
    \label{fig:qualitative_flash_pro_unanswerable}
\end{figure}

\begin{figure}[t!]
    \centering
    \begin{qualcardpair}
    \begin{qualcard}{Conversational detail}
        \qualframe{0.315\linewidth}{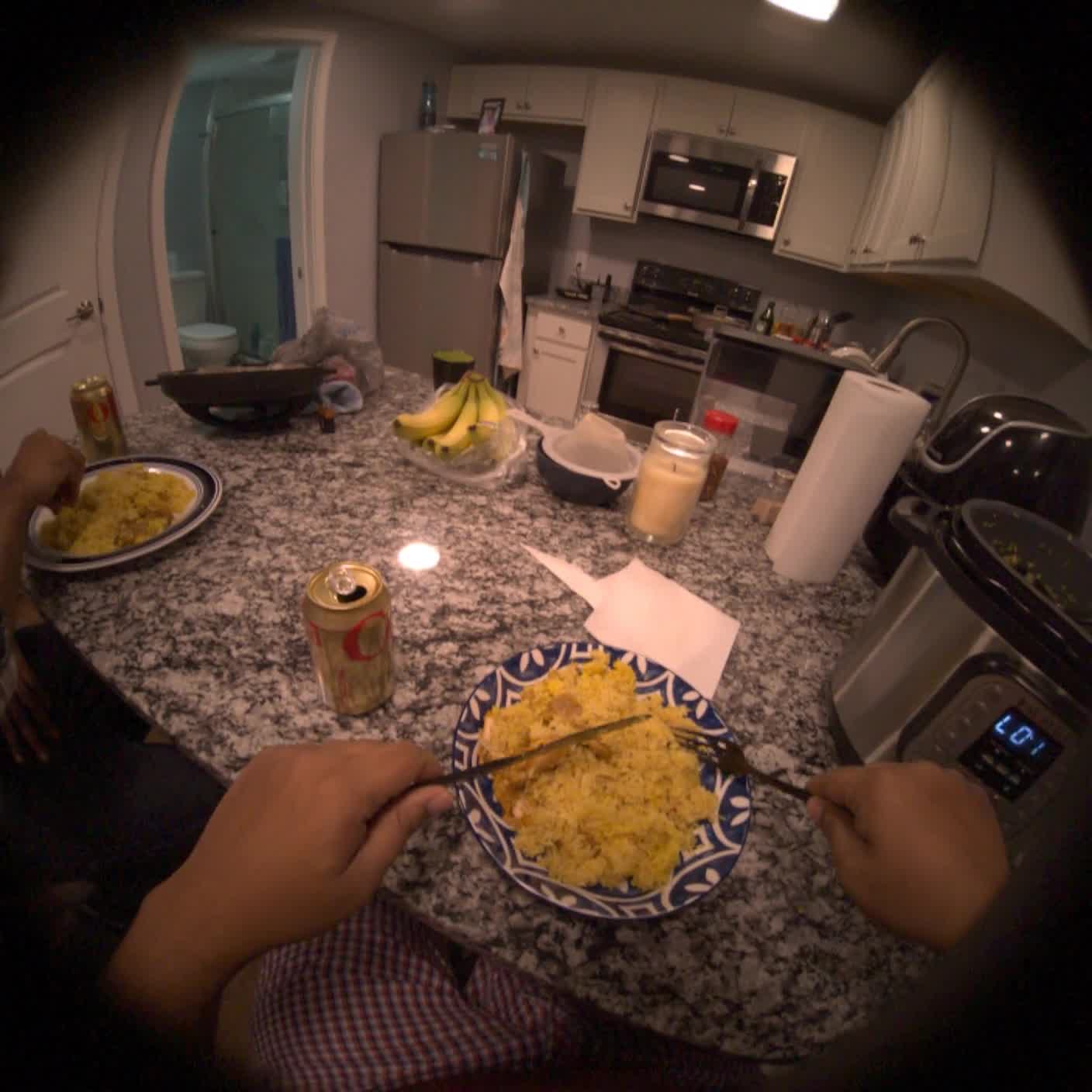}{talk}\hfill
        \qualframe{0.315\linewidth}{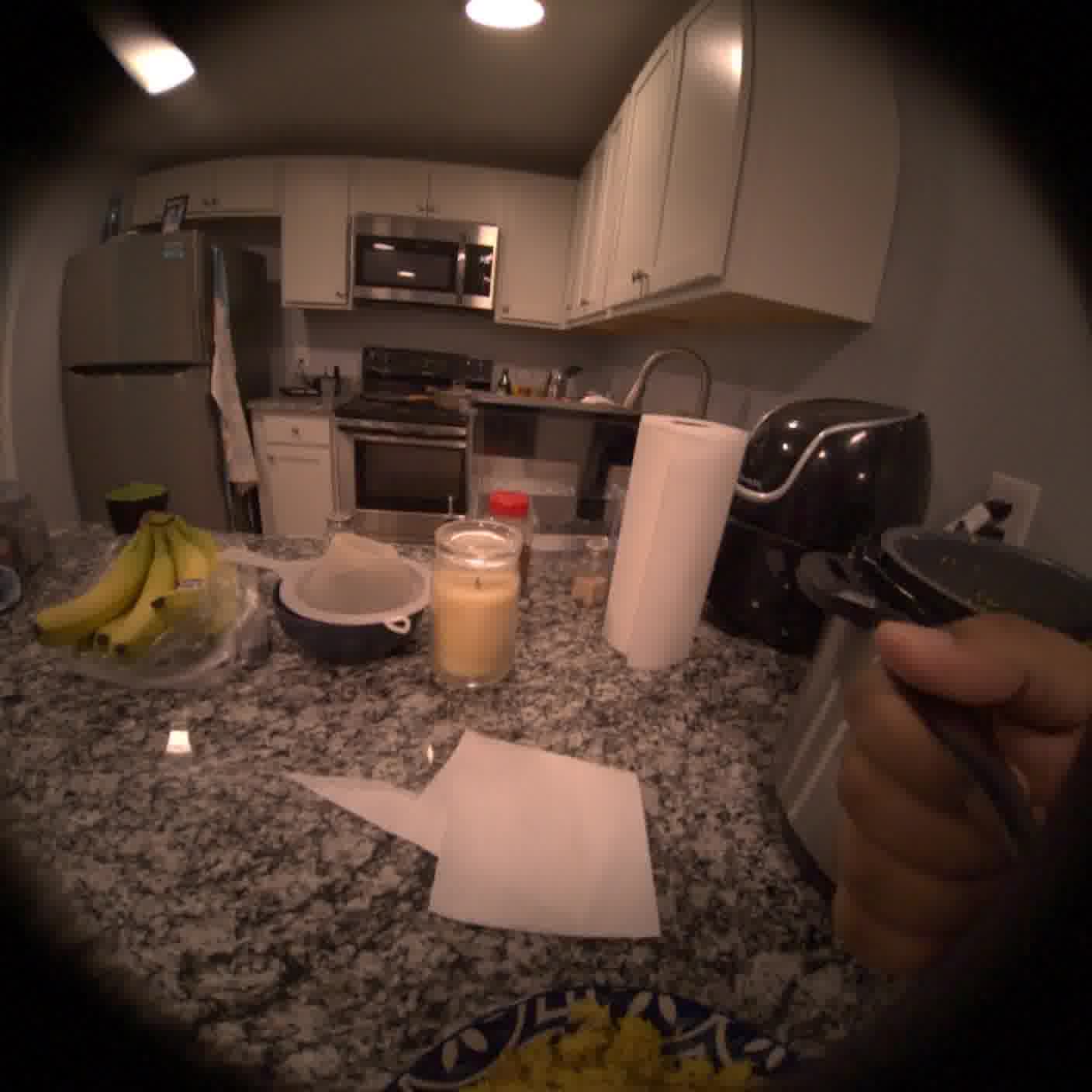}{meal}\hfill
        \qualframe{0.315\linewidth}{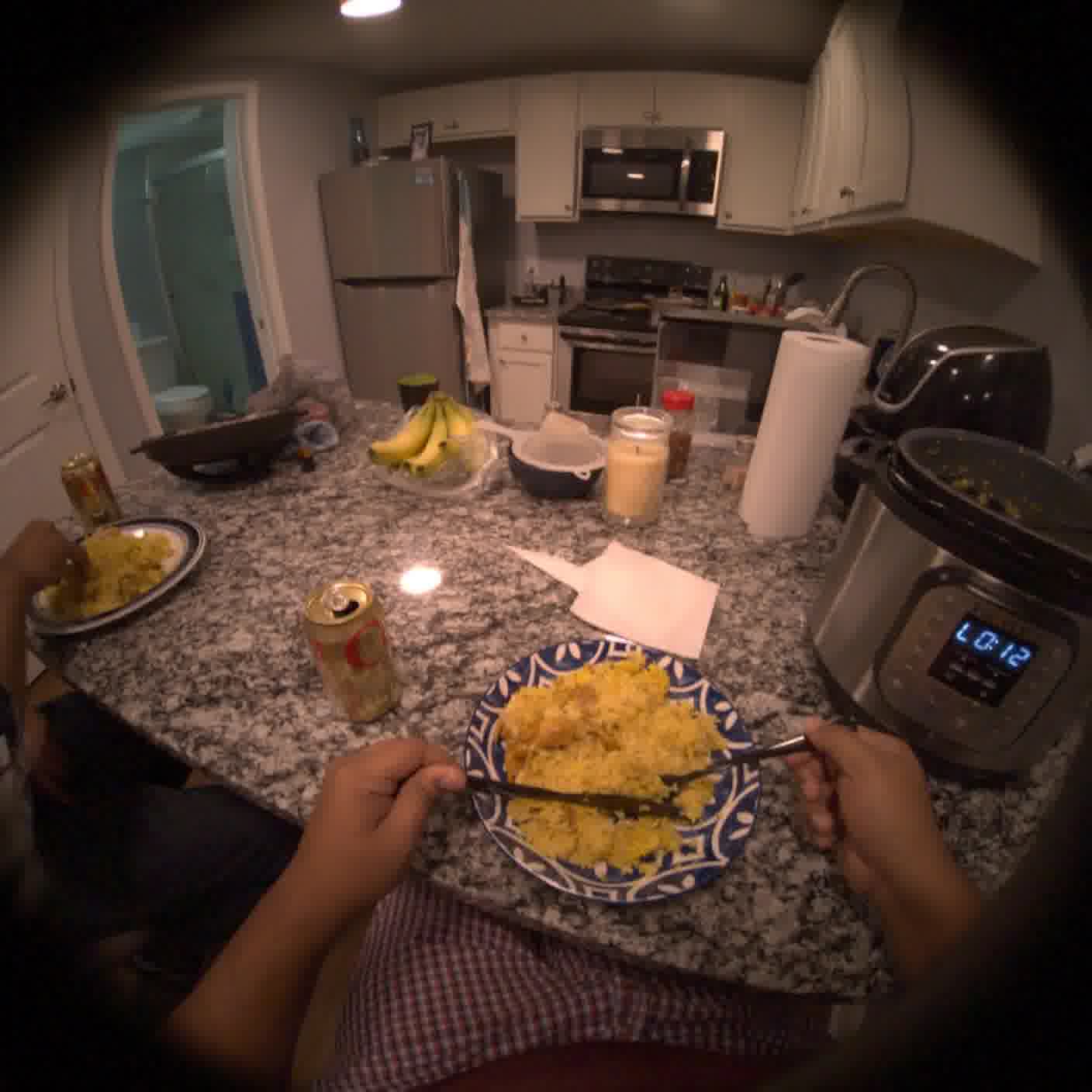}{later}
        \vspace{0.35mm}

        {\tiny\textbf{Q:} What did B say he uses his Instant Pot for when we were eating biryani?}
        \qualanswers{
            \qualchoice{\CorrectTag}{He said he uses it for cooking beef, because it makes the meat really soft compared to the stovetop.}
            \qualchoice{\VagueTag}{He said he uses it for cooking meat, because it produces a better texture than when using the stovetop.}
            \qualchoice{\WrongTag}{He said he uses it for cooking chicken, because it cooks the meat much faster than on the stovetop.}
            \qualchoice{\NATag}{This question cannot be answered.}
        }{\qualmodelselect{\CorrectTag}{\CorrectTag}{\NATag}}
    \end{qualcard}
    \end{qualcardpair}\hfill
    \begin{qualcardpair}

    \begin{qualcard}{Procedural sequence}
        \qualframe{0.24\linewidth}{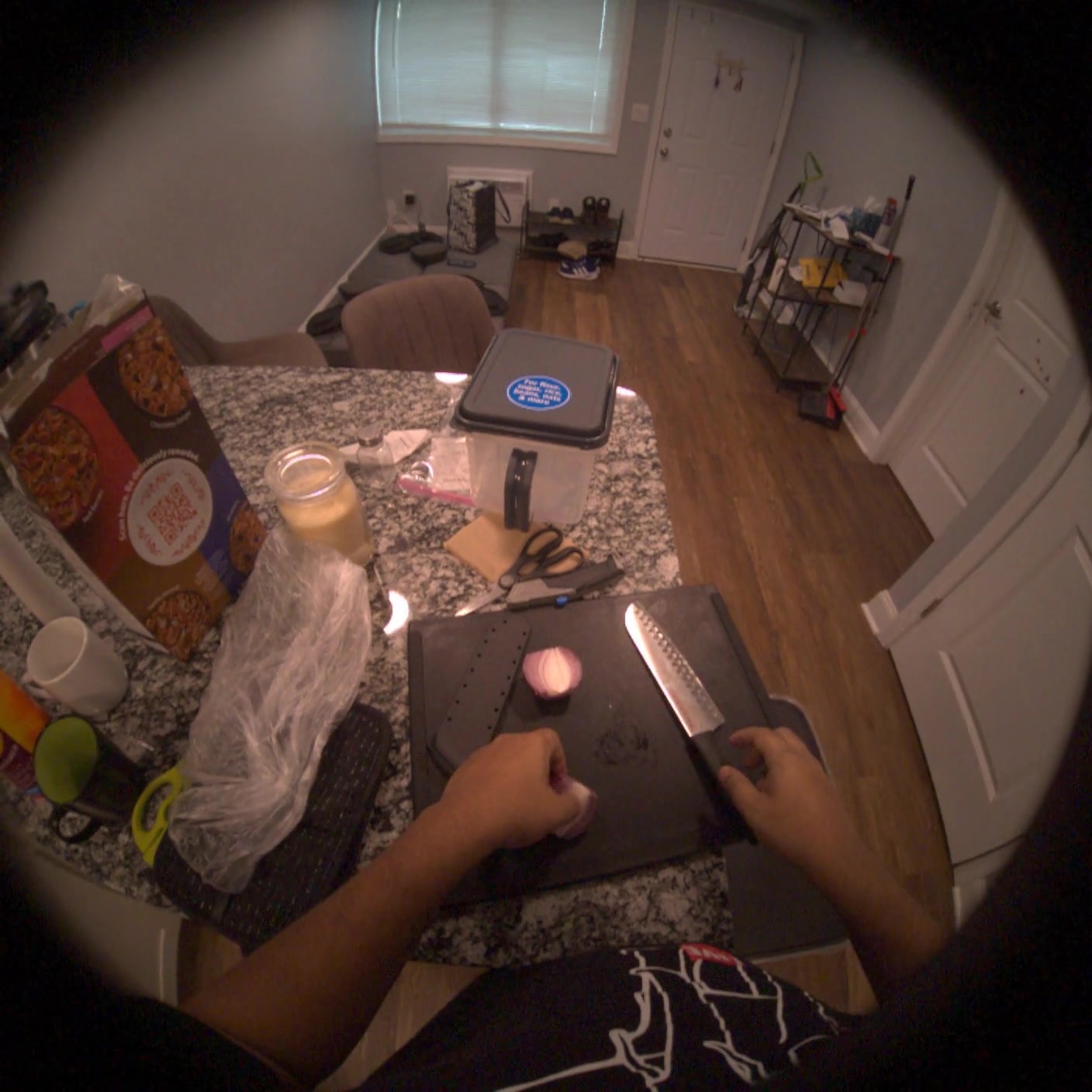}{ends}\hfill
        \qualframe{0.24\linewidth}{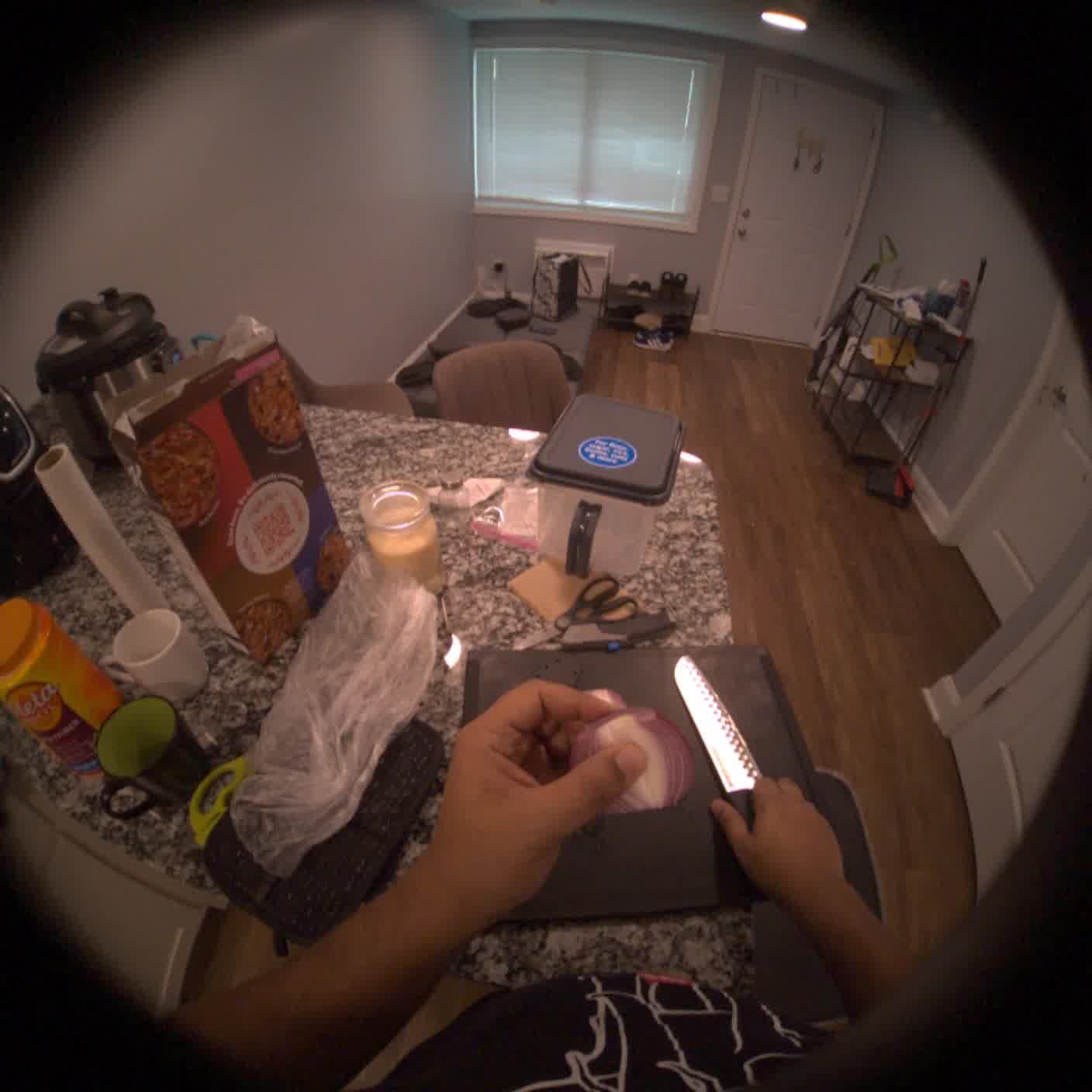}{peel}\hfill
        \qualframe{0.24\linewidth}{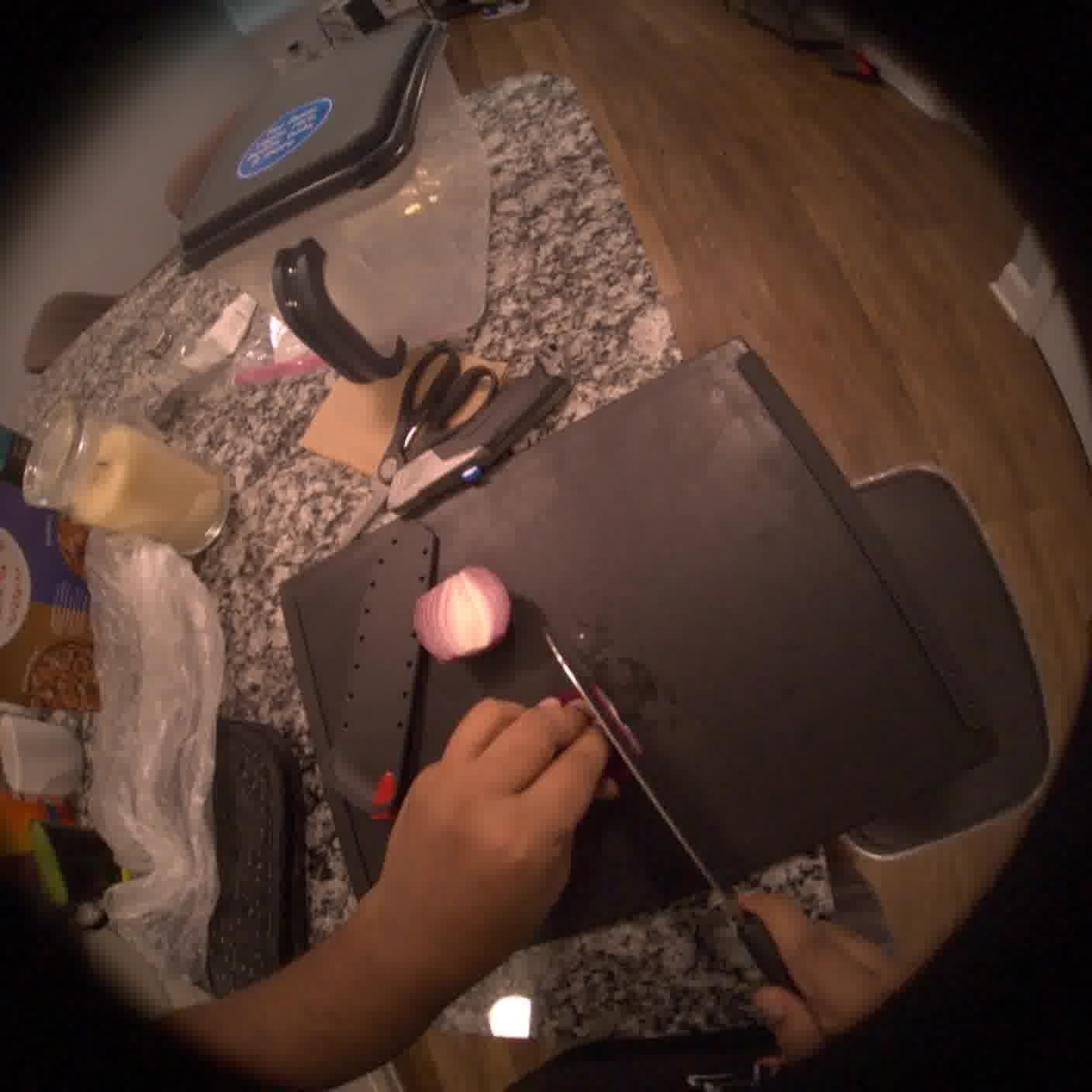}{half}\hfill
        \qualframe{0.24\linewidth}{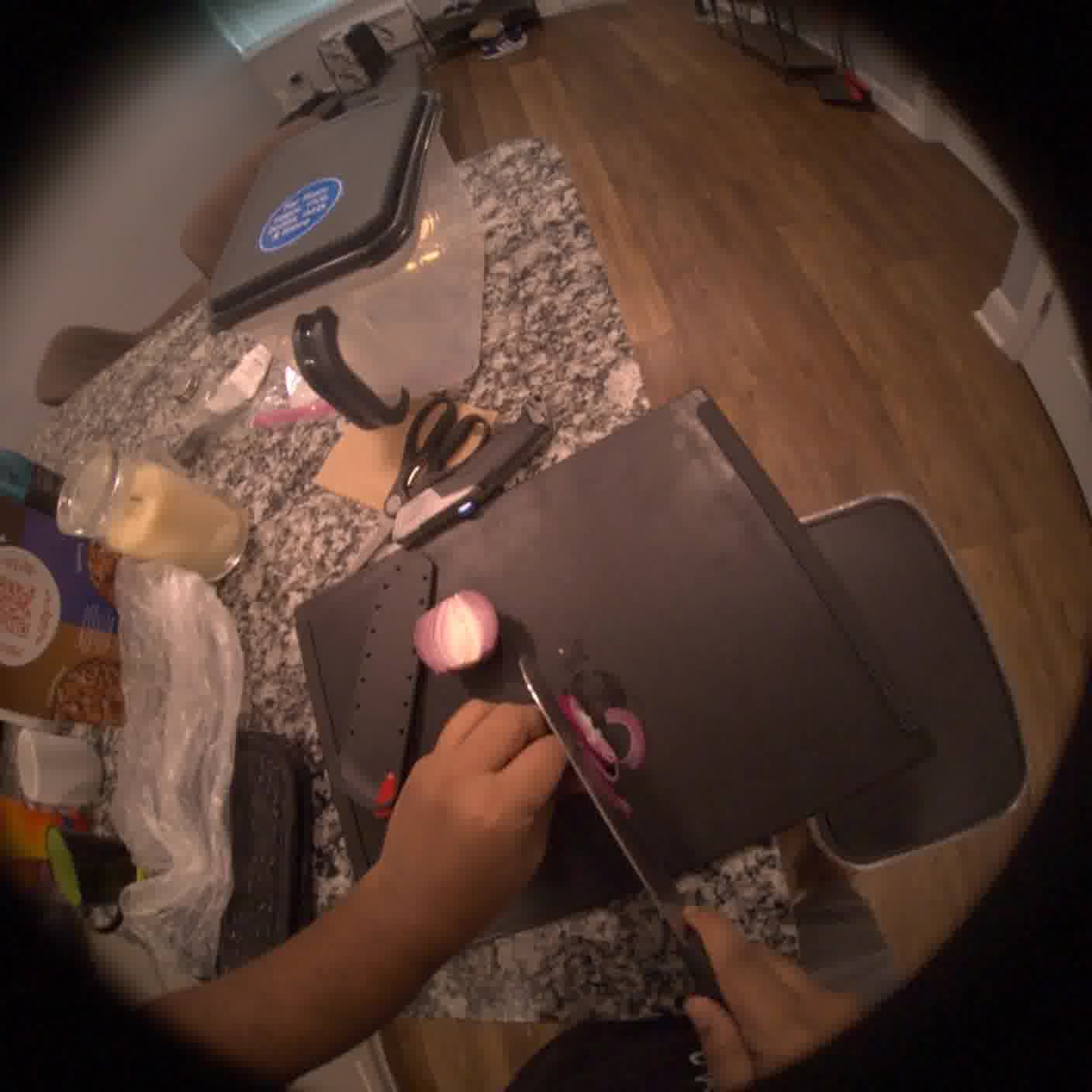}{slice}
        \vspace{0.35mm}

        {\tiny\textbf{Q:} I'm writing down the recipe steps for a friend. How exactly did I prep the red onion earlier?}
        \qualanswers{
            \qualchoice{\CorrectTag}{Ends off, outer layers peeled, onion cut in half, then sliced into thin strips.}
            \qualchoice{\VagueTag}{Ends and skin removed, then the onion was cut smaller and sliced into strips.}
            \qualchoice{\WrongTag}{Outer layers peeled first, then the ends were sliced off before cutting and slicing.}
            \qualchoice{\NATag}{This question cannot be answered.}
        }{\qualmodelselect{\CorrectTag}{\CorrectTag}{\NATag}}
    \end{qualcard}
    \end{qualcardpair}
    \caption{Additional answerable Flash and Gemini-3.1-Pro comparisons under EgoButler. Flash uses summarized conversation and procedural logs, while Gemini-3.1-Pro abstains.}
    \label{fig:qualitative_flash_pro_egobutler_answerable}
\end{figure}

\begin{figure}[t!]
    \centering
    \begin{qualcardpair}
    \begin{qualcard}{Package-label recall}
        \qualframe{0.315\linewidth}{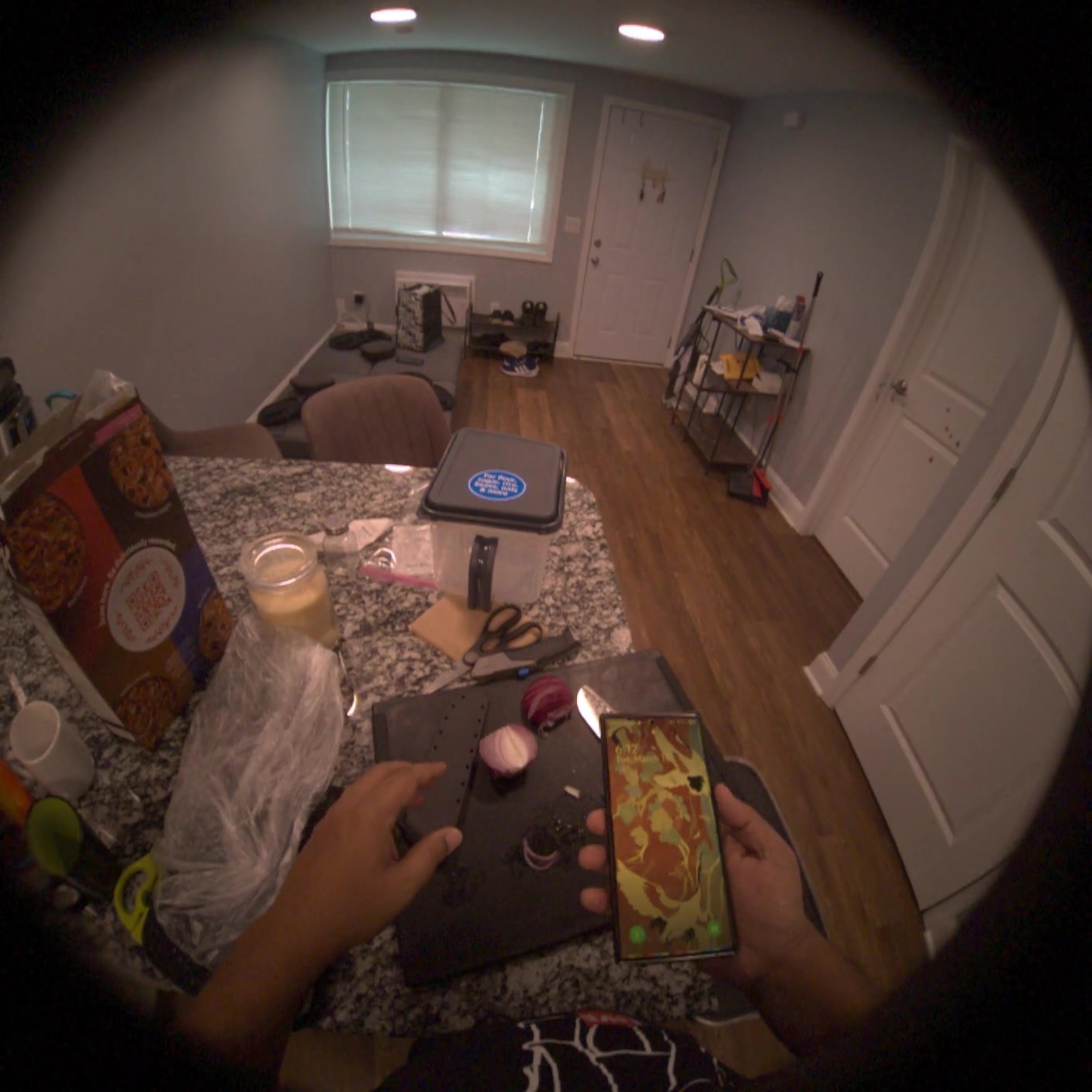}{salt}\hfill
        \qualframe{0.315\linewidth}{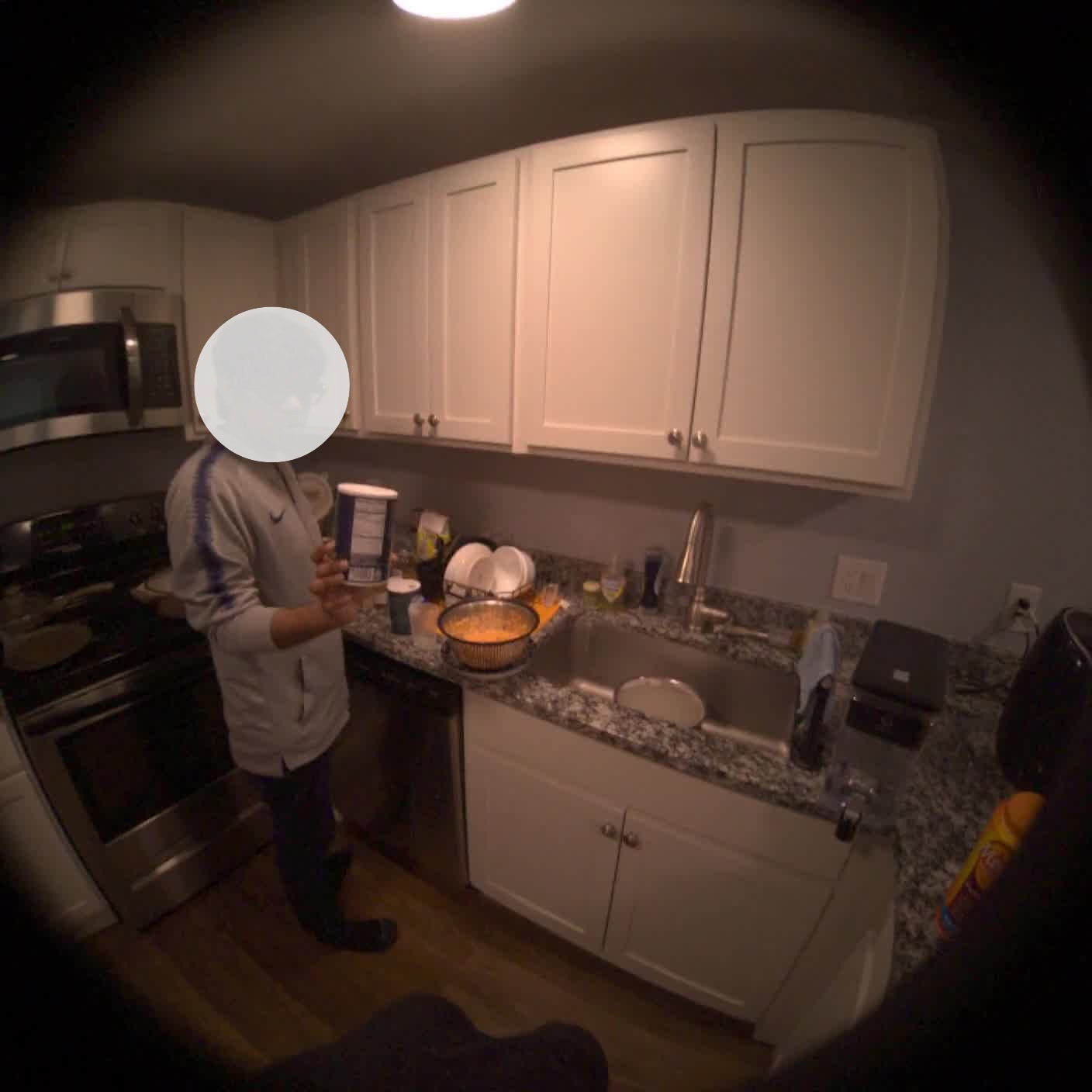}{label}\hfill
        \qualframe{0.315\linewidth}{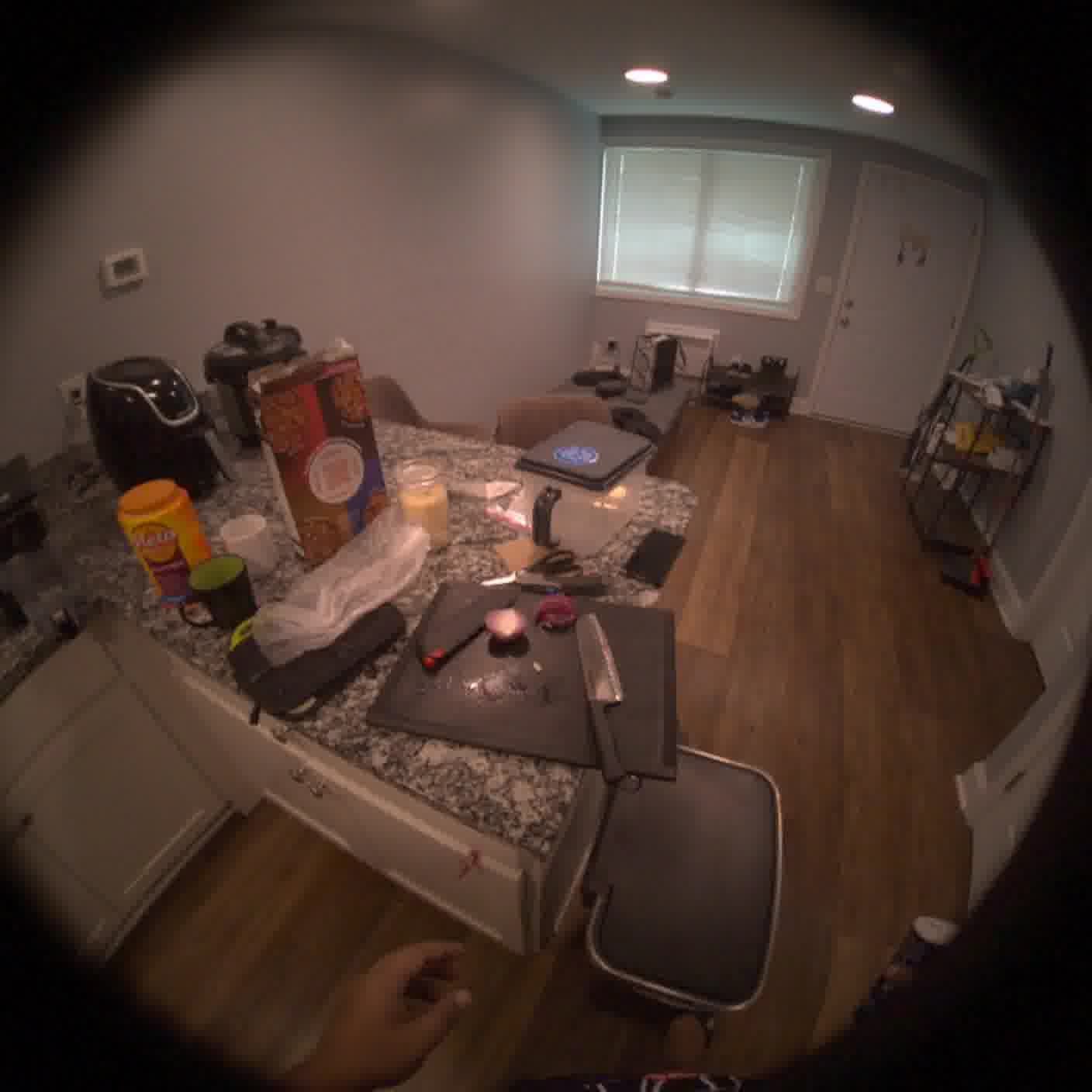}{later}
        \vspace{0.35mm}

        {\tiny\textbf{Q:} I'm checking the spices. Does the blue Morton salt container say if the salt is iodized?}
        \qualanswers{
            \qualchoice{\CorrectTag}{Yes, the blue Morton salt container indicates that the salt is iodized.}
            \qualchoice{\VagueTag}{The label on the blue container provides details about the salt type.}
            \qualchoice{\WrongTag}{No, the blue Morton salt container indicates that it is plain sea salt.}
            \qualchoice{\NATag}{This question cannot be answered.}
        }{\qualmodelselect{\CorrectTag}{\CorrectTag}{\NATag}}
    \end{qualcard}
    \end{qualcardpair}\hfill
    \begin{qualcardpair}

    \begin{qualcard}{Object placement}
        \qualframe{0.315\linewidth}{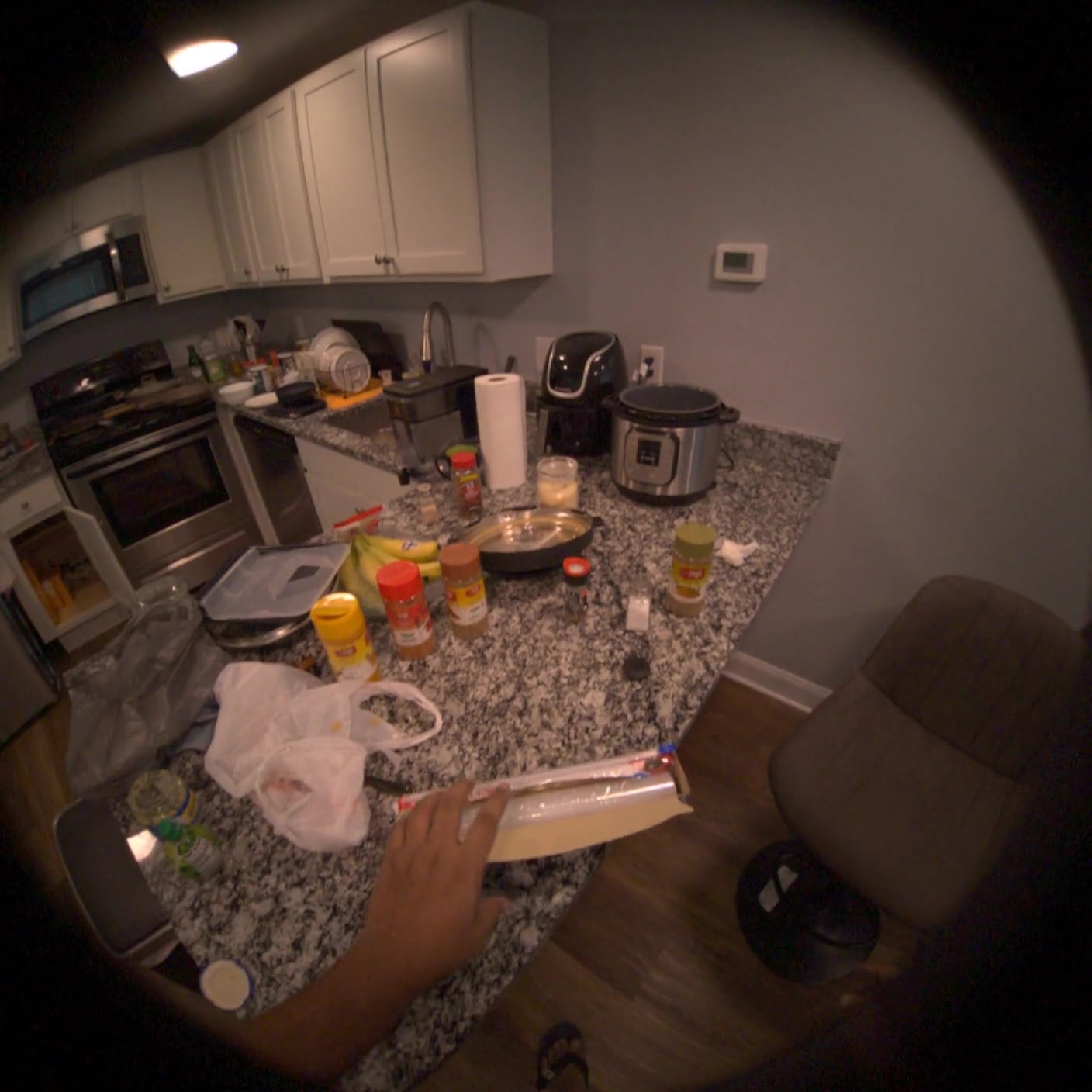}{box}\hfill
        \qualframe{0.315\linewidth}{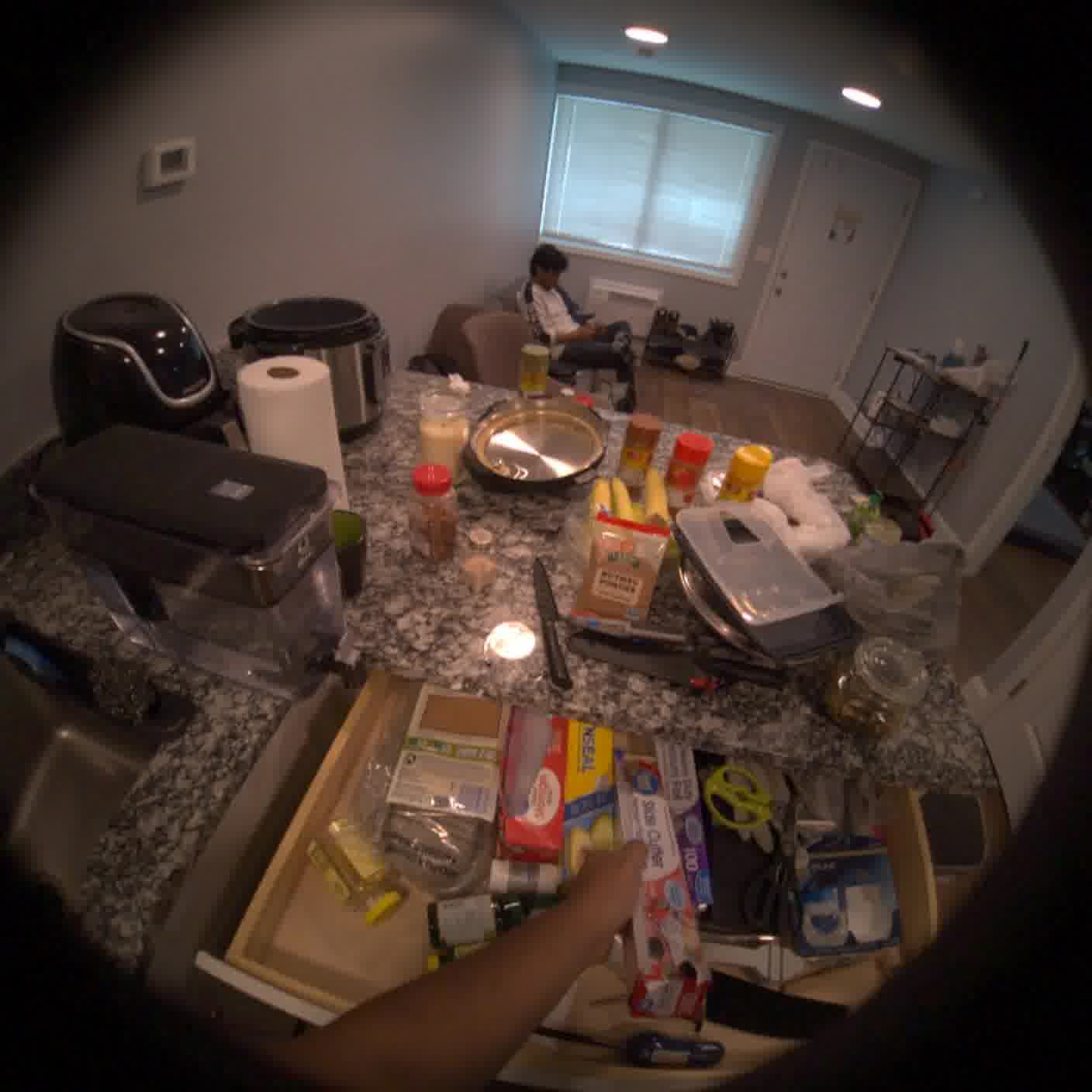}{drawer}\hfill
        \qualframe{0.315\linewidth}{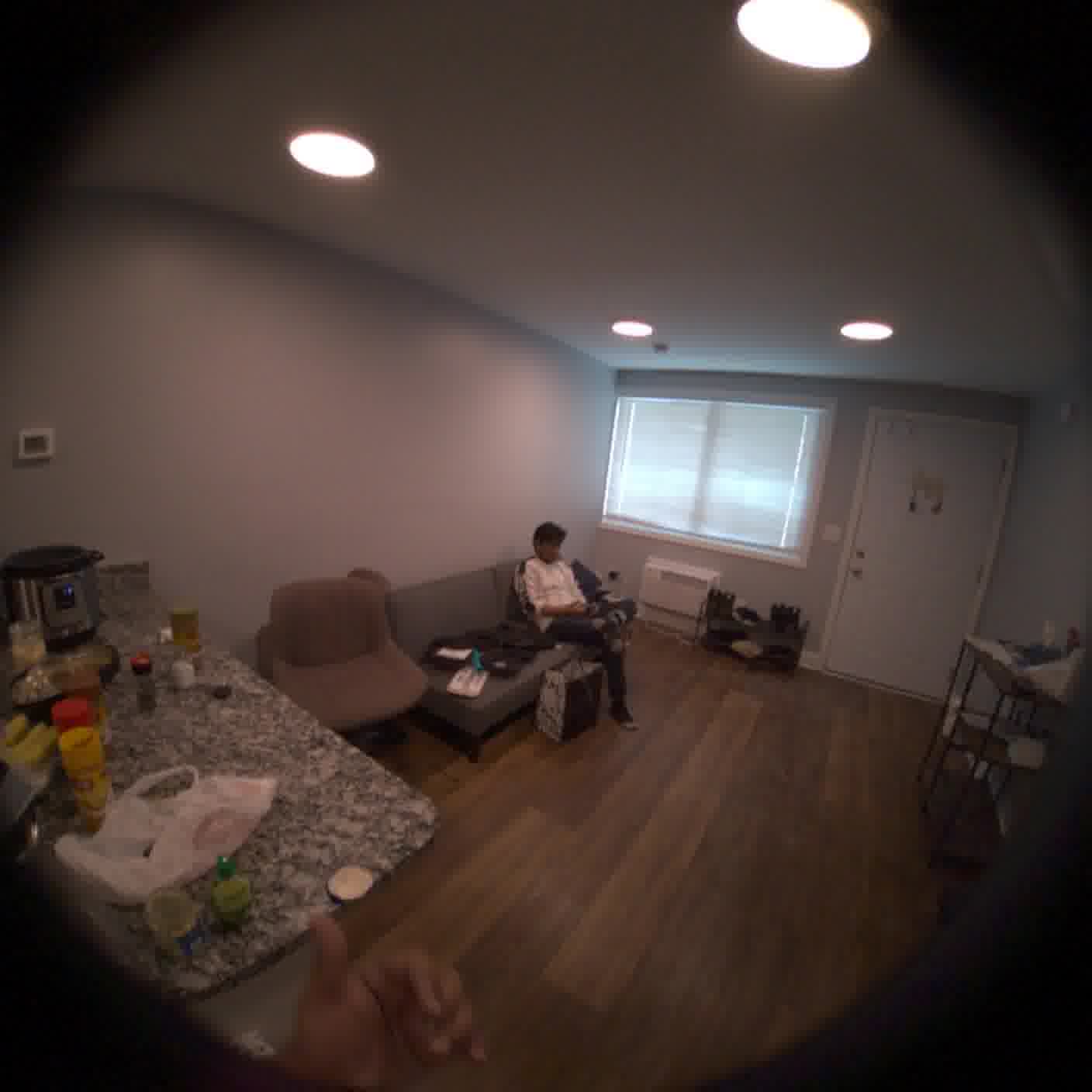}{closed}
        \vspace{0.35mm}

        {\tiny\textbf{Q:} I need to wrap up some leftovers. Where did I put the box of plastic wrap?}
        \qualanswers{
            \qualchoice{\CorrectTag}{You stored the box of plastic wrap in a lower drawer in the kitchen island.}
            \qualchoice{\VagueTag}{You put it away in one of the kitchen drawers.}
            \qualchoice{\WrongTag}{You stored the box of plastic wrap in the upper cabinet above the microwave.}
            \qualchoice{\NATag}{This question cannot be answered.}
        }{\qualmodelselect{\CorrectTag}{\CorrectTag}{\NATag}}
    \end{qualcard}
    \end{qualcardpair}
    \caption{Additional answerable Flash and Gemini-3.1-Pro comparisons under Video-RAG. Flash answers small-label and object-location questions, while Gemini-3.1-Pro abstains.}
    \label{fig:qualitative_flash_pro_videorag_answerable_more}
\end{figure}

\begin{figure}[t!]
    \centering
    \begin{qualcardpair}
    \begin{qualcard}{No visible brand}
        \qualframe{0.315\linewidth}{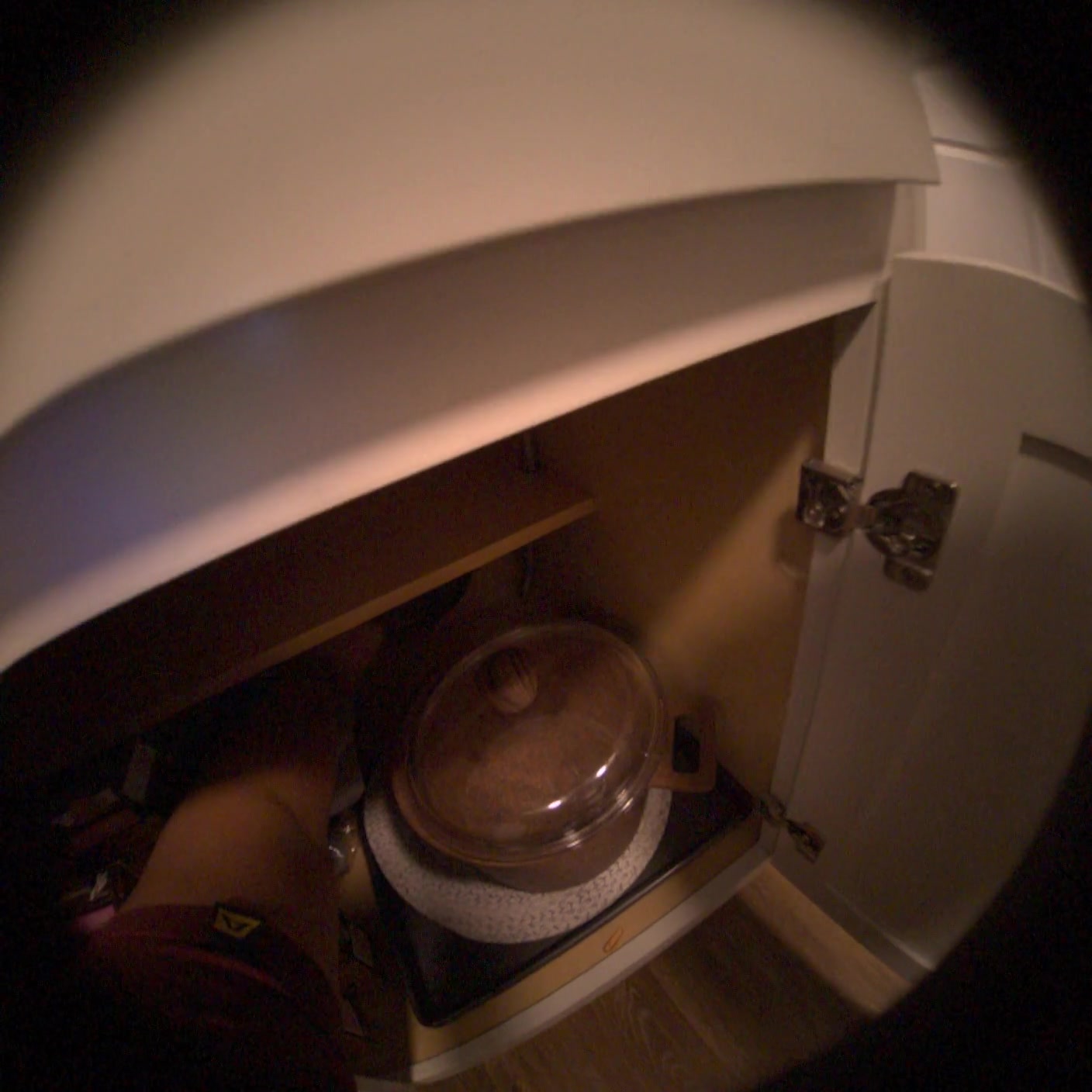}{strainer}\hfill
        \qualframe{0.315\linewidth}{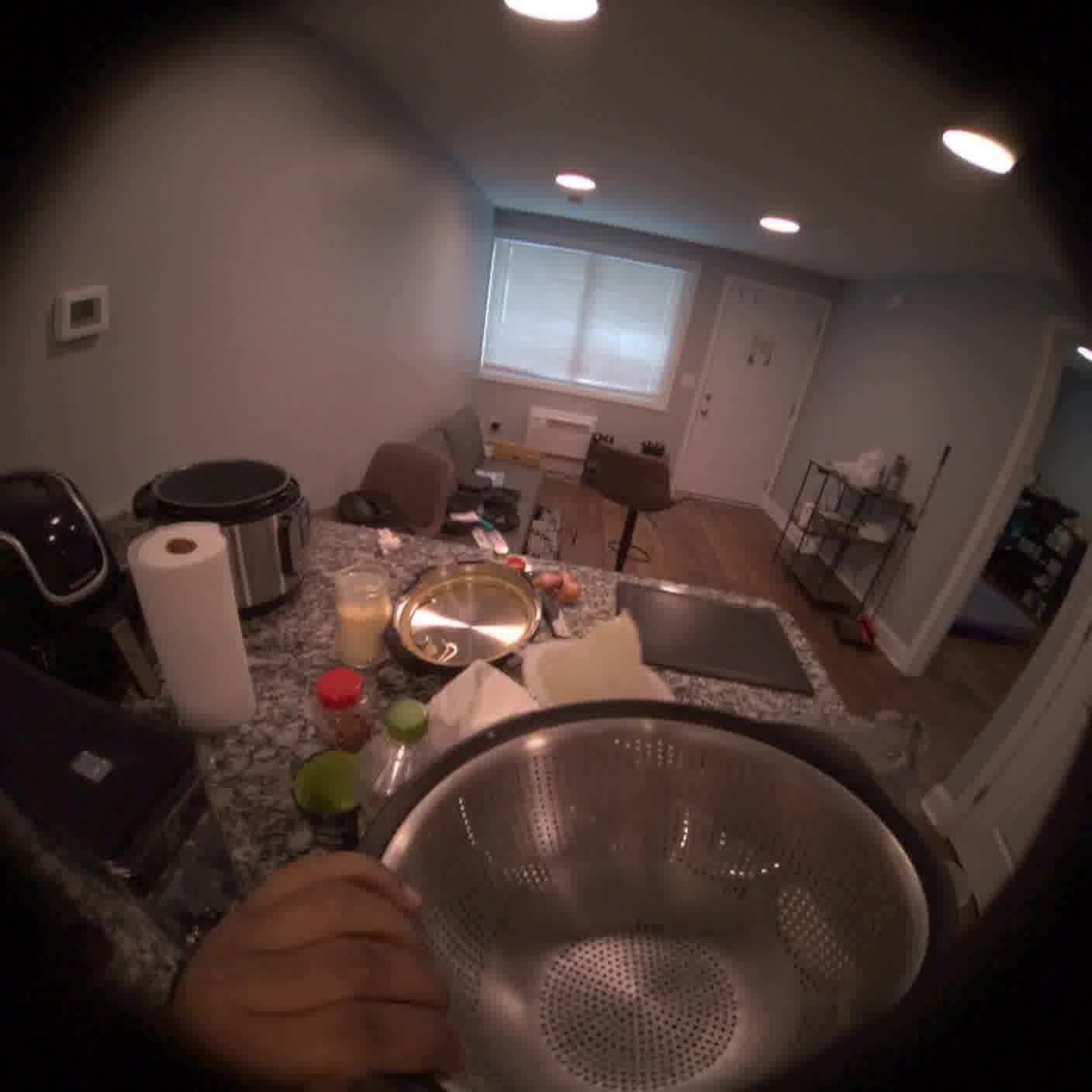}{colander}\hfill
        \qualframe{0.315\linewidth}{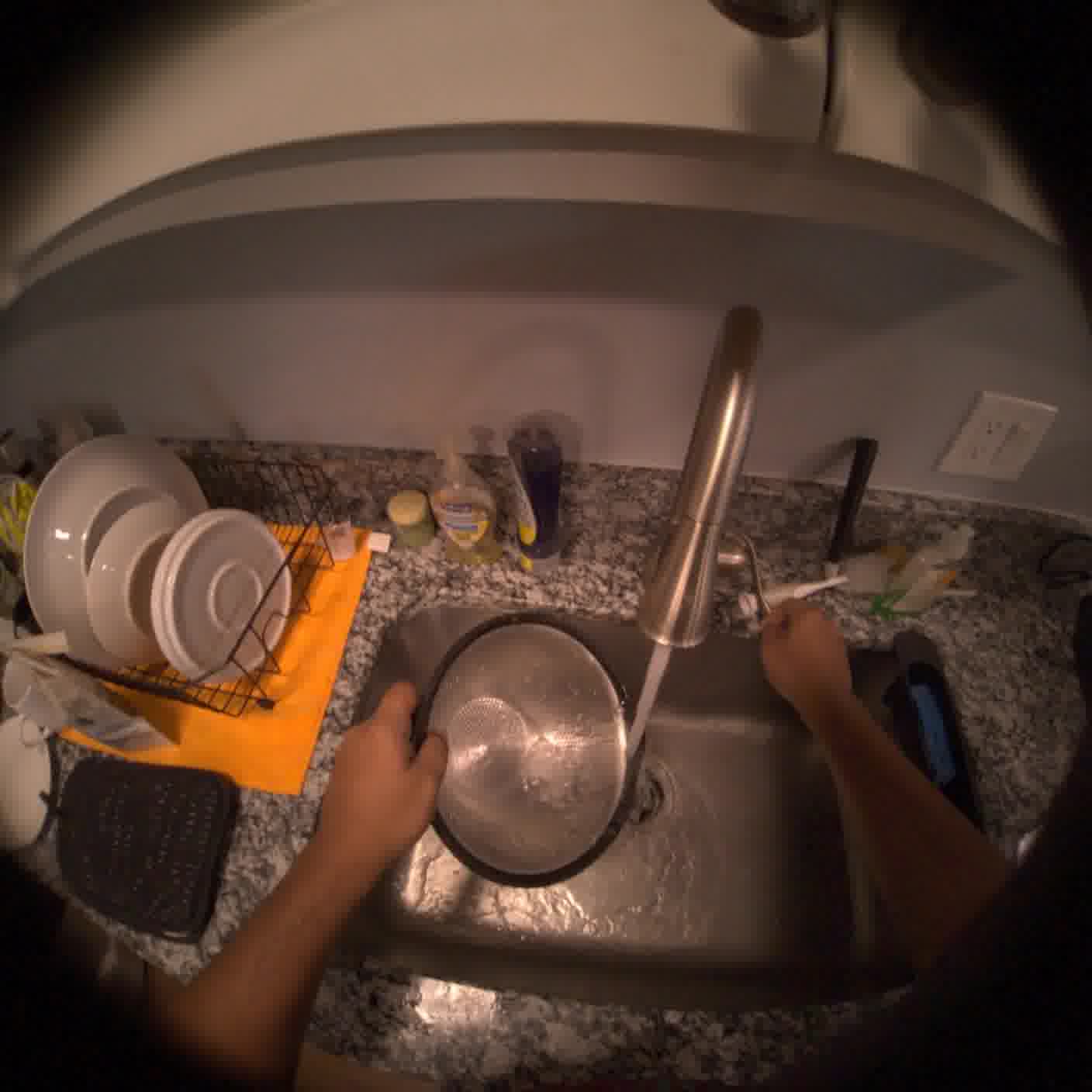}{sink}
        \vspace{0.35mm}

        {\tiny\textbf{Q:} Hey, I remember using a white mesh strainer and a metal colander earlier today. Was either of those made by Cuisinart?}
        \qualanswers{
            \qualchoice{\WrongTag}{The white mesh strainer is a Cuisinart model, but the brand of the metal colander is not visible.}
            \qualchoice{\WrongTag}{The metal colander is manufactured by Cuisinart, while the brand of the white mesh strainer is not recorded.}
            \qualchoice{\WrongTag}{Both the white mesh strainer and the metal colander are identified as Cuisinart products.}
            \qualchoice{\NATag}{This question cannot be answered.}
        }{\qualmodelselect{\NATag}{\WrongTag}{\NATag}}
    \end{qualcard}
    \end{qualcardpair}\hfill
    \begin{qualcardpair}

    \begin{qualcard}{False tool premise}
        \qualframe{0.315\linewidth}{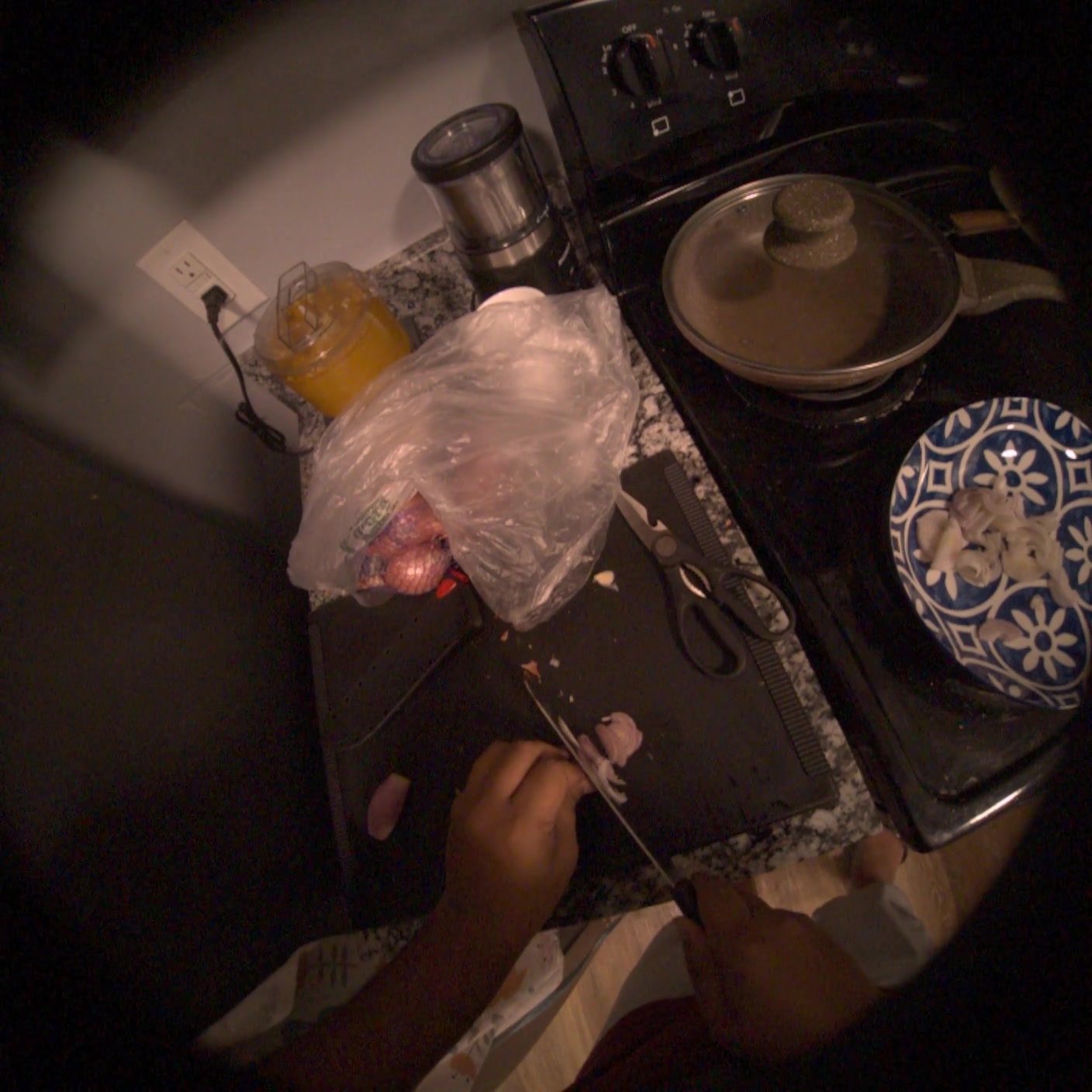}{lentils}\hfill
        \qualframe{0.315\linewidth}{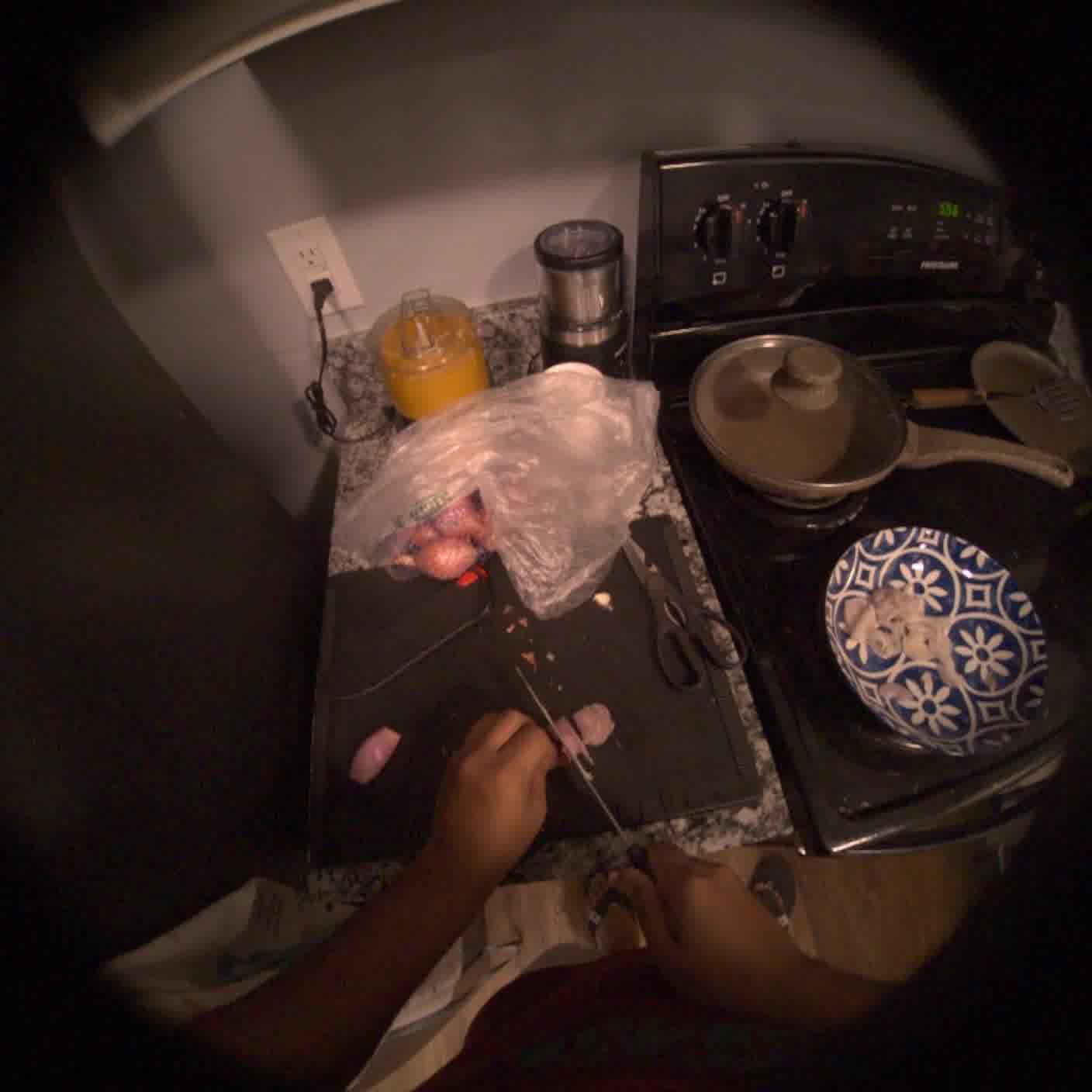}{spoon}\hfill
        \qualframe{0.315\linewidth}{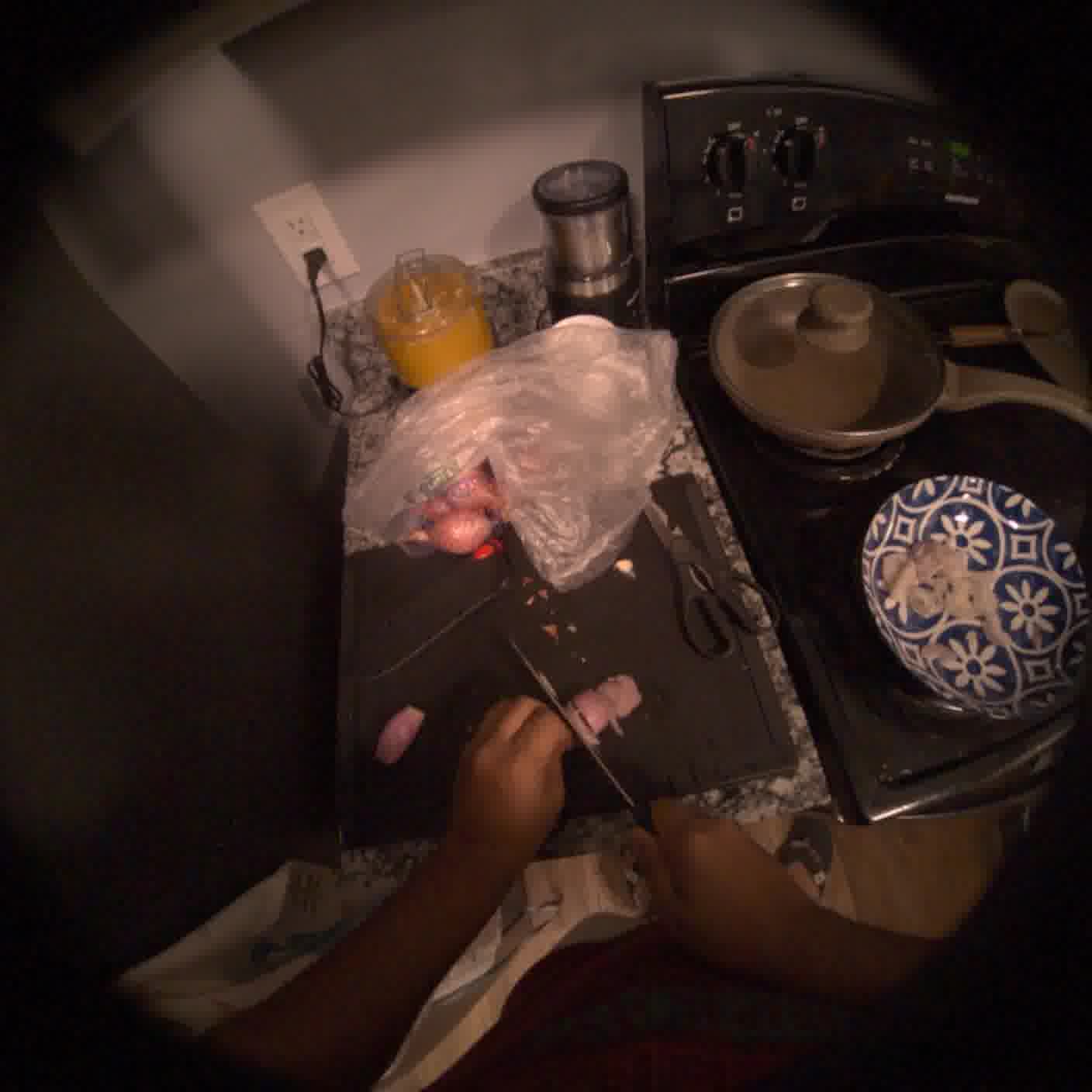}{prep}
        \vspace{0.35mm}

        {\tiny\textbf{Q:} I'm cleaning up my prep station. Where did I put the wooden spoon I used to scoop the lentils?}
        \qualanswers{
            \qualchoice{\WrongTag}{You used a silver spoon to scoop the lentils, not a wooden spoon.}
            \qualchoice{\WrongTag}{You used a white spoon to scoop the lentils, not a wooden spoon.}
            \qualchoice{\WrongTag}{You used a blue spoon to scoop the lentils, not a wooden spoon.}
            \qualchoice{\NATag}{This question cannot be answered.}
        }{\qualmodelselect{\NATag}{\WrongTag}{\NATag}}
    \end{qualcard}
    \end{qualcardpair}
    \caption{Additional unanswerable Flash and Gemini-3.1-Pro comparisons under EgoButler. Gemini-3.1-Pro rejects noisy summary evidence and false tool premises more reliably than Flash.}
    \label{fig:qualitative_flash_pro_egobutler_unanswerable}
\end{figure}

\textbf{Conversational fact conflation.}
Conversational memory can fail when nearby context provides a tempting but incorrect paraphrase. In these cases, the model often retrieves the right conversational neighborhood but collapses distinct statements into a broader summary. This is especially problematic for memory assistance because users often ask about small conversational commitments, preferences, or corrections. A vague summary may be semantically related, but it is not the remembered fact the user needs.

This pattern reveals a tension between compression and fidelity. Long-horizon systems need summarization to scale, yet conversational memory frequently depends on preserving exact attributes: who said what, which object was being compared, and whether a statement was a correction or a new fact. The errors here are not random hallucinations; they are plausible substitutions drawn from adjacent context. That makes them hard to catch unless the benchmark distinguishes correct answers from vague or misleading alternatives.

\textbf{Small visual evidence and OCR-style failures.}
Fine-grained visual memory is fragile when the answer depends on a small object, a short-lived view, or text that occupies only a small region of the frame. \Cref{fig:qualitative_visual_failures} shows two such cases. The important observation is that failure can take two different forms: conservative systems abstain even though evidence exists, while less conservative systems may answer with a visually plausible but incorrect detail. Both are undesirable for an AI memory assistant, but they have different user-facing costs.

These examples also show why egocentric memory cannot be reduced to generic video captioning. A captioner may accurately describe the overall scene while omitting the small object or tiny text that later becomes decisive. Once that information is absent from the memory representation, downstream retrieval has little chance to recover it. Robust systems likely need adaptive high-resolution inspection, object permanence over brief interactions, and OCR that can be triggered by later user intent rather than only by frame-level salience at indexing time.

\textbf{Temporal ordering and state-change tracking.}
Long-horizon memory requires preserving order or object state and activity, not merely retrieving a visually similar frame. \Cref{fig:qualitative_temporal_reasoning} contrasts a sequencing failure with a state-change success. The hard part is that the relevant frames can look repetitive: similar objects, similar hand motions, and similar locations recur across the session. A memory system must therefore maintain an event-level trace that records transitions rather than a bag of matching visual observations.

The state-change case illustrates what success looks like: the system must link an object's earlier use to a later cleaning event and then to its final storage location. This is closer to episodic state tracking than ordinary visual recognition. The sequencing failure, by contrast, shows that even when all evidence is present, the model may not preserve the ordering relation with enough confidence to answer. These failures point toward architectures that explicitly represent temporal predicates such as before, after, moved-to, cleaned, stored, and still-missing.

\textbf{Quantifier precision.}
Repeated near-identical actions expose counting failures. The issue is not only detecting an object category, but deciding whether multiple observations correspond to the same item, repeated handling of one item, or distinct instances. This is a common daily-memory need: users ask whether they already added an ingredient, how many pieces were packed away, or whether a repeated action happened once or several times.

The qualitative count failures suggest that current memory pipelines do not reliably maintain instance identity across repeated actions. Summaries tend to compress repeated visual events into phrases such as ``some items'' or ``several actions,'' which are useful for gist but insufficient for exact answers. A stronger memory assistant would need explicit count-preserving representations, uncertainty over duplicate observations, and mechanisms for reconciling repeated views of the same object.

\textbf{Abstention and premise validation.}
The unanswerable option is essential when the event is partially visible but the requested detail is missing, or when the question contains a false premise. \Cref{fig:qualitative_abstention} shows both cases. These examples are important because they separate retrieval from epistemic validation: finding a related visual moment is not enough. The model must also verify that the specific requested fact is supported.

False-premise questions are especially revealing. A system may retrieve the named object and then answer as though every clause in the query were true. From the user's perspective, this is a hallucination even if the object itself was seen. Correct abstention therefore requires checking the full proposition expressed by the question, including modifiers, causal claims, and implied actions. This is why \datasetName includes an explicit unanswerable option rather than treating all memory queries as answerable lookup problems.

Taken together, these qualitative cases show that long-horizon AI memory requires a coupled solution: retrieval must be sensitive enough to surface sparse details, memory representations must preserve exact attributes and event order, and the answerer must determine what is and is not supported by the evidence. Aggregate accuracy alone hides these distinctions. The qualitative breakdown makes clear that future progress will likely require explicit memory structures for object state, counts, temporal relations, and premise validation, in addition to stronger vision-language models.

\textbf{Gemini-3-Flash versus Gemini-3.1-Pro.}
The results expose a sharper model-level trade-off between Gemini-3-Flash and Gemini-3.1-Pro under the same retrieval setting. \Cref{fig:qualitative_flash_pro_answerable} shows answerable cases where Flash is willing to use retrieved evidence, while Gemini-3.1-Pro abstains despite support being present. In the Monopoly arithmetic case, the model must retrieve two distinct rent events and compute the difference. In the card-text case, the model must read a small piece of game text and preserve the exact wording. These examples are consistent with the aggregate result that Flash often obtains higher QA accuracy by committing when evidence is available.

The same tendency reverses on false-premise questions. \Cref{fig:qualitative_flash_pro_unanswerable} shows cases where the retrieved context is topically related but does not support the proposition in the question. Here, Flash produces specific but unsupported answers, whereas Gemini-3.1-Pro correctly abstains. This suggests that Gemini-3.1-Pro's conservatism can be useful when the evidence is missing or the prompt smuggles in an unverified event, even though the same conservatism hurts answerable recall.

In addition, \Cref{fig:qualitative_flash_pro_egobutler_answerable} shows answerable EgoButler cases where Flash uses conversational and procedural summaries while Gemini-3.1-Pro abstains. \Cref{fig:qualitative_flash_pro_videorag_answerable_more} shows the same answerable-recall pattern under Video-RAG for small-label reading and object placement. \Cref{fig:qualitative_flash_pro_egobutler_unanswerable} shows the complementary unanswerable setting, where Gemini-3.1-Pro's caution helps reject noisy summaries and false tool premises. Together, these additional examples show that the Flash/Gemini-3.1-Pro trade-off persists across retrieval formats. Flash is often better matched to answerable retrieved evidence, especially when the task requires committing to small visual text, conversational facts, object placement, arithmetic over events, or procedural order. Gemini-3.1-Pro is more cautious and can therefore lose recall, but that caution protects against false positives when the evidence does not support the user's premise.

\FloatBarrier

\subsection{Survey Results}
\label{sec:results.survey}

\begin{figure}[h!]
    \centering
    \includegraphics[width=0.6\linewidth]{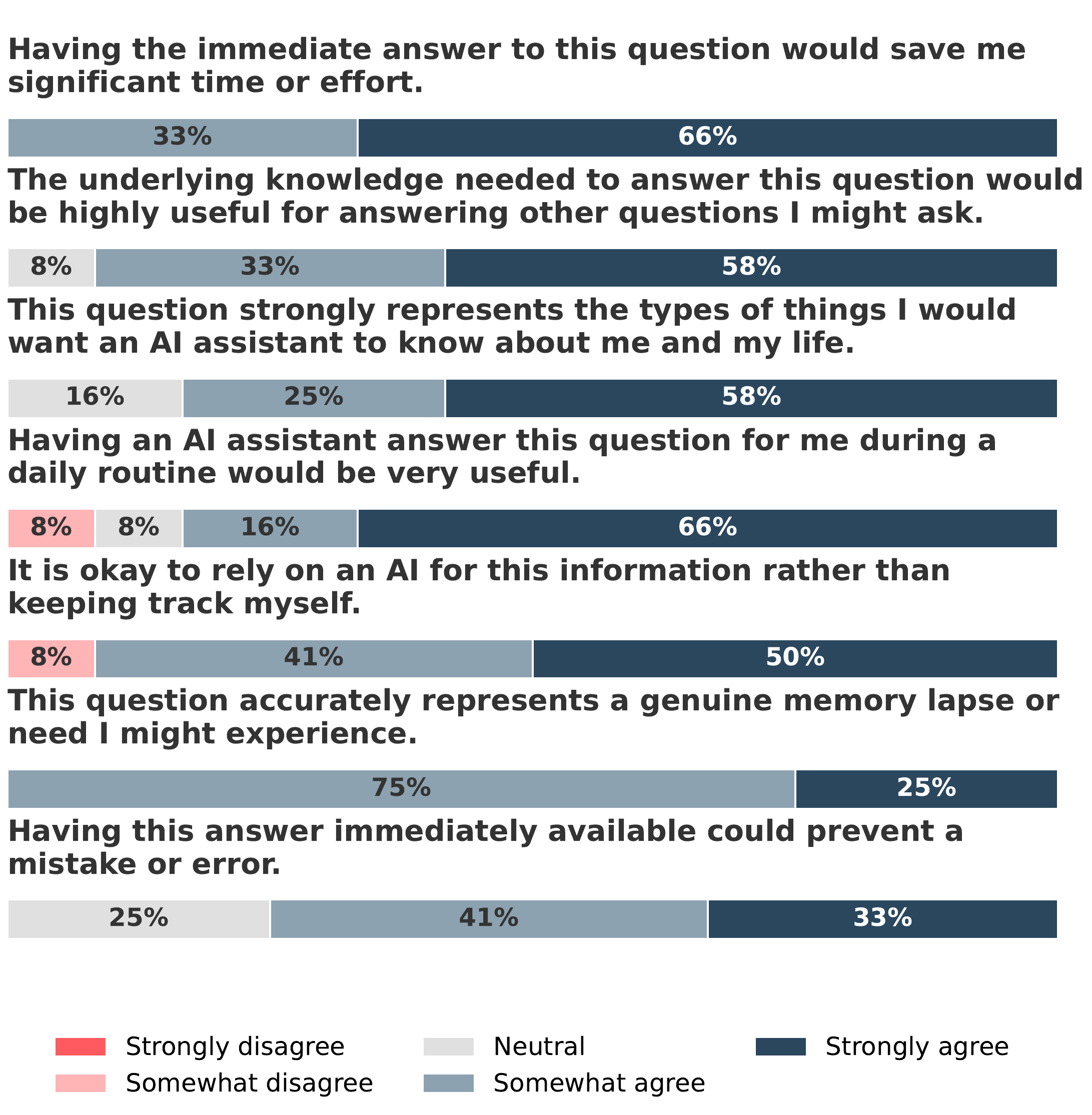}
    \caption{Survey responses for QA quality.}
    \label{fig:survey}
\end{figure}

We conducted a participant survey with eight participants to measure perceived question realism, usefulness, and alignment with everyday memory needs. Users reviewed $18$ questions from their own recordings and rated seven statements about question quality and utility. As shown in \Cref{fig:survey}, responses were strongly positive, with no disagreement across statements: 86\% agreed that questions captured genuine memory lapses, and 82\% found the answers useful during daily routines.
The survey also suggests that \datasetName captures reusable personal knowledge rather than isolated QA pairs: 78\% agreed that the underlying knowledge would also help answer future questions. This pattern supports the broader motivation of \datasetName: long-horizon AI memory should be useful, personal, and grounded, while still requiring careful treatment of user trust and control.

\FloatBarrier

\section{Conclusion}
We introduced \datasetName, an egocentric VQA benchmark for long-horizon memory in AI assistant settings. It combines multimodal AR-glass recordings with temporally grounded questions, evidence annotations, and ordered answer choices that distinguish accurate, vague, incorrect, and unanswerable responses.
Our preliminary evaluation shows that current video understanding and retrieval-augmented systems remain unreliable: strong VLMs still struggle with answerability, long temporal gaps, and multi-moment evidence integration. Participant feedback confirms practical daily relevance. Overall, \datasetName moves evaluation toward grounded, situated memory systems and highlights the need for accurate, hallucination-robust models that answer only when the available evidence is sufficient.

{
  \small
  \bibliographystyle{plainnat}
  \bibliography{main}
}

\appendix
\raggedbottom

\newpage
\section{Task Description Supplementary Information}
\label{app:tasks}

\subsection{Task Taxonomy}
\label{sec:tasks}
This section expands the task taxonomy introduced in \Cref{sec:dataset} and compared against prior datasets in \Cref{sec:comparison}.

The \datasetName dataset categorizes question-answer pairs into six distinct functional tasks. Each task mimics a real-world application of human memory augmentation. Below, we describe these tasks.

\vspace{1mm}
\textbf{Object \& Location Memory.} This task requires the system to identify the last known position of a specific object or track its movement across different time spans and locations, maintaining object permanence. This task is based on the concept of episodic memory~\citep{tulving1972episodic} and has been previously used in egocentric vision \citep{grauman2022ego4d}. It evaluates an AI agent's ability to index a wearer's data and locate an object based on its last observed time. This effectively augments human memory on demand, answering queries such as, ``I cannot find the black scissors. Where did I leave them last?'' or ``I forgot where I put my keys. Where did I put them?''.

\vspace{1mm}
\textbf{Conversational Memory.} Systems must also recall specific facts from multi-topic chats, including tracking commitments, deferred answers, and mid-conversation corrections. Conversational Memory, also known as Dialogue State Tracking (DST), is a formal component within task-oriented conversational AI \citep{henderson2014word, wu2019transferable}. Robust DST is critical for maintaining awareness across exchanges without losing context due to natural conversational shifts. Examples of this task include queries like, ``Where did I say we should meet after this?'' or ``What did B tell me to get after I finished working on this?''.

\vspace{1mm}
\textbf{Visual Scene Recall.} This task evaluates the ability to recall specific visual details from past environments, such as text on whiteboards, instructions in a manual, or displayed on a screen. While rooted in episodic memory~\citep{tulving1972episodic}, it concurrently evaluates the system's \emph{semantic memory}~\citep{tulving1972episodic, squire2004memory}. To successfully manage dense information retrieval, the AI must go beyond visual recall to recognize objects, read text, and link these elements to general world knowledge and factual meaning independent of the temporal context. Representative questions include: ``Was I supposed to add the spices now or after frying?'' or ``I forgot how many cups of rice I put in the pot. How much did I put in?''.

\vspace{1mm}
\textbf{In-Context Retrieval.} This task requires the system to perform multi-hop reasoning by chaining multiple disjoint facts retrieved from the user's history. It evaluates the system's capacity for \emph{relational memory}~\citep{eichenbaum1997declarative, cohen1997memory}—the cognitive ability to represent and navigate associations between independent elements of an experience. Unlike simple retrieval, in-context retrieval requires the AI to identify a primary fact (e.g., a mentioned time or location) and use it as a prerequisite context to evaluate or retrieve a secondary piece of information (e.g., the status of an ongoing task). This task models complex real-world situational awareness and planning where memory serves as a substrate for logical deduction. Example queries include, ``Given the meeting time B mentioned, do I have time to finish my presentation?'' or ``My bus leaves in 20 minutes. Will the dryer finish up in time?''.

\vspace{1mm}
\textbf{Timeline Reconstruction.} The agent is required to sequence events chronologically, such as listing all locations visited during a multi-location errand in the correct order. Timeline reconstruction directly evaluates the temporal aspect of episodic memory~\citep{tulving2002episodic} through the temporal localization and chronological sequencing of disjointed events across a longer horizon. Furthermore, when applied to structured activities, this task models the tracking of \emph{procedural memory}~\citep{cohen1980preserved, squire2004memory}---the implicit knowledge of how to perform tasks and specific action sequences. By reconstructing timelines, the agent demonstrates an understanding of the step-by-step procedures required to complete an activity. For instance, the system might be asked, ``Did I miss any ingredients when preparing the marinade?'' or ``Does anyone own this property on the Monopoly board?''.

\vspace{1mm}
\textbf{Intent Recall.} Systems must retrieve information based on explicit, verbal reminders or evaluate the passive recall of goals that were implied but ultimately not completed by the user. This aligns directly with \emph{prospective memory}, the cognitive process that allows individuals to remember to perform intended actions in the future \citep{einstein1990normal, brandimonte1996prospective}. Prospective memory acts as a bridge between intentions and actions, requiring continuous monitoring to execute tasks at the appropriate time. Example queries include explicit intent like, ``What did I intend to do after I finished this?'' or implicit reminders like ``Washer cycle is done. Remind user to put the clothes into the dryer.'' Reminders can be tied to a specific point in time or be triggered based on location or interaction with a person.

To validate that our task categories reflect distinct memory behaviors, we asked participants to answer a representative sample question from each task and describe the thought process they would use if answering from memory. The resulting mapping between task categories and recalled reasoning strategies is summarized in \Cref{fig:skill_answer_sankey}. The flows show that the taxonomy is not only based on surface question wording: different tasks elicit different memory operations, such as locating an object, reconstructing temporal order, checking prior intent, or integrating evidence across events. This supports our categorization of \datasetName tasks as distinct forms of long-horizon egocentric memory.

\section{Data Collection Supplementary Information}
\label{app:data_collection}
\label{sec:data_collection}

\subsection{Hardware and Modalities}
\label{app:data_collection.hardware}

Participants wore Gen 1 Meta Aria Glasses to capture high-frequency, egocentric multimodal data. The collected raw data streams included one $1408\times1408$ RGB video stream collected at $30$ fps, two $640\times480$ grayscale video streams for SLAM collected at $30$ fps, a $320\times240$ eye-tracking camera array operating at $60$ fps to monitor gaze behavior, and $7$ channels of high-fidelity audio sampled at $48$ kHz. Additionally, the device logged motion and orientation data via two Inertial Measurement Units (IMUs) operating at $1000$ Hz and $800$ Hz, a Magnetometer at $10$ Hz, and a Barometer at $50$ Hz. We also computed open and closed loop trajectories, as well as 3D point clouds of the entire space from SLAM and IMU motion data. Eye tracking data were also obtained from processing the raw eye tracking camera streams.

\subsection{Protocol}
\label{app:data_collection.protocol}
Data collection was conducted under a protocol approved by the Institutional Review Board (IRB). The released dataset described in this paper contains recordings from \numberOfParticipants participants recruited from the general population and university settings. The IRB protocol frames the study as an evaluation of the memory capabilities of an AI-driven wearable system, ``Supermemory,'' that integrates multimodal sensor data from Meta Aria Glasses with long-term memory retrieval. Unlike studies that focus on perception within a fixed timespan, the protocol targets dynamic, non-linear memory formation across everyday activities over longer time spans. Using synchronized high-frequency eye-tracking, audio, IMU, RGB, and SLAM video data, the study analyzes how contextual and temporal cues influence recall, intent tracking, and memory construction. The protocol allowed either single-subject recordings or multiple participants working together and conversing while wearing the same type of AR glasses.

Potential participants were identified through recruitment materials and contacted the study team by email if interested. The study team confirmed eligibility before scheduling sessions, shared the types of tasks participants would perform, discussed whether participants were comfortable with outdoor tasks or tasks involving other participants, and noted preferences about study locations and timing. During screening, participants were also told how the data would be processed and that residual identifiability could remain even after processing.

Participants completed an informed consent process before any study procedures using a legally valid electronic signature workflow. After eligibility was confirmed and a session was scheduled, the study team emailed a DocuSign link to the consent form. Participants were encouraged to review the consent form in advance and could ask questions by email, phone, or in person before signing. At check-in, the study team confirmed the participant's identity and completion of the correct consent form; a copy of the signed consent was automatically provided through DocuSign. The consent process emphasized that participation was voluntary, that participants could decline or discontinue at any time without penalty, and that compensation was not contingent on completing tasks beyond the prorated time completed. Participants were also informed that participation required consent for face-blurred study data to be shared publicly through the Hugging Face Hub; individuals unwilling to allow public release of their processed data were asked not to participate.

Data were collected in a short-term rental (Airbnb) to simulate a home environment, with both indoor and outdoor segments included under the IRB and non-participant privacy protocol. Each recording session had at least one member of the lab participating in it to guide the workflow with one to three other external participants. All participants calibrated their glasses before starting to record. Participants were then asked to explore the environment and take quick notes about object locations. After that, the tasks were explained to them based on a predefined script, though they were encouraged to follow the script loosely to keep all interactions natural. Scripts included cooking instructions for preparing a recipe, manuals for playing a board game, or assembling a puzzle. Participants read or were orally dictated the specific instructions by the lab member and were asked to follow them. Participants planned among themselves how to implement the script and collaboratively finish it. To prevent leaks of sensitive information, participants were assigned code names and used those names instead of their real names. After $30$ to $45$ minutes, participants took breaks and then changed glasses before continuing.

The protocol allowed follow-up sessions using the same general procedure with different scripts, either to collect additional task permutations or to replace unusable data. Follow-up participation was voluntary, and participants could decline further sessions. Participants could incur parking or transportation costs, which were not reimbursed. Incentives were provided as physical Amazon gift cards and were prorated when participation was incomplete. Researchers who collected protocol-testing data as study subjects did not receive incentives.

The protocol describes the study as minimal risk and non-invasive. The main physical risks were slight fatigue from wearing the glasses or performing scripted tasks, mitigated by breaks between scripts. The Aria glasses include onboard temperature sensors and shutoff behavior for overheating, and attendants monitored glass temperature during sessions. The protocol identifies no direct benefit to individual participants, but describes societal benefits from enabling future AR systems that support long-term episodic recall and situational awareness and from providing an open dataset to accelerate research in human-centered AI. Primary endpoints include recall accuracy measured by automated vision-language models against human-verified ground truth from the scripted scenarios.

The full participant-facing protocol materials, including task scripts, consent language, and supplementary study materials, are included with the dataset release.

\subsection{Privacy and Anonymization Steps}\
\label{app:data_collection.privacy}

To minimize the publication of identifiable and sensitive information, we publish the RGB video, processed eye gaze data, trajectory data, SLAM point-cloud data, and IMU data only. We do not release the raw audio data, but instead provide transcribed text from audio using the WhisperX~\citep{bain2023whisperx} model and manually remove any private information accidentally divulged by participants. We also blur faces and license plates from the RGB video data using the EgoBlur~\citep{raina2023egoblurresponsibleinnovationaria} model. Outdoor captures of non-participants from passing glances were blurred. However, any direct interaction, such as talking with non-participants, was manually removed from the recordings.

The IRB protocol used a multi-layer privacy plan for incidental capture of non-participants. Public-facing activities were scheduled, when possible, during off-peak times and in low-traffic areas, and the study avoided areas with heightened privacy expectations. For indoor recordings, any segment containing facial data from a non-consenting person was removed from the release. For necessary outdoor public recordings, faces were blurred automatically and then manually verified so missed faces could be removed or redacted. The study team did not actively interact with non-participants; any direct interaction with an outside person was removed from the released dataset.

We used a multi-stage de-identification and quality-control process. Automated tools, including EgoBlur and multimodal LLMs, were used to detect and mask faces and to flag frames containing potentially identifying text such as IDs, name tags, license plates, or personal account information on screens. We then manually reviewed outputs to identify missed faces, unwanted segments, and sensitive text. Only processed data with blurred faces and transcribed, redacted audio is shared broadly.

\subsection{Data, Code, and License}
\label{sec:data_collection.release}
\label{app:data_collection.release}

The dataset is released at \dataURL. Code for dataset processing and evaluation is released at \codeURL. The dataset was collected and created by the authors and will be released under the \textsc{CC BY-NC} license.

\section{Annotation Pipeline Supplementary Information}
\label{app:annotation_pipeline}
\label{sec:annotation_pipeline}
To scale the generation of question-answer pairs in \datasetName, we developed an automated, agentic question-answer generation pipeline. This pipeline leverages multiple specialized LLM agents and is structured into two phases. Each phase goes through human review to maintain high data fidelity.

\subsection{Phase 1: Dense Video Captioning}
\label{sec:annotation.phase1}
The pipeline initializes with egocentric Session Videos and Temporal Metadata (video recording time and date, duration, etc.). Due to the context limits of LLMs, these videos are first split into shorter chunks. Concurrently, an \textit{Audio Extraction} process is performed, with the audio of all sessions for a participant combined and transcribed using \textit{WhisperX}. An \textit{LLM Captioning} agent then processes each video chunk along with the transcription, a \textit{Person Registry} (containing pseudonyms and descriptions of the individuals as they appear in the recordings), and previously accumulated captions. This agent extracts dense descriptions of visual actions, detected objects, auditory events, and conversation summaries. These discrete captions are subsequently temporally aggregated, producing consolidated Video Captions for the session. New individuals are added to the registry upon detection. The first phase is illustrated in \Cref{fig:qa_pipeline_phase1}. To ensure the text accurately reflects the raw video, this intermediate output undergoes an initial stage of \textit{Human Review}.

\subsection{Phase 2: Agentic QA Generation}
\label{sec:annotation.phase2}

Phase 2 consists of $4$ agents: \textit{QA Planner}, \textit{Verifier}, \textit{Retriever}, and \textit{Enhancer}. \Cref{fig:qa_pipeline_phase2} illustrates this phase. It begins with a \textit{Build Super Ledger} process, which aggregates the Metadata and Video Captions from all sessions into a unified ``Super Ledger''. A \textit{QA Planner} agent reads this Super Ledger to propose diverse question-answer pairs designed to target the dataset's core dimensions and tasks~\citep{wang2023selfinstruct}.
We employ a rationale-first schema design principle during question answer generation~\citep{willard2023efficient}. We enforce \textit{instance-level chain-of-thought} by embedding reasoning fields before generating an annotation with a question and answer data~\citep{xiao2024efficient}. This localizes the thought process of the agent to the local context.

Note that the QA Planner only uses the text of the Super Ledger for this.
To ensure that the generated pairs are grounded in the source media, the pipeline employs a closed-loop verification system. For each generated annotation, a \textit{Verifier} agent asks a \textit{Retriever} for relevant information. The Retriever searches the Super Ledger and returns relevant data to the Verifier. The Verifier then evaluates the proposed annotation against a strict set of criteria. Factual correctness ensures that the information is grounded in the data. Objective alignment checks whether the question-answer pair is relevant to the objective of \datasetName, specifically whether it adheres to the predefined tasks and whether the temporal gap is significant enough to justify a possible memory lapse. Causality criteria require the Verifier to ensure the question is answerable using only previously recorded data. Answer Choice Balance asks the Verifier to ensure the answer cannot be inferred from only the question and answer choices. Finally, QA Naturalness criteria ensure that the question is practical and naturally worded.
The Verifier also follows the same principle as the QA planner and generates an individual criterion score reason before generating the actual score~\citep{zheng2023judging}.
Based on this evaluation, the Verifier determines if updates are needed. If updates are required, the pair is routed to an \textit{Enhancer} agent, which updates question-answer pairs according to the verifier's suggestions and feeds them back to the Verifier in an iterative loop. If the verifier gives no suggestions and the annotation is marked as incorrect, the pair is considered unusable and routed to the \textit{Rejected set}. If the annotation is marked correct and has no further suggestions, it is considered approved. Approved pairs are routed to the \textit{Accepted Set}. Finally, the pairs in the Accepted Set undergo a final \textit{Human Review} phase to yield the definitive benchmark dataset.

The QA Planner and Verifier agents use the \modelname{gemini-3.1-pro-preview} model, whereas the LLM Captioning Agent, Video Retriever, and Enhancer agents use \modelname{gemini-3-flash-preview}.

\subsection{Pipeline Cost Analysis}
\label{sec:annotation.cost}

We estimate the variable API cost of the annotation pipeline from duration-based media tokenization rather than frame counts. This distinction matters for long egocentric recordings: in the current Gemini API accounting, video and audio inputs are converted at fixed rates of $263$ and $32$ tokens per second, respectively, for a combined media density of $295$ tokens per second for video with audio.\footnote{\url{https://ai.google.dev/gemini-api/docs/tokens}} Pricing is computed using the May 2026 Gemini Developer API standard paid tier: \modelname{gemini-3-flash-preview} charges $\$0.50$/M input tokens for text/image/video, $\$1.00$/M input tokens for audio, and $\$3.00$/M output tokens; \modelname{gemini-3.1-pro-preview} charges $\$2.00$/M input tokens and $\$12.00$/M output tokens for prompts up to $200$k tokens, with higher prices for larger prompts.\footnote{\url{https://ai.google.dev/gemini-api/docs/pricing}} Unless otherwise stated, the estimates below use online standard pricing, assume no Batch/Flex discounts, and exclude human review, local preprocessing, storage, taxes, and non-LLM compute.

\begin{tcolorbox}[
  enhanced,
  colback=tblNote,
  colframe=tblRule,
  boxrule=0.45pt,
  arc=1.5mm,
  left=1.6mm,
  right=1.6mm,
  top=1mm,
  bottom=1mm
]
\footnotesize
\textbf{Cost model.}
Stage 1 splits each hour into thirty $120$s chunks and runs captioning on Flash. Each chunk contributes $120\times(263+32)=35.4$k media tokens plus approximately $5$k prompt/context tokens. Stage 2 includes one Pro QA-generation pass over the Super Ledger, followed by a closed-loop verification handshake: a text-only verifier request, a Flash ledger-retrieval request, one Pro verifier pass over $10$ evidence clips totaling $300$s, and one Flash enhancer pass over the same evidence budget. We assume three verification loops per QA, $90\%$ cache reuse for the session-level Super Ledger, and $15\%$ reuse for recurring evidence clips. Output lengths are estimated at $2$k tokens per caption chunk, $1.5$k tokens for QA generation, and $0.5$--$1.5$k tokens per verification-loop agent call.
\end{tcolorbox}

\begin{table}[t]
\centering
\footnotesize
\caption{Unit cost assumptions for the SuperMemory annotation pipeline. Costs use current Gemini standard-tier pricing and duration-based video/audio tokenization.}
\label{tab:pipeline_cost_units}
\setlength{\tabcolsep}{4pt}
\renewcommand{\arraystretch}{1.15}
\begin{tabularx}{\linewidth}{@{}>{\raggedright\arraybackslash}p{2.25cm}Y>{\raggedleft\arraybackslash}p{1.65cm}>{\raggedleft\arraybackslash}p{1.65cm}>{\raggedleft\arraybackslash}p{2.2cm}@{}}
\toprule
\rowcolor{tblHeader}
\textbf{Unit} & \textbf{Dominant inputs} & \textbf{Input tokens} & \textbf{Output tokens} & \textbf{Est. cost} \\
\midrule
\rowcolor{tblSubHeader}
\multicolumn{5}{@{}l}{\textbf{Stage 1: Dense video captioning}} \\
1 hour, Flash captioner & $30$ chunks; $946.8$k video tokens, $115.2$k audio tokens, $150$k text/context tokens & $1.212$M & $60$k & $\$0.84$ \\
\midrule
\rowcolor{tblSubHeader}
\multicolumn{5}{@{}l}{\textbf{Stage 2: QA generation, per QA}} \\
QA generation & Super Ledger text at $\sim15$k tokens/hour; $90\%$ cached, \modelname{gemini-3.1-pro-preview} & $15T{+}1$k & $1.5$k & $\$0.0200+\$0.0057T$ \\
\midrule
\rowcolor{tblSubHeader}
\multicolumn{5}{@{}l}{\textbf{Stage 2: Closed-loop verification, per QA per loop}} \\
Verifier setup & QA pair and rubric text on \modelname{gemini-3.1-pro-preview} & $1.0$k & $0.5$k & $\$0.008$ \\
Retriever & Super Ledger text at $\sim15$k tokens/hour; $90\%$ cached, \modelname{gemini-3-flash-preview} & $15T{+}1$k & $1.0$k & $\$0.0035+\$0.001425T$ \\
Verifier evidence pass & $10$ clips totaling $300$s plus prompt; $15\%$ cached, \modelname{gemini-3.1-pro-preview} & $93.5$k & $1.5$k & $\$0.180$ \\
Enhancer evidence pass & Same evidence budget on \modelname{gemini-3-flash-preview} & $93.5$k & $1.5$k & $\$0.049$ \\
\rowcolor{tblOurs}
\textbf{Total loop} & Pro verifier with Flash retriever/enhancer & $195.5{+}15T$k & $4.5$k & \begin{tabular}[t]{@{}r@{}}$\mathbf{\$0.2407}$\\$\mathbf{+\$0.001425T}$\end{tabular} \\
\bottomrule
\end{tabularx}
\vspace{0.25em}
\begin{flushleft}
\footnotesize \textit{Note.} $T$ denotes total session duration in hours. For $T{=}1$, QA generation costs approximately $\$0.026$ and one closed-loop verification iteration costs approximately $\$0.242$; at three loops, generation plus verification is $\$0.752$ per QA. At $T{=}50$, ledger context dominates the text-only generation/retrieval steps, raising generation plus three verification loops to approximately $\$1.241$ per QA.
\end{flushleft}
\end{table}

\begin{table}[htbp]
\centering
\small
\caption{Worst case scale projection for the default model mix: Flash captioning, retrieval, and enhancement; Pro QA generation and verification; and three verification loops per QA.}
\label{tab:pipeline_cost_scale}
\setlength{\tabcolsep}{4pt}
\renewcommand{\arraystretch}{1.12}
\begin{tabularx}{0.92\linewidth}{@{}>{\centering\arraybackslash}p{1.45cm}>{\centering\arraybackslash}p{1.8cm}>{\raggedleft\arraybackslash}p{1.45cm}>{\raggedleft\arraybackslash}p{2.1cm}>{\raggedleft\arraybackslash}X@{}}
\toprule
\rowcolor{tblHeader}
\textbf{Hours} & \textbf{QA density} & \textbf{QAs} & \textbf{Total tokens} & \textbf{Default cost} \\
\midrule
\rowcolor{tblSubHeader}
\multicolumn{5}{@{}l}{\textbf{Small-to-medium annotation runs}} \\
$1$ & $0.5$/min & $30$ & $21.1$M & $\$23.40$ \\
$1$ & $1.0$/min & $60$ & $41.0$M & $\$45.97$ \\
$1$ & $2.0$/min & $120$ & $80.8$M & $\$91.09$ \\
$2$ & $0.5$/min & $60$ & $45.9$M & $\$47.41$ \\
$2$ & $1.0$/min & $120$ & $89.2$M & $\$93.13$ \\
$2$ & $2.0$/min & $240$ & $175.9$M & $\$184.57$ \\
$10$ & $0.5$/min & $300$ & $373.5$M & $\$260.98$ \\
$10$ & $1.0$/min & $600$ & $734.2$M & $\$513.52$ \\
$10$ & $2.0$/min & $1{,}200$ & $1.456$B & $\$1{,}018.60$ \\
\midrule
\rowcolor{tblOurs}
$50$ & $1.67$/min & $5{,}000$ & $18.076$B & $\mathbf{\$6{,}246.18}$ \\
\bottomrule
\end{tabularx}
\vspace{0.35em}
\begin{flushleft}
\footnotesize The $50$ hour, $5{,}000$ QA extrapolation is intentionally conservative for text context: it sends a $15$k-token/hour Super Ledger prefix during Pro QA generation and each Flash retrieval loop, with cache-aware pricing. If the planner or retriever first narrows the ledger with an index or symbolic search before calling Gemini, the Stage 2 text cost drops; if evidence bundles exceed $300$s or Pro requests cross the $200$k-token pricing threshold, costs rise. Note, the above data is for the worst case when all QA pairs are resent for enhancement. From our experience, this number falls by 30-50\% after every iteration. \textbf{The approximate cost for our data was around \$3900}
\end{flushleft}
\end{table}

The dominant cost driver is therefore not Stage 1 captioning but repeated multimodal evidence verification and Pro QA generation over a long Super Ledger. For the default configuration, captioning $50$ hours costs approximately $\$42$, whereas Pro QA generation plus three-loop verification for $5{,}000$ QAs costs approximately $\$6.20$k under the cache-aware ledger model. This suggests that future scaling should prioritize (i) evidence-clip pruning before verifier calls, (ii) reusing cached high-value evidence clips across related QAs, and (iii) indexing the Super Ledger before QA generation and retrieval.

\section{Verification Criteria and Annotation Format}
\label{app:verification_annotation_format}

\subsection{Verification Criteria}
\label{app:verification_criteria}
To guarantee that the \datasetName benchmark meets the highest standard of scientific rigor, we employ a multi-stage validation pipeline consisting of agentic scoring, deterministic temporal filtering, structural schema verification, and visual spatial-grounding validation.
\begin{enumerate}
    \item \textbf{Agentic Quality Scoring (Stage 2 Verifier Agent)}:
    The candidate QA pairs are first evaluated across three continuous dimensions ($\in [0.0, 1.0]$) by a specialized Verifier Agent:
    \begin{itemize}
        \item \textit{Factual Correctness:} Ensures complete alignment between the answer text and visual/audio evidence.
        \item \textit{Objective Relevance:} Evaluates question logicality, requiring a minimum temporal gap between evidence and query ($>10$ minutes) and multi-clip reasoning.
        \item \textit{Causal Answerability:} Verifies that the question can be resolved using only the evidence available up to the question's execution timestamp.
    \end{itemize}
    A candidate pair is rejected if any score falls below the threshold $\tau = 0.6$.
    \item \textbf{Deterministic Causal Filtering and Temporal Grounding}:
    Beyond agentic intuition, the pipeline programmatically enforces strict physical causality:
    \begin{itemize}
        \item \textit{Causal Evidence Pruning:} A deterministic filter calculates the absolute timeline offsets of all video chunks. Any answer evidence whose start timestamp $t_{\text{evidence}}$ is greater than the question start timestamp $t_{\text{question}}$ is programmatically pruned from the final dataset.
        \item \textit{Anchor Validity:} All timestamps (MM:SS) must strictly fall within the boundaries of the associated video segments, preventing temporal hallucinations or drift.
    \end{itemize}
    \item \textbf{Structural and Semantic Integrity Verification}:
    Every verified pair is validated against a strict Pydantic-enforced schema to guarantee downstream compatibility:
    \begin{itemize}
        \item \textit{Multiple Choice Divergence:} The stored schema for every QA pair contains exactly three custom choices. For answerable queries ($\text{is\_answerable} = \text{True}$), these consist of exactly one `correct', one `vague' (technically correct but ambiguous), and one `incorrect' distractor. For unanswerable queries ($\text{is\_answerable} = \text{False}$), all three choices are classified as `incorrect' distractors. During evaluation, a standard, fixed fourth option (``N/A'' or ``This question cannot be answered'') is dynamically appended to form a complete four-way multiple-choice setup. Each choice contains an `explanation' field justifying its classification.
        \item \textit{Coherent Room \& Modality Mapping:} Both questions and answers are validated for spatial and sensory consistency by requiring a designated room/location tag and a combination of active sensor modalities (e.g., \textit{Video, Audio, Gaze, Trajectory, Depth, OCR}).
        \item \textit{Task-Category Alignment:} Every pair is strictly classified under one of our six defined task categories (e.g., \textit{object\_location\_memory, timeline\_reconstruction}), and checked for alignment with that category's specific reasoning patterns.
    \end{itemize}
\end{enumerate}
Only QA pairs passing all agentic, programmatic, and structural criteria are kept in the final output.

\begin{figure}[t]
    \centering
    \includegraphics[width=\linewidth]{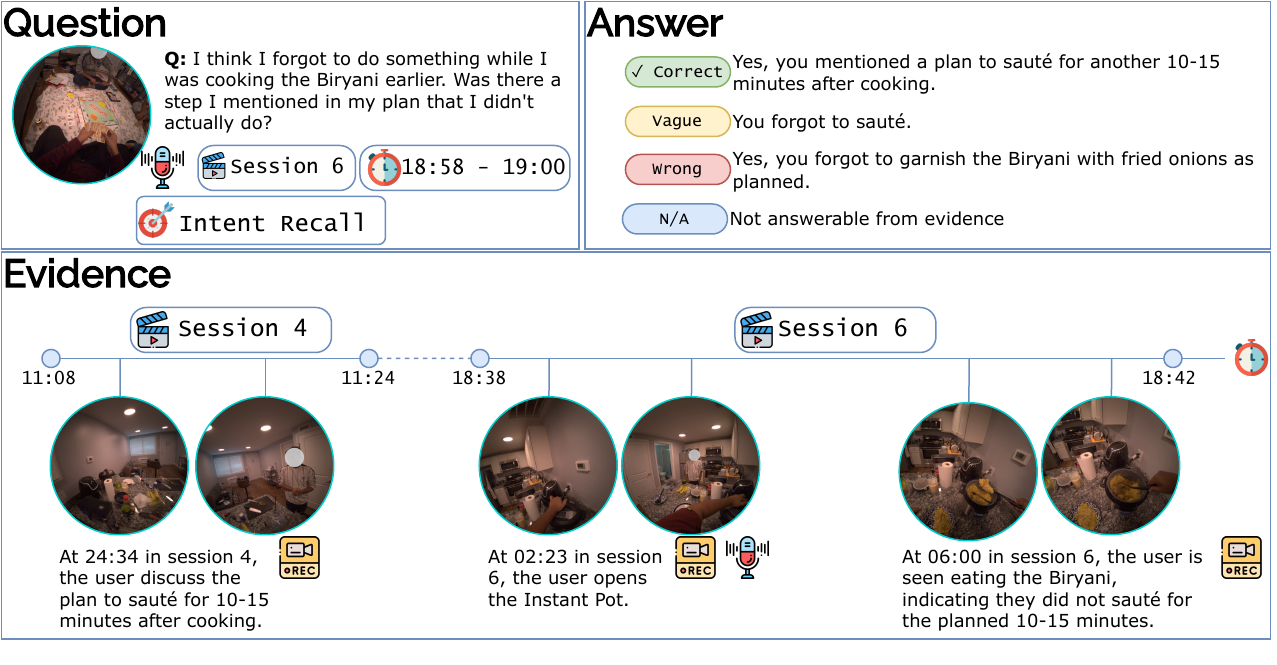}
    \caption{Question Answer Annotation in \datasetName}
    \label{fig:annotation}
\end{figure}

\subsection{Detailed Input/Output Format}
\label{app:annotation.output_format}
\label{sec:annotation.output_format}

Question-Answer pairs in \datasetName are multi-modal, i.e., both question and answers may require input from multiple modalities to comprehend. In addition to the text, questions reference a time span, indicating where the query may realistically be asked. Each question also has metadata like location details and relevant sensor modalities needed to understand the question, e.g., Video, Audio, Gaze.

Answers in \datasetName are naturally phrased multiple-choice sentences. For evaluation, each question is presented with four ordered choices:
\begin{description}[leftmargin=2.2cm, labelwidth=1.8cm, labelsep=0.35cm, nosep]
    \item[\CorrectTag] An accurate answer that directly addresses the query.
    \item[\VagueTag] A technically plausible answer that is ambiguous, underspecified, or less useful.
    \item[\WrongTag] An answer that is inconsistent with the evidence.
    \item[\NATag] An abstention option for queries that lack sufficient grounding in the available evidence.
\end{description}
Each answer is grounded in video evidence, which can span multiple sessions to evaluate long-term memory and recall. Each evidence item includes a text description, referenced session, and video IDs, video time span, location, and relevant sensor modalities. The evidence for an answer always occurs before the question starts.

\begin{schemabox}[label={schema:final_qa_format_full}]
{Final Verified QA Pair}
\footnotesize
\setlength{\columnsep}{1.0em}
\begin{multicols}{2}
\dirtree{%
.1 OutputQAPairFormat.
.2 balance\_reasoning: String.
.2 question: Object.
.3 text: String \hfill first-person query.
.3 question\_reasoning: String \hfill temporal justification.
.3 room: String \hfill ledger environment.
.3 video\_id: String \hfill primary video.
.3 modalities[]: List[Modality].
.3 time\_spans[]: List[Object].
.4 video\_id: String.
.4 start\_time: String \hfill MM:SS.
.4 end\_time: String \hfill MM:SS.
.2 answer: Object.
.3 text: String.
.3 is\_answerable: Boolean.
.3 evidence\_list[]: List[Object].
.4 reason: String \hfill ledger-grounded evidence description.
.4 room: String.
.4 video\_id: String.
.4 time\_span: Object.
.5 start\_time: String \hfill MM:SS.
.5 end\_time: String \hfill MM:SS.
.4 modalities[]: List[Modality].
.3 answer\_choices[]: List[Object].
.4 explanation: String.
.4 choice\_ltype: ChoiceType.
.4 text: String.
}
\columnbreak
\dirtree{%
.1 OutputQAPairFormat.
.2 metadata: Object.
.3 task: MemoryTask.
.3 task\_reasoning: String.
.3 primary\_video\_id: String.
.3 verification\_score: Object.
.4 factual\_correctness\_score: Float.
.4 factual\_correctness\_reasoning: String.
.4 objective\_correctness\_score: Float.
.4 objective\_correctness\_reasoning: String.
.4 causal\_answerability\_score: Float.
.4 causal\_answerability\_reasoning: String.
.4 naturalness\_score: Float.
.4 naturalness\_reasoning: String.
.4 is\_correct: Boolean.
.4 is\_guessable: Boolean.
.4 guessability\_justification: String.
.4 contains\_pii: Boolean.
.4 privacy\_audit\_reasoning: String.
.4 is\_salvageable: Boolean.
.4 suggestions[]: List[String].
.4 suggested\_chunks[]: List[Object].
.5 video\_id: String.
.5 start\_time: String \hfill MM:SS.
.5 end\_time: String \hfill MM:SS.
.5 relevance\_reason: String.
}
\end{multicols}
\end{schemabox}

\section{Dataset Statistics Supplementary Information}
\label{sec:data_stats}

\begin{figure}[t]
    \centering
    \begin{subfigure}[b]{0.45\textwidth}
        \centering
        \includegraphics[width=\textwidth]{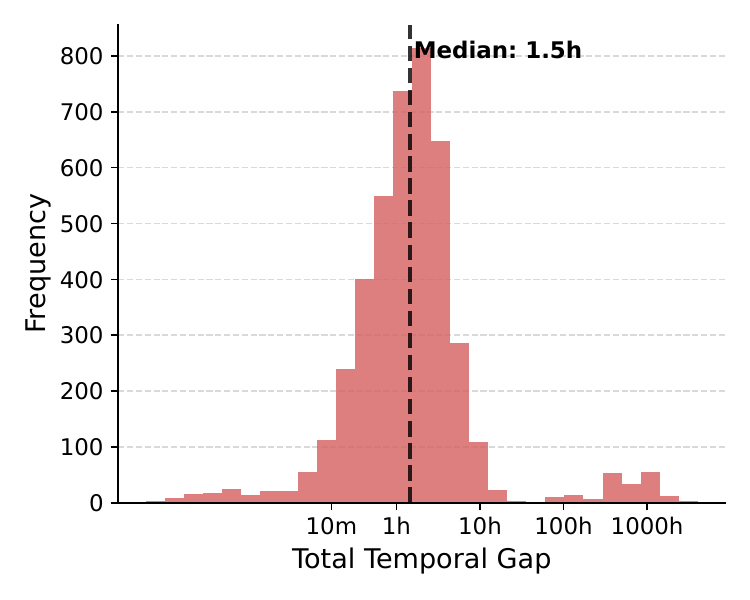}
        \caption{Temporal Gap Distribution}
        \label{fig:temporal_gap}
    \end{subfigure}
    \hfill
    \begin{subfigure}[b]{0.49\textwidth}
        \centering
        \includegraphics[width=\textwidth]{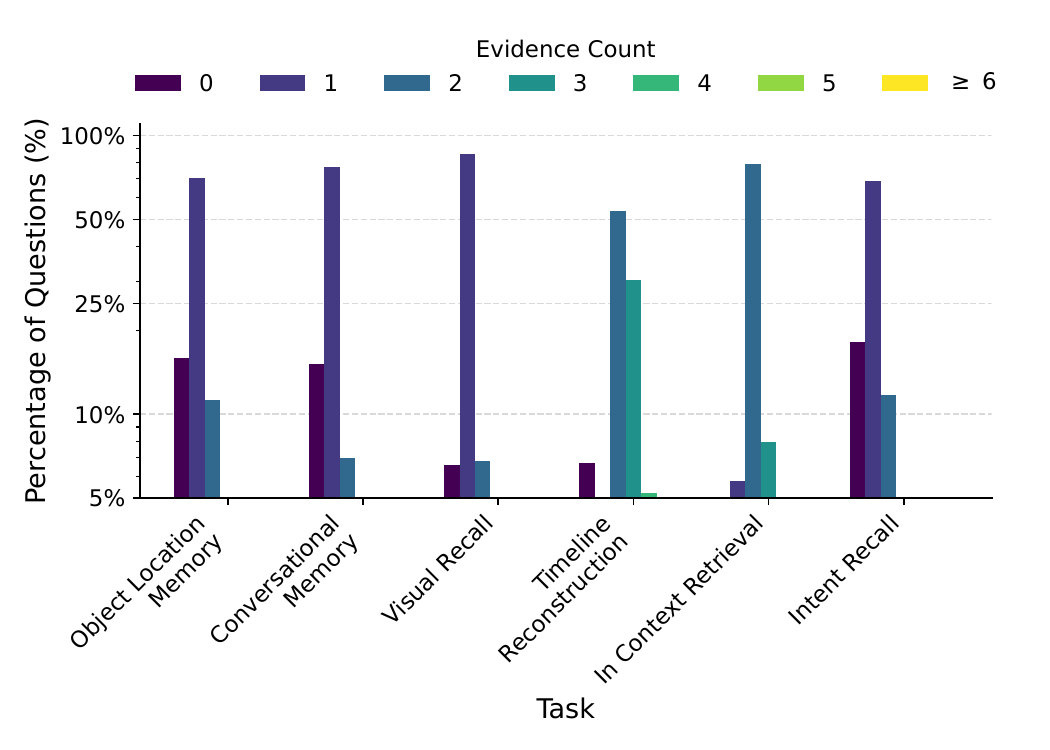}
        \caption{Evidence Count by Task}
        \label{fig:evidence_count_by_skill}
    \end{subfigure}
    \caption{Overview of dataset statistics (a) temporal gap (in hours) between the timestamp of the visual evidence and the user's query, and (b) the number of distinct evidence clips required to answer a question.}
    \label{fig:combined_stats}
\end{figure}

\textbf{Evidence Complexity:} The amount of context a model needs depends heavily on the task (\Cref{fig:evidence_count_by_skill}). Tasks such as object localization typically require grounding in only a single temporal or spatial segment. However, complex tasks such as Timeline Reconstruction require models to piece together clues across multiple separate clips.

\textbf{Temporal Reasoning Horizons:} The \datasetName benchmark is also designed such that answering a question requires searching over long temporal horizons. \Cref{fig:temporal_gap} illustrates the ``temporal gap''---the time elapsed between a user asking a question and when the evidence required to answer it was recorded. While our dataset includes extreme cases where models may have to recall events recorded over a week ago, most questions fall in the $1$ to $2$ hour range.

\begin{figure}[t]
    \centering
    \begin{subfigure}[b]{0.49\textwidth}
        \centering
        \includegraphics[width=\textwidth]{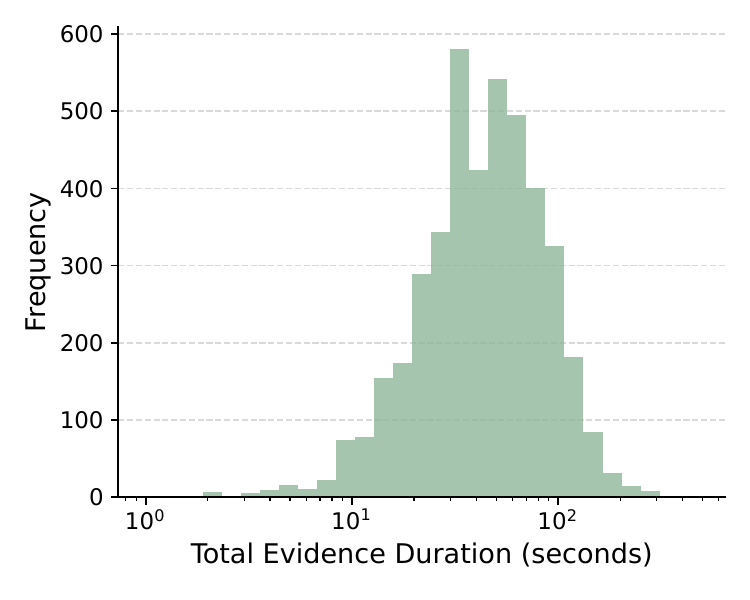}
        \caption{Evidence Duration Distribution}
        \label{fig:evidence_duration}
    \end{subfigure}
    \hfill
    \begin{subfigure}[b]{0.49\textwidth}
        \centering
        \includegraphics[width=\textwidth]{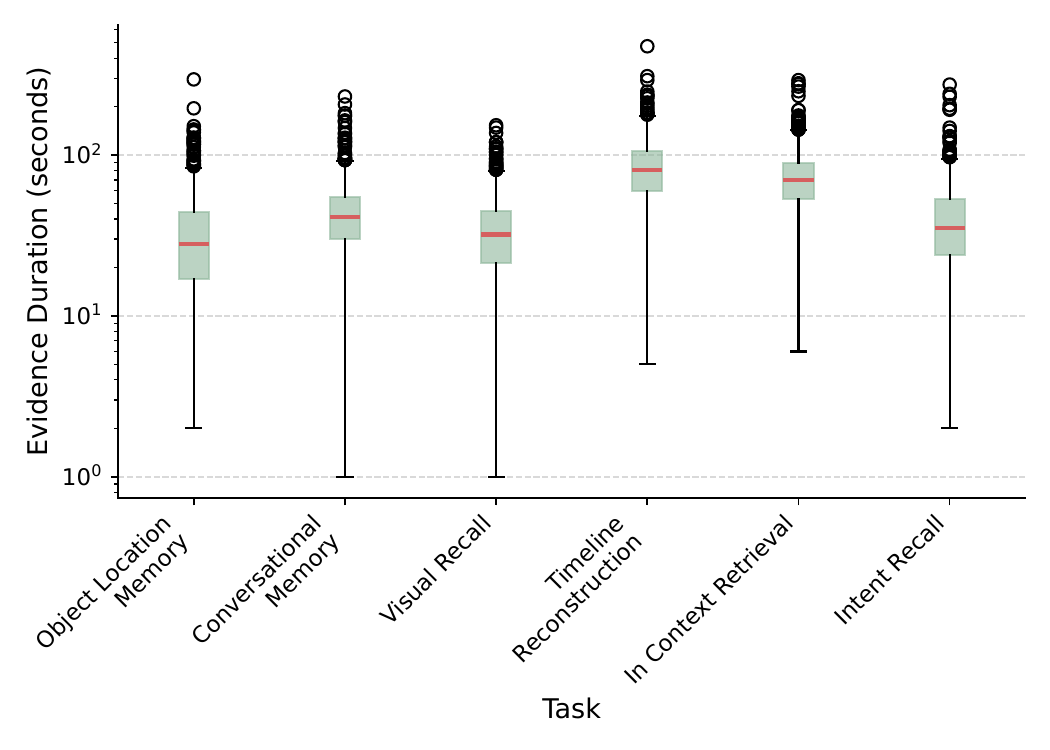}
        \caption{Evidence Duration by Task}
        \label{fig:evidence_duration_by_skill}
    \end{subfigure}
    \caption{Additional dataset statistics showing (a) the total duration of evidence clips required to answer each question, and (b) the distribution of evidence duration across task categories.}
    \label{fig:combined_evidence_duration_stats}
\end{figure}

\textbf{Evidence Duration:} In addition to the number of evidence clips, we measure how much total video time must be inspected to answer each question (\Cref{fig:evidence_duration}). This statistic captures a complementary source of difficulty: some questions can be grounded by a short visual moment, while others require tracking longer activities or aggregating observations over extended periods. Breaking evidence duration down by task category further shows which categories require sustained temporal grounding rather than isolated retrieval (\Cref{fig:evidence_duration_by_skill}).

\section{Human Review}
\label{app:human_review}
\label{sec:human_review}

\begin{figure}[htpb]
    \centering
    \begin{subfigure}[b]{0.49\textwidth}
        \includegraphics[width=\linewidth]{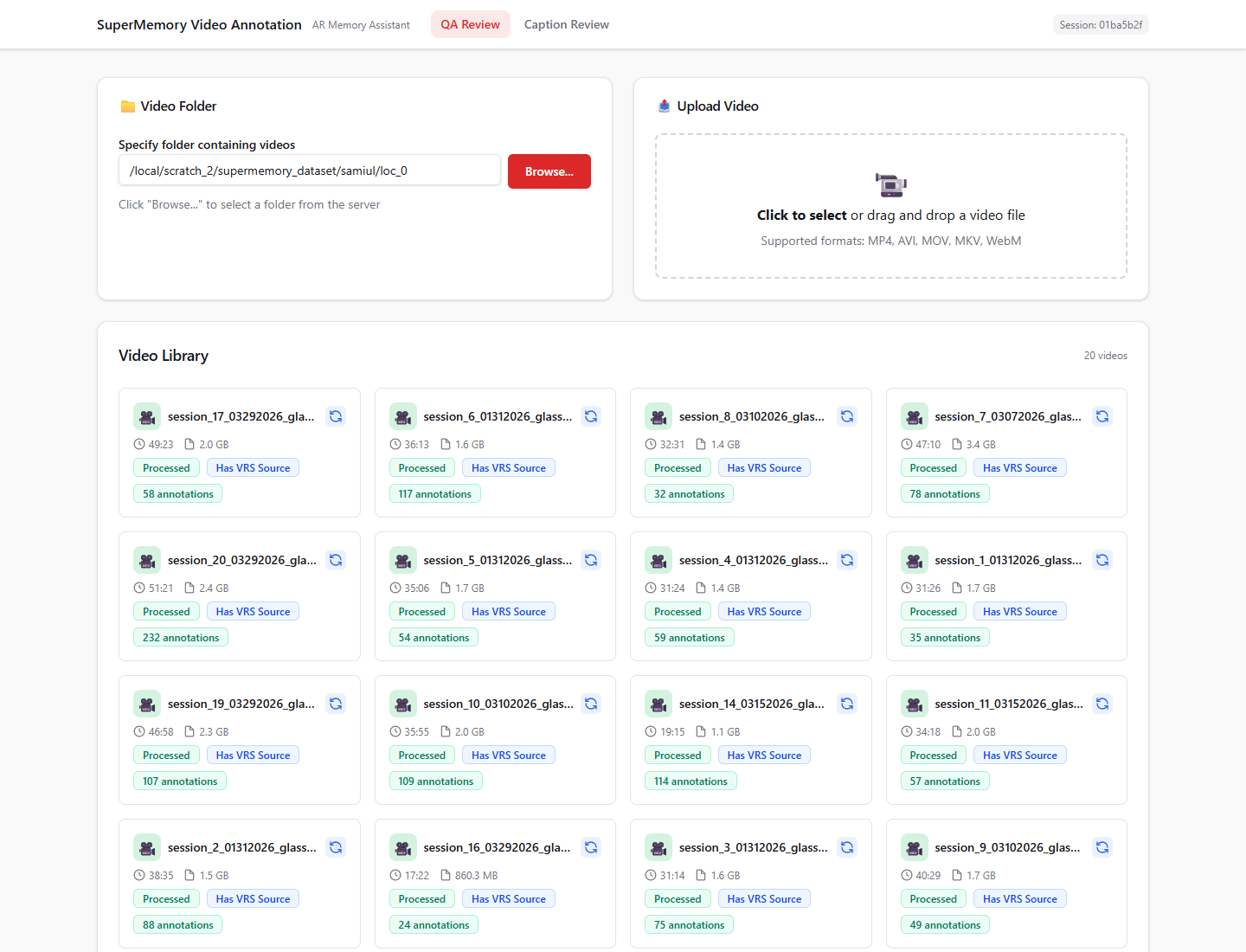}
        \caption{Initial shared review screen.}
        \label{fig:human_review_initial}
    \end{subfigure}
    \hfill
    \begin{subfigure}[b]{0.49\textwidth}
        \includegraphics[width=\linewidth]{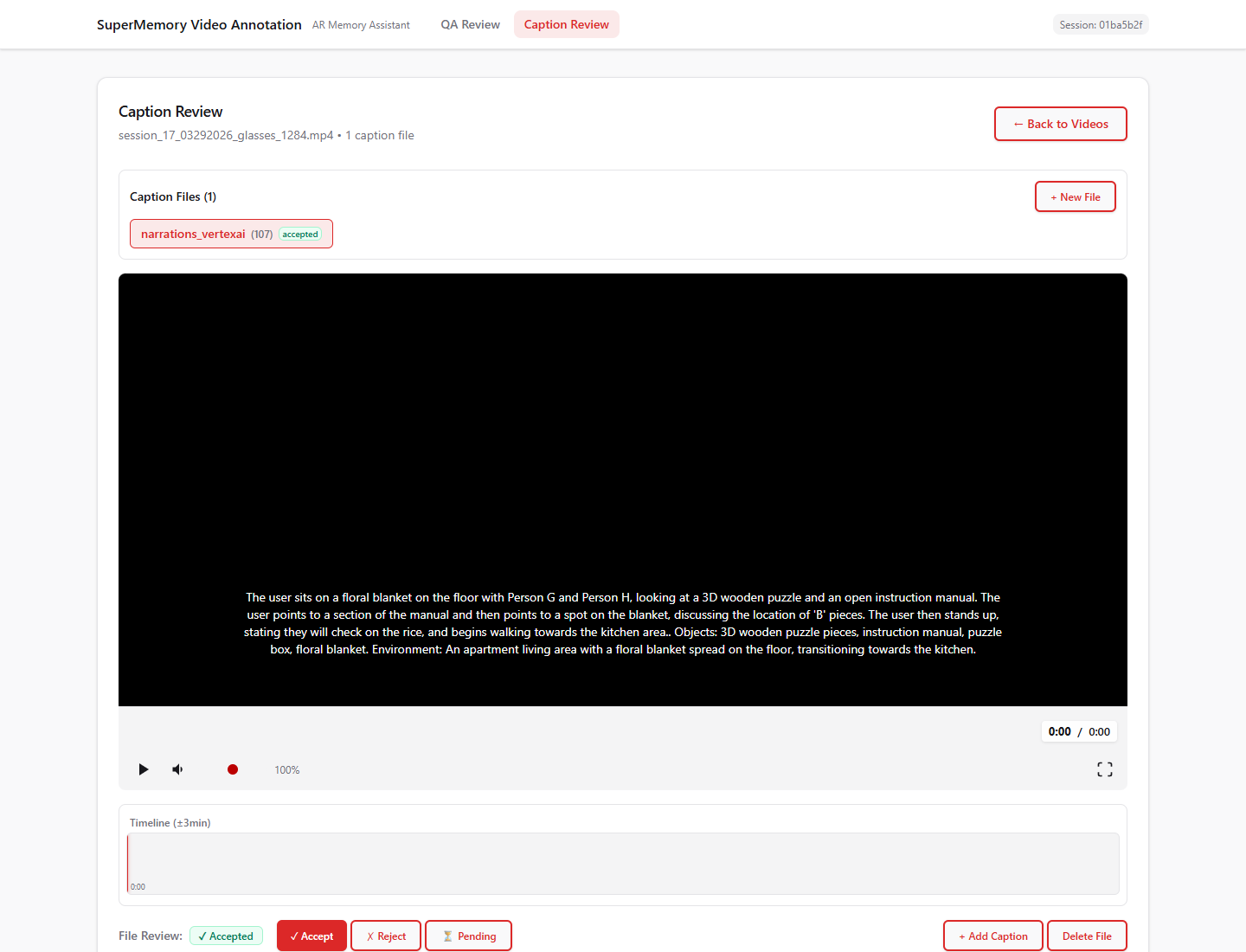}
        \caption{Caption file and local timeline.}
        \label{fig:human_review_caption_review}
    \end{subfigure}

    \vspace{0.5em}

    \begin{subfigure}[b]{0.49\textwidth}
        \includegraphics[width=\linewidth]{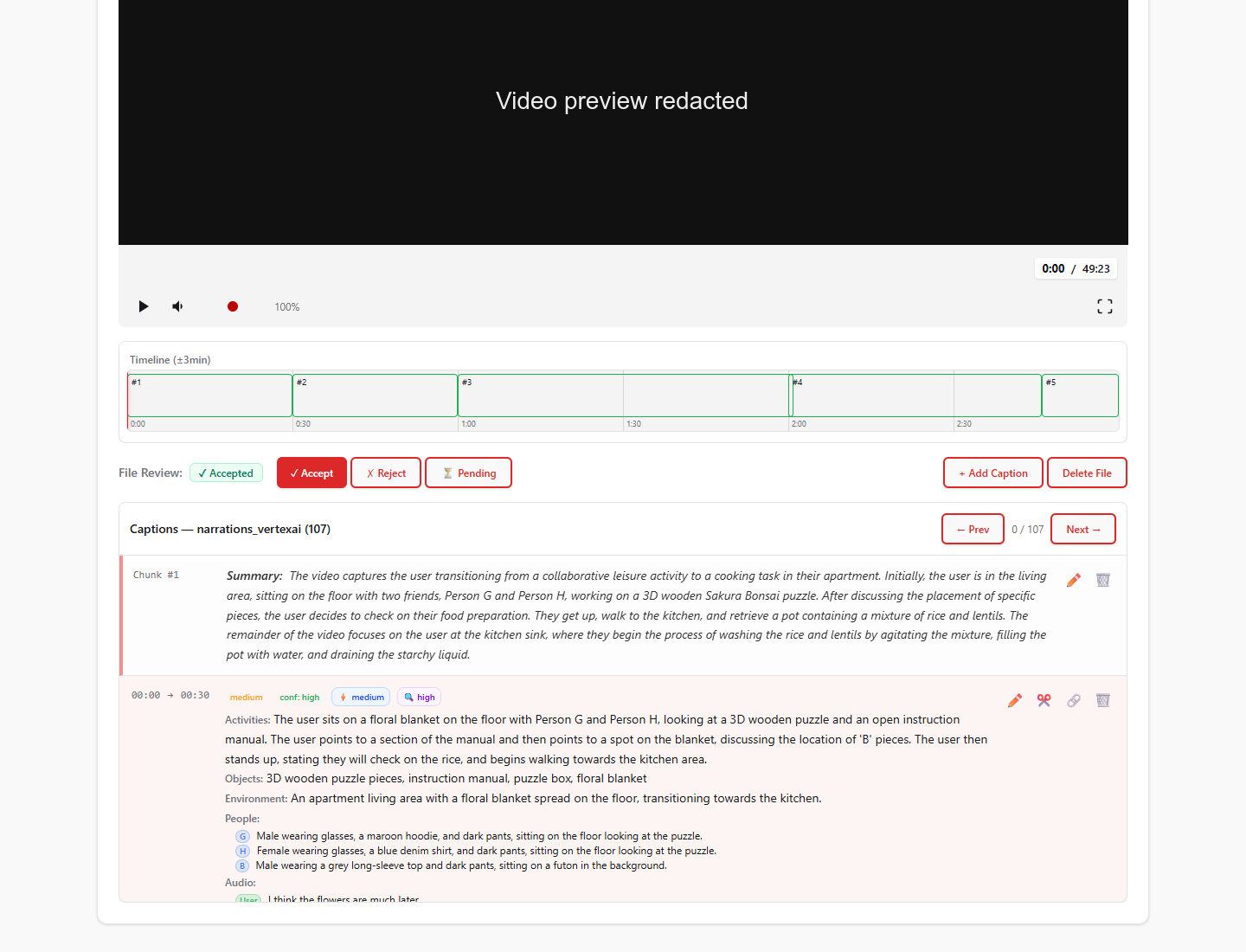}
        \caption{Caption chunks and review controls.}
        \label{fig:human_review_caption_chunks}
    \end{subfigure}
    \hfill
    \begin{subfigure}[b]{0.49\textwidth}
        \includegraphics[width=\linewidth]{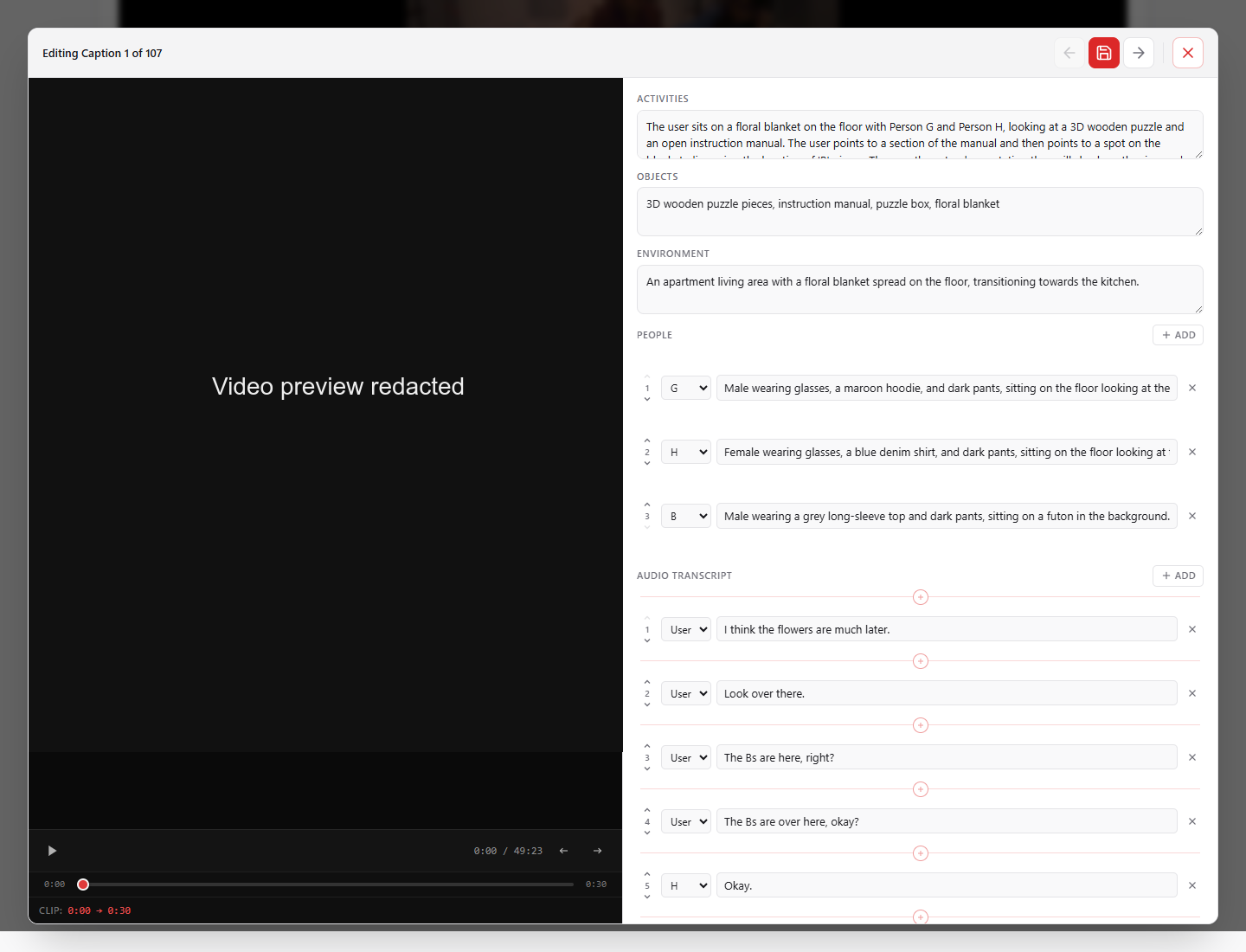}
        \caption{Caption edit screen.}
        \label{fig:human_review_caption_edit}
    \end{subfigure}
    \caption{Caption-review flow after Phase 1. Reviewers select a video, inspect caption files and temporal chunks, assign file-level review status, and edit structured caption fields. Video previews are redacted in the figure for privacy.}
    \label{fig:human_review_caption_flow}
\end{figure}

\begin{figure}[htpb]
    \centering
    \begin{subfigure}[b]{0.49\textwidth}
        \includegraphics[width=\linewidth]{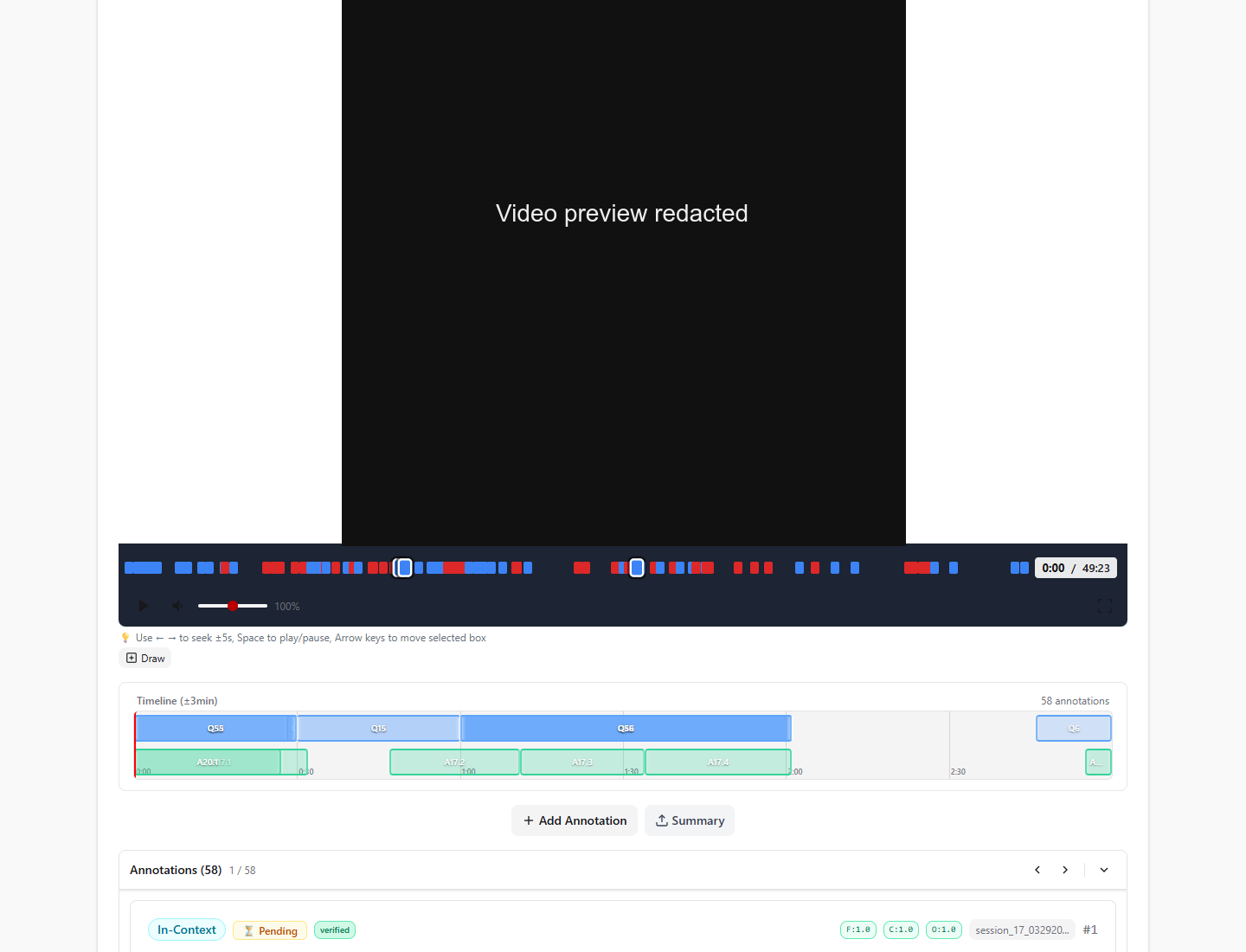}
        \caption{QA review timeline and annotation panel.}
        \label{fig:human_review_qa_review}
    \end{subfigure}
    \hfill
    \begin{subfigure}[b]{0.49\textwidth}
        \includegraphics[width=\linewidth]{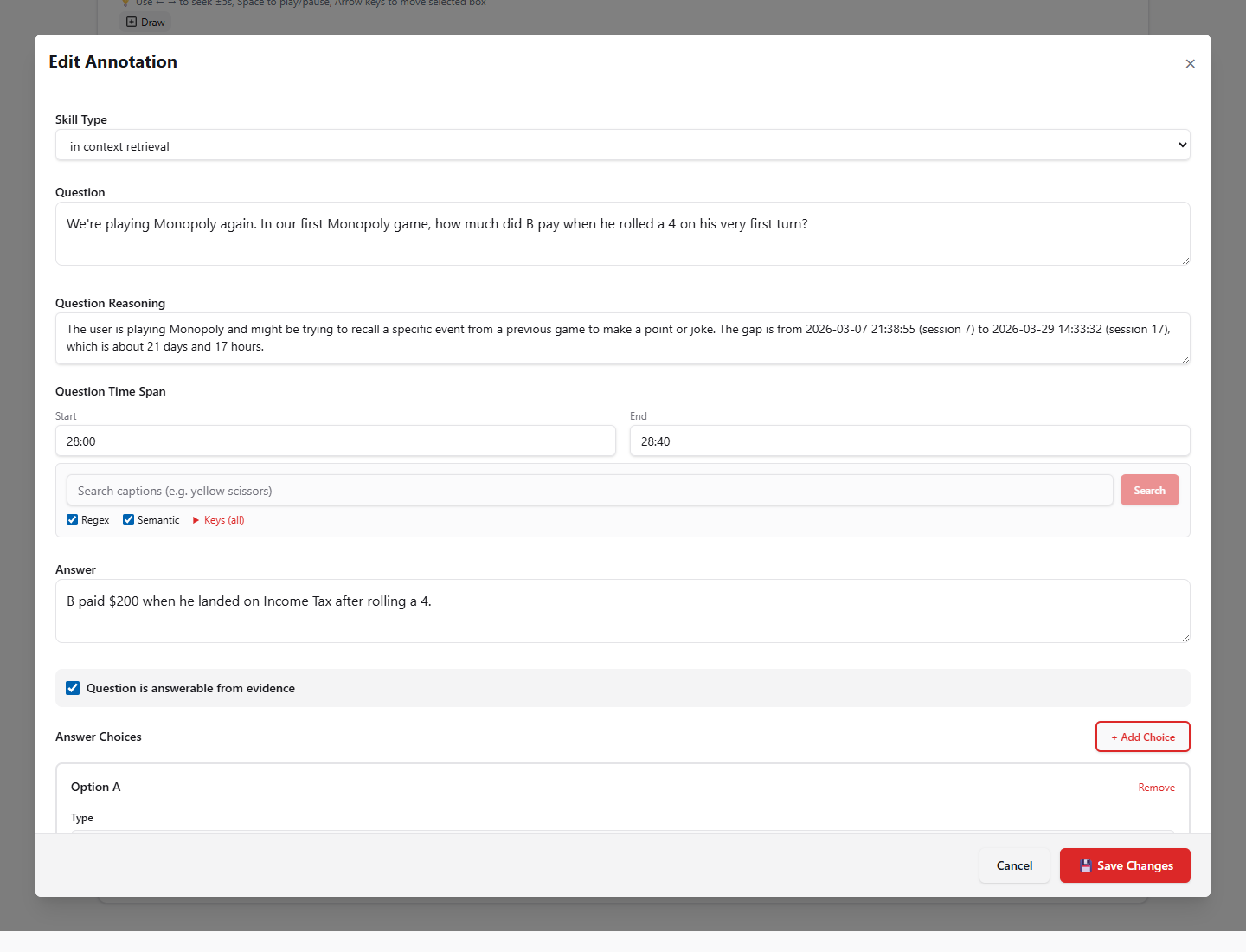}
        \caption{QA edit screen.}
        \label{fig:human_review_qa_edit}
    \end{subfigure}
    \caption{QA-review flow after Phase 2. Reviewers inspect temporal evidence markers and generated question-answer annotations, then edit task labels, question text, reasoning, answer choices, evidence, and human-review status before saving. Video previews are redacted in the figure for privacy.}
    \label{fig:human_review_qa_flow}
\end{figure}

Human review is the final quality-control step after each automated annotation phase. The same review platform supports two reviewer workflows (Figures~\ref{fig:human_review_caption_flow} and~\ref{fig:human_review_qa_flow}). After Phase 1, reviewers use the caption-review interface to inspect the dense video captions produced for each session. The interface loads a processed video, the corresponding caption file, a local timeline window, and the caption chunks generated for that interval. Reviewers can mark a caption file as accepted, rejected, or pending, add new captions, and edit, split, link, or delete individual chunks. In edit mode, each caption exposes structured fields for activities, objects, environment, people, audio transcripts, and timestamps. This stage checks that the captions preserve the visual activity, objects, people, environment, audio events, and text observations needed for downstream question generation.

After Phase 2, reviewers use the QA-review interface to inspect generated question-answer annotations against the source video and evidence timeline. The platform displays accepted and rejected annotation files, the video source, dense temporal markers for questions and supporting evidence, and a side panel for the current annotation. Reviewers can navigate among annotations, review task labels and verification scores, inspect answer options, add or delete annotations, and open raw model responses or model reasoning when debugging failures. In edit mode, reviewers can update the task category, question, reasoning, question time span, answer text, answer choices, evidence spans, modalities, bounding boxes, and human-review status before saving. This workflow lets reviewers verify grounding, factual correctness, answerability, and option quality before examples are included in the final benchmark.

The human review stage is part of the annotation workflow rather than a separate post-hoc audit. Reviewers checked factual grounding against the source video and evidence timeline, temporal causality, answerability, question naturalness, and answer-choice balance before examples were included in the final benchmark. Reviews were performed by the session attendant or reviewer familiar with the task context, which helped resolve ambiguous references and intended actions. Because this workflow is not an independent user-utility study, we separately use participant survey responses in \Cref{sec:results.survey} to assess perceived realism and usefulness.

\section{Evaluation Protocol and Data Use}
\label{app:evaluation_protocol}
\label{sec:experiment.protocol}

The numbers in \Cref{tab:supermemory_results} use \datasetName as an open zero-shot evaluation benchmark. The frameworks and VLMs reported there are not trained or fine-tuned on \datasetName QA labels; each system receives the question, answer choices, and only evidence available before the question time. We intentionally release labels rather than maintaining a hidden test split, prioritizing reproducibility and flexible research use over leaderboard-style evaluation. The released labels may also support finetuning, retrieval diagnostics, and other supervised analyses, but researchers should clearly state whether a result is zero-shot, finetuned, or otherwise uses \datasetName supervision.

\textbf{Baseline modality use.}
The released dataset includes RGB video, transcripts, gaze, motion, trajectory, IMU, and SLAM-derived data, but the current baselines do not exhaust all sensor streams. In practice, Video-RAG mainly consumes RGB frames, ASR transcripts, OCR, object detections, and retrieved auxiliary text, while EgoButler uses RGB/audio-derived clip captions and hierarchical text memories. Thus \Cref{tab:supermemory_results} should be read as a strong baseline suite for current long-video memory agents, not as an upper bound on methods that explicitly exploit gaze, trajectory, IMU, or SLAM. Currently, there are no agentic system baselines that fully utilize all these modalities.

\section{Reproducibility and Compute}
\label{app:evaluation_compute}
\label{sec:evaluation_compute}

We evaluate all systems using the metrics reported in the main paper: answerability F1, QA accuracy, and QA mean reciprocal rank.
Open-source model evaluations were run on a server with $4\times$A100 GPUs. Gemini-family closed-source models were accessed through the Google Cloud Platform API, and OpenAI-family models were accessed through Azure OpenAI APIs. The released code repository contains the evaluation scripts needed to reproduce the reported baseline comparisons.

\end{document}